\documentclass[11pt]{article}
\PassOptionsToPackage{dvipsnames,table}{xcolor}
\usepackage[utf8]{inputenc}
\usepackage[T1]{fontenc}
\usepackage{lmodern}
\usepackage[margin=1in]{geometry}
\usepackage[round,authoryear]{natbib}
\usepackage{url}
\usepackage{booktabs}
\usepackage{microtype}
\usepackage[dvipsnames, table]{xcolor}
\usepackage{arydshln}
\newcommand{\modelseparator}{\noalign{\vskip 2pt}\hdashline[4pt/2pt]\noalign{\vskip 2pt}}
\usepackage{graphicx}
\usepackage{amsmath,amssymb}
\usepackage{pifont}
\usepackage{fontawesome5}
\usepackage{enumitem}
\usepackage{placeins}
\usepackage{float}
\usepackage[font=small,labelfont=bf]{caption}
\usepackage{titlesec}
\titleformat{\section}{\large\bfseries}{\thesection}{0.8em}{}
\titleformat{\subsection}{\normalsize\bfseries}{\thesubsection}{0.8em}{}
\titlespacing*{\section}{0pt}{1.6ex plus .4ex minus .2ex}{0.8ex plus .2ex}
\titlespacing*{\subsection}{0pt}{1.2ex plus .3ex minus .2ex}{0.6ex plus .1ex}
\titlespacing*{\paragraph}{0pt}{0.6ex plus .2ex}{0.8em}
\setlist{itemsep=2pt,topsep=3pt}

\usepackage{bm}

\def\vt{{\bm{t}}}
\def\vv{{\bm{v}}}

\newcommand{\cmark}{\ding{51}}
\newcommand{\xmark}{\ding{55}}

\definecolor{accentorange}{HTML}{FA500E}
\definecolor{heat}{HTML}{6C8EBF}
\definecolor{accentblue}{HTML}{10069F}
\definecolor{C4}{RGB}{0,148,181}
\newcommand\myshade{100}

\usepackage[
citecolor=accentblue!\myshade!black,
colorlinks=true,
linkcolor=accentblue!\myshade!black,
urlcolor=C4,
bookmarksnumbered=true
]{hyperref}
\usepackage[nameinlink]{cleveref}
\crefname{figure}{Fig.}{Figs.}\Crefname{figure}{Fig.}{Figs.}
\crefname{table}{Tab.}{Tabs.}\Crefname{table}{Tab.}{Tabs.}

\crefname{section}{Sec.}{Secs.}\Crefname{section}{Sec.}{Secs.}
\crefname{subsection}{Sec.}{Secs.}\Crefname{subsection}{Sec.}{Secs.}
\crefname{subsubsection}{Sec.}{Secs.}\Crefname{subsubsection}{Sec.}{Secs.}
\crefname{appendix}{App.}{Apps.}\Crefname{appendix}{App.}{Apps.}
\crefname{subappendix}{App.}{Apps.}\Crefname{subappendix}{App.}{Apps.}
\crefname{subsubappendix}{App.}{Apps.}\Crefname{subsubappendix}{App.}{Apps.}
\hypersetup{pdftitle={Clipstral: Image, Video, and Multilingual Text Embedder},pdfauthor={Mistral AI}}
\newcommand{\linkicon}[1]{\raisebox{-0.25em}{#1}}
\newcommand{\ours}{Clipstral}
\newcommand{\vstarcls}{\texorpdfstring{V$^*$-classification}{V*-classification}}

\makeatletter
\renewcommand{\@maketitle}{%
  \begin{center}%
\rule{\linewidth}{3pt}\par
\vspace{1.2em}
    {\LARGE\bfseries \@title \par}%
\vspace{1.2em}
\rule{\linewidth}{0.6pt}\par
  \end{center}%
\vspace{0.6em}}
\makeatother

\title{Clipstral:\\Image, Video, and Multilingual Text Embedder}
\date{}

\begin{document}

\maketitle

\begin{center}
\includegraphics[width=0.8\linewidth,keepaspectratio]{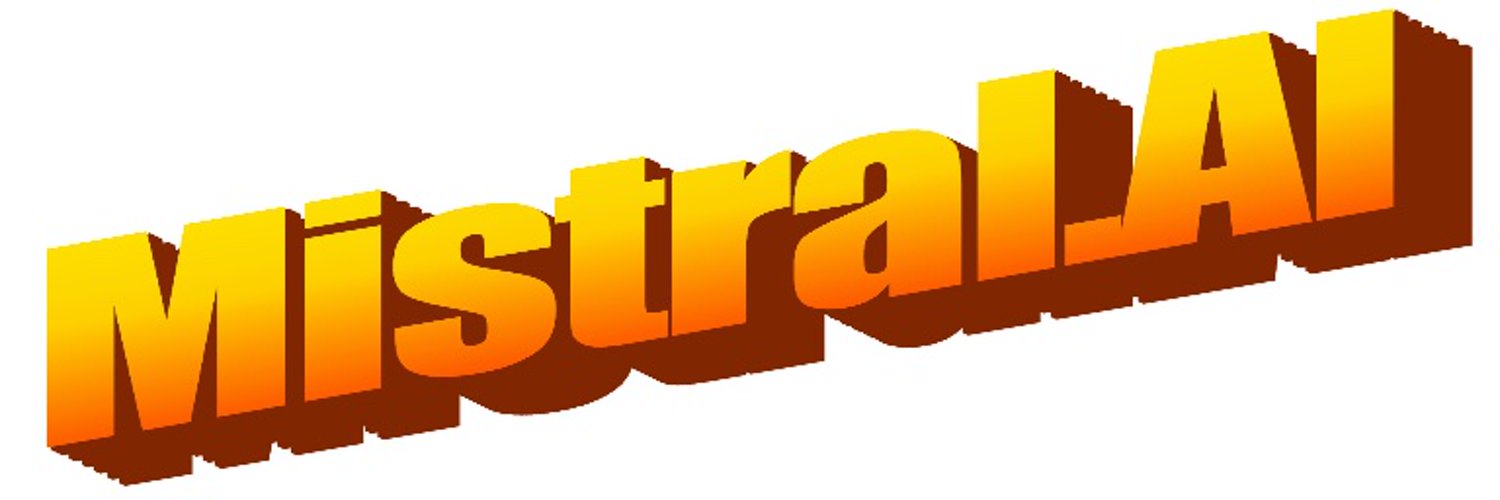}
\end{center}

\newcommand{\oursParams}{1.24B}
\newcommand{\oursVisParams}{409M}
\newcommand{\oursSTParams}{50M}
\newcommand{\oursTxtParams}{783M}

\newcommand{\oursImgClsAvg}{79.2}

\newcommand{\oursVstarCrop}{67.5}

\newcommand{\vstarChance}{36.0}
\newcommand{\qwenParamRatio}{1.7}
\newcommand{\qwenEightParamRatio}{6.6}

\newcommand{\dVidClsVsQwen}{0.8}
\newcommand{\clsPeG}{81.6}

\newcommand{\truncOtherMax}{1.6}

\newcommand{\lxXmOthersEngLo}{3.7}
\newcommand{\lxXmOthersEngHi}{11.6}

\newcommand{\stageOneImgRetGain}{1.5}
\newcommand{\stageOneImgClsGain}{0.1}

\newcommand{\vidDurActivityNet}{117}
\newcommand{\vidDurMsrvtt}{14}
\newcommand{\vidDurMsvd}{8}

\begin{abstract}
\noindent
Embedding images, videos and text in a shared space enables retrieval, curation and zero-shot classification
over large media collections.
Real collections are heterogeneous, with images and videos across a wide range of resolutions and descriptions
that can be long and multilingual.
Yet, no compact open model excels at both retrieval and classification across this full range.
We introduce Clipstral, a 1.2B open-weight dual encoder built for this setting.
The vision tower is a vision transformer that can process images at their native resolution and aspect ratio,
and extends to video through a lightweight spatio-temporal encoder that attends jointly to the patch tokens of
all frames; the text tower embeds text of any length in many languages.
Under one zero-shot protocol covering 37 image and video benchmarks at four resolutions, we compare Clipstral
with 13 open-weight encoders up to 4B parameters.
Clipstral is the strongest image and video retriever, leading on long-caption and multilingual retrieval, and
ranks second on image and video classification, behind models 1.7--2$\times$ larger.
Accuracy holds across all tested resolutions (224--1036\,px), and the model pushes the accuracy--size Pareto
frontier for image and video classification and retrieval.
We also introduce V$^*$-classification, a zero-shot benchmark of small details in native high-resolution
images.
On full images, every evaluated model scores within 10 points of chance on this benchmark, exposing
fine-detail recognition as an open problem.
The model weights, inference code and benchmark are released under an open license.
\end{abstract}

\begin{center}
\small
\renewcommand{\arraystretch}{1.25}%
\begin{tabular}{@{}c@{\hspace{0.5em}}l@{\hspace{0.8em}}l@{}}
\linkicon{\includegraphics[height=1.05em]{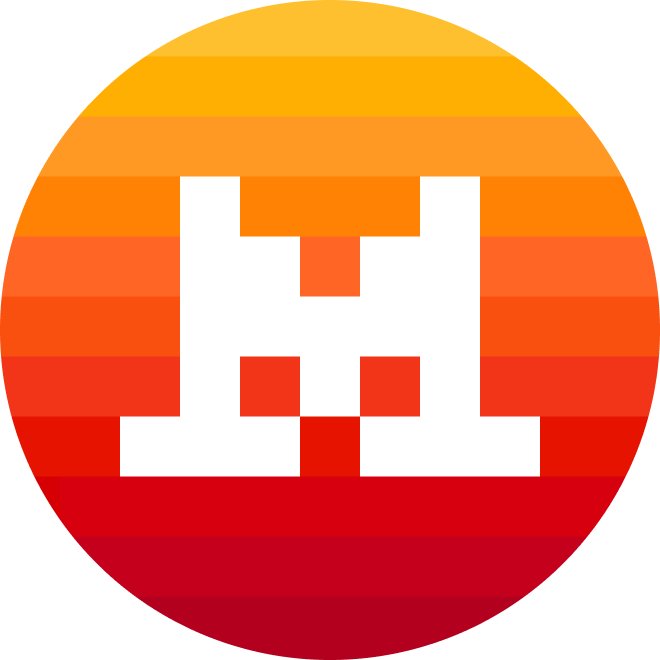}} & Blog post
  & \href{https://mistral.ai/news/mistral-multimodal-embed}{\texttt{mistral.ai/news/mistral-multimodal-embed}} \\
\linkicon{\includegraphics[height=1.05em]{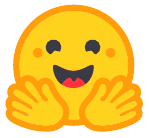}} & Model weights and inference code
  & \href{https://huggingface.co/mistralai/mistral-multimodal-embed-1B}%
  {\texttt{huggingface.co/mistralai/mistral-multimodal-embed-1B}} \\
\linkicon{\large\faGithub} & Examples and demo
  & \href{https://github.com/mistralai/mistral-multimodal-embed}%
  {\texttt{github.com/mistralai/mistral-multimodal-embed}} \\
\linkicon{\includegraphics[height=1.05em]{figures/logos/huggingface.pdf}} & \vstarcls{}
  & \href{https://huggingface.co/datasets/mistralai/vstar-classification}%
  {\texttt{huggingface.co/datasets/mistralai/vstar-classification}} \\
\end{tabular}
\end{center}

\section{Introduction}
\label{sec:intro}

\begin{figure}[!t]
\centering
\includegraphics[width=\linewidth]{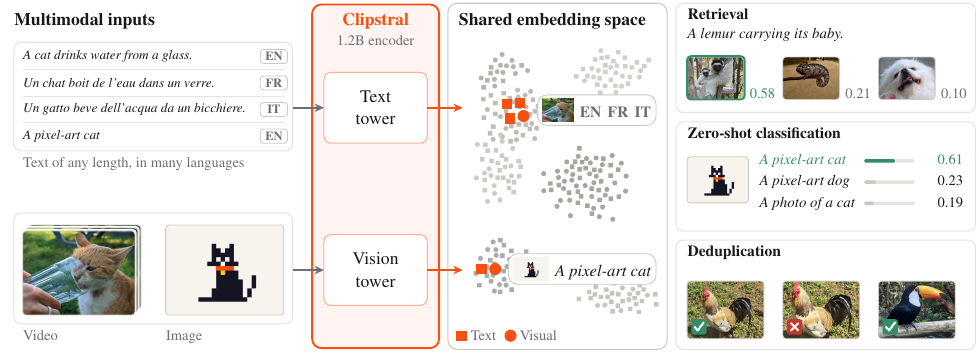}
\caption{\textbf{\ours{} overview.} The text and vision towers map text in many languages, images at any
resolution and aspect ratio, and videos into one shared embedding space, where matching text and visuals lie
close together.
On the right, use cases of \ours{}: text-to-image/video retrieval, zero-shot classification against class
prompts and the deduplication of image collections, where near-duplicate copies are discarded.
Retrieval and classification scores are illustrative cosine similarities.}
\label{fig:concept}
\end{figure}
Multimodal embedding models map text, images and videos into a shared space in which matching content lies
close together (\cref{fig:concept}).
This makes them the standard tool for retrieval over large media collections: a text query returns the nearest
images or videos, and vice versa.
The same space supports zero-shot classification, by comparing the embedding of an image or video with the
embeddings of class names, and the curation of large datasets.
These models are trained on large collections of image-text and video-text pairs with a contrastive
loss~\citep{radford2021clip,jia2021align,zhai2023siglip}, which pulls matching pairs together and pushes
non-matching pairs apart.
Once trained, they handle new tasks without retraining, and can still be fine-tuned on domain-specific data.

\paragraph{Limitations of prior encoders.}
We categorize contrastive embedding models into two broad families: dual encoders and VLM-based embedders.
Dual encoders embed text and visual inputs with separate towers.
Perception Encoder (PE)~\citep{bolya2025perception} supports images and video, but its L variant loses
substantial accuracy at high resolution (\cref{sec:results_resolution}), its text context is limited to 32
tokens (L) or 72 tokens (G) (\cref{sec:long_captions}), and it is English-centric
(\cref{sec:multilinguality}).
SigLIP2~\citep{tschannen2025siglip2}, TIPSv2~\citep{cao2026tipsv2}, C-RADIOv4~\citep{ranzinger2026cradiov4}
and MetaCLIP~2~\citep{chuang2025metaclip2} do not support video natively and cap text at 64--77 tokens.
TIPSv2 retrieves well in English, but this does not carry over to other languages
(\cref{sec:multilinguality}).
VLM-based embedders adapt VLM backbones with contrastive
training~\citep{jiang2025vlm2vec,zhang2025gme,meng2026vlm2vecv2}, so that a single backbone encodes text,
visual and interleaved inputs.
Qwen3-VL-Embedding~\citep{li2026qwen3vlembedding} comes in 2B and 8B variants, and
EmbeddingGemma~2~\citep{google2026embeddinggemma2} has 0.44B parameters excluding its audio encoder.
These models are competitive on retrieval but trail the strongest dual encoders on image classification
(\cref{sec:results_image}).
Hosted embedders such as Gemini Embedding~2~\citep{shanbhogue2026geminiembedding2}, Voyage
multimodal-3.5~\citep{voyageai2026multimodal} and Cohere Embed~\citep{cohere2026embed} are available only
through APIs, so we focus on open-weight models, which can be deployed on-premise, fine-tuned and compared
under one controlled protocol.

Overall, each of these open-weight models meets only some of the following requirements: strong retrieval and
classification from low to high resolution for (i) images and (ii) videos, (iii) text of arbitrary length,
(iv) native multilingual support, and (v) a compact size for indexing large collections.

\paragraph{Limitations of prior benchmarks.}

Existing benchmarks typically evaluate models at a single input resolution and therefore miss failure modes
that only appear at higher resolutions.
Our protocol instead measures retrieval and classification performance while varying the input size, resizing
every image and video to several resolutions (\cref{sec:eval}).
This exposes models that degrade on high-resolution inputs, i.e., on long patch sequences, as we observe for
PE-Core-L~\citep{bolya2025perception}.
However, resizing does not test whether a model can recognize a detail spanning a few pixels in a native
multi-megapixel image.
High-resolution benchmarks exist for VLM question answering and
retrieval~\citep{wu2024vstar,wang2025hrbench,zhang2025mmerealworld,liu2025sorce}, but, to the best of our
knowledge, none evaluates the zero-shot classification of small details in native high-resolution images.

\begin{figure}[!t]
\centering
\includegraphics[width=0.9\linewidth]{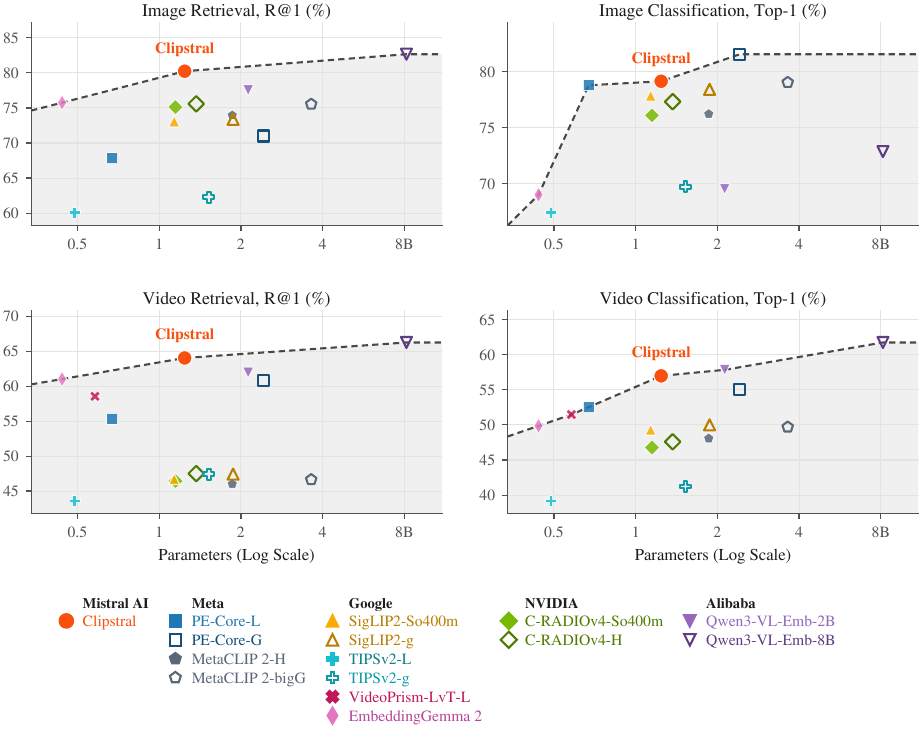}
\caption{\textbf{Performance vs.\ model size.}
We include open-weight models from Mistral, Meta, Google, NVIDIA and Alibaba.
The dashed line traces the Pareto frontier of performance vs.\ size.
No model smaller than \ours{} scores higher on image and video retrieval and zero-shot classification.
Among larger models, only Qwen3-VL-Embedding (8B in retrieval, 2B and 8B in video classification) and
PE-Core-G (image classification) score higher.
For each model we report best-over-resolutions performance average over several benchmarks.
For protocol and benchmark details, see \cref{sec:protocol,sec:benchmarks} and \cref{app:eval}.}
\label{fig:larger_models}
\end{figure}

\begin{figure}[t]
\centering
\includegraphics[width=\linewidth]{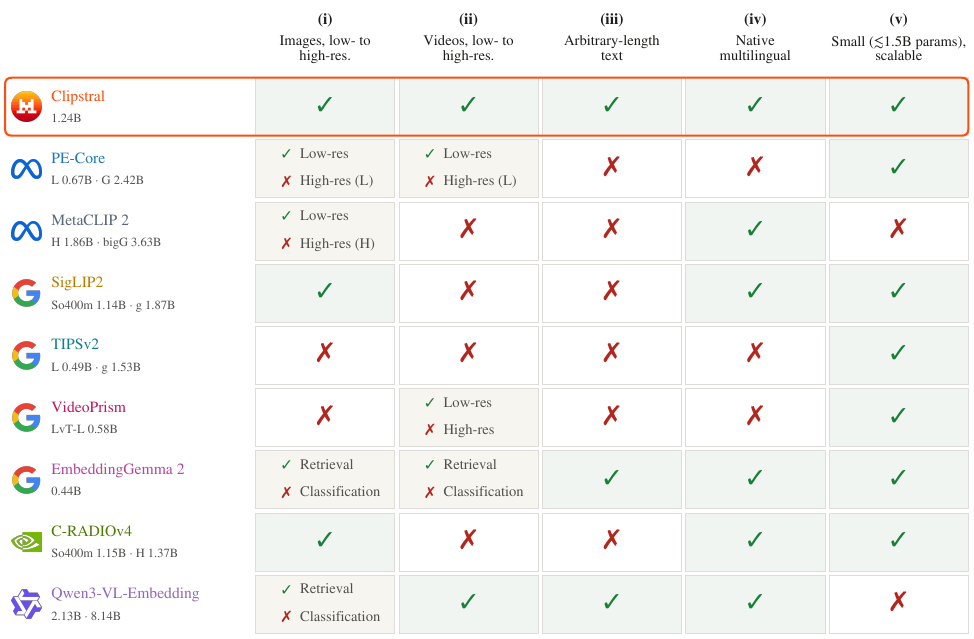}
\caption{\textbf{Requirements vs.\ model suite.} Requirements (i)--(v) identified in \cref{sec:intro} for
\ours{} and every model suite of our main comparison (one row per suite, with its evaluated sizes).
A suite meets (v) if one of its variants does; ``(L)'' and ``(H)'' mark the variant that fails at high
resolution (224 vs.\ 1036\,px group means, \cref{sec:results_resolution}).
For protocol and benchmark details, see \cref{sec:protocol,sec:benchmarks} and \cref{app:eval}.}
\label{fig:properties}
\end{figure}

\paragraph{\ours{}.}
We introduce \ours{}, a 1.2B open-weight dual encoder that meets requirements (i)--(v)
(\cref{fig:properties}).
The vision tower is a Vision Transformer (ViT,~\citealp{dosovitskiy2021vit}) that can process images at their
native resolution and aspect ratio, and extends to video through a lightweight spatio-temporal (ST) encoder
that attends jointly to the patch tokens of all frames.
The text tower embeds multilingual text of any length (\cref{sec:method}).
We compare \ours{} with 13 open-weight encoders up to 4B parameters from Meta, Google, NVIDIA and Alibaba,
plus Qwen3-VL-Embedding-8B as a reference, under a unified zero-shot protocol covering 18 image
classification, 6 image retrieval, 8 video classification and 5 video retrieval benchmarks of natural images
and videos, at four input resolutions (\cref{sec:eval}), and under a standardized throughput protocol
(\cref{sec:throughput}).
Among open-weight models up to 4B parameters, \ours{} is the strongest image and video retriever, leads on
long-caption and multilingual retrieval, and ranks second on image and video classification, behind models
1.7--2$\times$ larger (\cref{fig:larger_models}).
Accuracy holds across all four tested resolutions (224--1036\,px; \cref{sec:results_resolution}), and the
model pushes the accuracy--latency Pareto frontier in image and video retrieval and classification
(\cref{fig:acc_latency}).

\paragraph{\vstarcls{}.}
To evaluate fine detail at native resolution, we introduce \vstarcls{}, to our knowledge the first zero-shot
classification benchmark of small details in native high-resolution images, derived from
V$^*$~Bench~\citep{wu2024vstar} (\cref{sec:vstar}).
On full images, every evaluated model scores within 10 points of chance, so fine-detail recognition remains
open (\cref{sec:results_vstar}).
We release the weights and inference code of \ours{} together with the benchmark.

\section{Related work}
\label{sec:competitors}

We review existing vision encoders, grouped into contrastive models (\cref{sec:related_contrastive}) and other
vision encoders (\cref{sec:related_other}).

\subsection{Contrastive models}
\label{sec:related_contrastive}

\paragraph{Dual encoders.}
CLIP and ALIGN~\citep{radford2021clip,jia2021align} established separate vision and text towers trained on
paired examples.
Later work improved the data through open corpora, metadata balancing and
filtering~\citep{schuhmann2022laion5b,cherti2023openclip,xu2024metaclip,chuang2025metaclip2,fang2024dfn}, the
contrastive objective, including sigmoid, captioning, self-distillation and masked-prediction
losses~\citep{zhai2023siglip,yu2022coca,tschannen2025siglip2,maninis2025tips,cao2026tipsv2,tang2025tulip}, and
the recipe at scale~\citep{zhai2022lit,sun2023evaclip,sun2024evaclip18b,li2025openvision,bolya2025perception}.
Different directions addressed the requirements of \cref{sec:intro} separately.
For (i) images across resolutions, sequence packing~\citep{dehghani2023navit} and randomized patch
sizes~\citep{beyer2023flexivit} let a ViT accept arbitrary input sizes, a contrastive encoder can be trained
at native resolution~\citep{qiao2025univitar}, and multi-resolution training or feature distillation reduce
the accuracy drop at unseen resolutions~\citep{heinrich2025radiov25,ranzinger2025featsharp,yang2026sigliphd};
\ours{} trains contrastively on a mixture of resolutions.
For (ii) video, space--time transformers factorize or jointly apply attention over frames and
patches~\citep{bertasius2021timesformer,arnab2021vivit}, while CLIP-based video models mostly pool frame
features or are pretrained on video--text
data~\citep{luo2022clip4clip,ma2022xclip,yan2022videococa,wang2022internvideo,li2023umt,wang2024internvideo2,%
zhao2024videoprism,wang2024videoclipxl}; \ours{} adds joint space--time attention over the patch tokens of all
frames to an image encoder.
For (iii) long text, variants stretch~\citep{zhang2024longclip} or replace~\citep{najdenkoska2025tulipclip}
positional embeddings, train on long re-captions~\citep{zheng2024dreamlip,xie2025fgclip}, or replace the text
tower with an LLM~\citep{huang2026llm2clip}, although long-caption training can degrade short-text
tasks~\citep{wu2024lotlip}.
For (iv) multilingual text, variants distill~\citep{carlsson2022mclip,chen2023altclip} or
replace~\citep{visheratin2023nllbclip} the text tower, or train on worldwide
data~\citep{chuang2025metaclip2,xie2026fgclip2}.
These models address subsets of requirements (i)--(v); \ours{} combines all five in one dual encoder
(\cref{fig:properties}).

\paragraph{VLM embedders.}
Instruction-conditioned VLMs can be adapted into embedding models with contrastive
training~\citep{jiang2024e5v,jiang2025vlm2vec,zhang2025gme,liu2025lamra,lin2025mmembed,gu2025unime,%
chen2025mme5,meng2026vlm2vecv2,faysse2025colpali}.
We evaluate the 2B and 8B Qwen3-VL-Embedding variants~\citep{li2026qwen3vlembedding} and
EmbeddingGemma~2~\citep{google2026embeddinggemma2}.
The smallest Qwen variant (2B) has \qwenParamRatio{}$\times$ as many parameters as \ours{}.
Nevertheless, \ours{}, which uses separate vision and text towers, leads or is competitive with open-weight
models of up to 4B parameters on natural-image and video retrieval and classification.
Its advantage over Qwen3-VL-Embedding-2B and EmbeddingGemma~2 is largest on image classification
(\cref{fig:larger_models}).

\paragraph{Closed models.}
Hosted embedders include Gemini Embedding~2~\citep{shanbhogue2026geminiembedding2}, Voyage
multimodal-3.5~\citep{voyageai2026multimodal}, and Cohere Embed v3 and v4~\citep{cohere2026embed}.
We do not include them in our quantitative comparison, since we focus on open-weight models.

\subsection{Other vision encoders}
\label{sec:related_other}

\paragraph{Self-supervised vision encoders.}
\texttt{dino.txt}~\citep{jose2024dinotxt} aligns a text encoder to a vision backbone trained with
self-supervision on images alone: DINOv2~\citep{oquab2024dinov2} in the original work and
DINOv3~\citep{simeoni2025dinov3}, which applies the same recipe to its own backbone.
Both text-aligned variants are competitive on zero-shot image classification and strong on open-vocabulary
segmentation, but their image--text retrieval trails that of SigLIP variants. \ours{}, in contrast,
outperforms SigLIP2 on image--text retrieval and also supports retrieval and classification of videos
(\cref{sec:method_arch}).
For video, V-JEPA~2~\citep{assran2025vjepa2} predicts masked video representations without an image--text
contrastive loss.
Its released encoder has no text tower, so it cannot embed text queries.
VL-JEPA~\citep{chen2026vljepa} adds a text encoder and a predictor and reports results on zero-shot video
classification and text-to-video retrieval.
It is not open source, so it falls outside the scope of our comparison.

\paragraph{Distilled vision encoders.}
Agglomerative models distill several foundation-model teachers into a single student backbone.
They are strong general-purpose vision backbones, and some support text--image tasks by mapping their features
into a teacher's embedding space.
AM-RADIO and RADIOv2.5~\citep{ranzinger2024amradio,heinrich2025radiov25} distill image teachers and use
CLIP-aligned image adapters, while C-RADIOv4~\citep{ranzinger2026cradiov4} uses a SigLIP2-g-aligned adapter.
In each case, text embeddings come from the separately loaded teacher text encoder.
All three are image backbones.
For video, our C-RADIOv4 baseline averages frame embeddings (\cref{app:baseline_protocol}).
EUPE~\citep{zhu2026eupe} distills vision teachers through a large proxy model into efficient image backbones.
Its released checkpoints are vision-only.
The reported zero-shot classification requires a teacher text tower and an adapter, so the checkpoints alone
support neither text-query classification nor retrieval. \ours{} trains a separate text tower together with
the vision tower on image--text and video--text pairs, yielding one contrastive embedding space for both
media.

\section{Method}
\label{sec:method}

We first describe the architecture of \ours{} (\cref{sec:method_arch}), then its training
(\cref{sec:method_training}): the model is trained on image--text pairs first (\cref{sec:method_image}),
followed by a training stage on videos (\cref{sec:method_video}).

\begin{figure}[t]
\centering
\includegraphics[width=\linewidth]{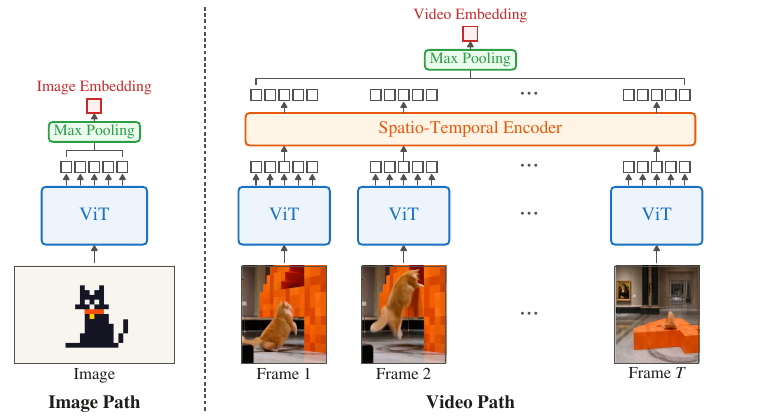}
\caption{\textbf{Vision paths of \ours{}.} Left: an image is encoded by the ViT and its patch tokens are
max-pooled.
Right: $T$ video frames are encoded separately by the ViT, and their tokens are then processed jointly by the
spatio-temporal encoder and finally max-pooled.
A linear head maps the pooled token to the shared 1024-d embedding space, into which the causal text tower
(not shown) embeds text from its final token.
All final image, video and text embeddings are L2-normalized.}
\label{fig:arch}
\end{figure}

\subsection{Architecture}
\label{sec:method_arch}

\ours{} is a dual-encoder vision-language model trained with a CLIP-style contrastive objective: a vision
tower processes images and videos, while a text tower encodes the associated text.
It has \oursParams{} parameters in total (\cref{tab:model_architecture}).

\paragraph{Vision tower.}
The vision encoder is a Vision Transformer (ViT,~\citealp{dosovitskiy2021vit}) with \oursVisParams{}
parameters~(\mbox{\cref{tab:model_architecture}}), patch size 14, a $2{\times}2$ patch merger and 2D rotary
position embeddings~\citep{su2024roformer}, which let it process any image size and aspect ratio.
The embedding of an $H\times W$ image, with $H$ and $W$ multiples of 28, is the element-wise maximum over the
$N=HW/28^{2}$ merged tokens (max pooling), followed by a linear head into the shared 1024-d embedding space.
The vision tower is initialized from an internal vision encoder~\citep{mistral2026large4}, which had not been
trained contrastively.

\paragraph{Video path.}
A video clip is represented by $T$ sampled frames, each encoded separately by the ViT. The resulting $N T$
tokens are concatenated and processed by a spatio-temporal (ST) encoder: a 4-layer transformer with
\oursSTParams{} parameters~(\mbox{\cref{tab:model_architecture}}), full attention~\citep{arnab2021vivit} and
3D rotary embeddings~\citep{wang2024qwen2vl,wei2025videorope}.
The ST encoder lets the model learn spatio-temporal relationships between patches across all frames.
The video embedding is the max pooling over all output tokens, followed by the image head.
In \cref{sec:abl_video_pool}, we ablate the ST encoder and the pooling operator, and find that the ST encoder
is most effective when applied to the patch tokens of all frames and that max pooling is the best pooling
operator in our setting.

\paragraph{Relation to prior video architectures.}
X-CLIP~\citep{ma2022xclip} passes one ViT class token per frame through a three-layer temporal transformer.
Our ST encoder instead receives the merged patch tokens from every frame, which better represent spatial
information.
InternVideo2~\citep{wang2024internvideo2} applies joint space--time attention throughout its video ViT,
whereas \ours{} confines it to four layers after the per-frame ViT, which reduces the computational cost per
video.
Both Frozen in Time~\citep{bain2021frozen} and VideoPrism~\citep{zhao2024videoprism} factorize space--time
attention, but arrange it differently.
Frozen in Time applies temporal and then spatial attention within each transformer block.
VideoPrism, like \ours{}, first encodes frames spatially and then passes their patch tokens to a separate
encoder, but its temporal attention operates only between corresponding patch locations across frames.
Our ST encoder instead attends jointly to all patch tokens, allowing any location in one frame to interact
with any location in another.

\paragraph{Text tower.}
The text encoder is a transformer~\citep{vaswani2017attention} with \oursTxtParams{}
parameters~(\mbox{\cref{tab:model_architecture}}), full attention with a causal mask, rotary positions, a
131k-entry multilingual tokenizer and no fixed context limit.
The text embedding is the final hidden state of the end-of-sequence token, followed by a linear head
projecting from the internal dimension to the shared 1024-d embedding space.
The text tower is initialized from an internal Mistral checkpoint~\citep{mistral2026large4} that had not been
trained contrastively.

\begin{table}[tbp]
\centering
\caption{\textbf{Architecture of the vision and text towers.} The ViT is shared by images and video frames; a $2{\times}2$ patch merge precedes the video-only ST encoder. Both towers produce 1024-dimensional embeddings.}
\label{tab:model_architecture}
\small
\setlength{\tabcolsep}{5pt}
\begin{tabular}{@{}lccc@{}}
\toprule
 & \multicolumn{2}{c}{Vision tower} & Text tower \\
\cmidrule(lr){2-3}\cmidrule(l){4-4}
 & ViT-L/14 & ST encoder & Causal transformer \\
\midrule
Parameters & \oursVisParams{} & \oursSTParams{} & \oursTxtParams{} \\
Hidden width & 1024 & 1024 & 1792 \\
Layers & 24 & 4 & 14 \\
Attention heads & 8 & 8 & 14 \\
Feed-forward width & 4096 & 4096 & 4864 \\
Position encoding & 2D RoPE & 3D RoPE & 1D RoPE \\
\bottomrule
\end{tabular}
\end{table}

\subsection{Training}
\label{sec:method_training}

Both training stages use the same objective.
Given a global batch of $B$ (image, text) or (video, text) pairs, we train the model to minimize the symmetric
InfoNCE loss~\citep{oord2018cpc} used in CLIP~\citep{radford2021clip}
\begin{equation}
\mathcal{L} = -\frac{1}{2B}\sum_{i=1}^{B}\left[\log\frac{e^{\gamma\, \vv_i^\top \vt_i}}{\sum_{j} e^{\gamma\,
\vv_i^\top \vt_j}} + \log\frac{e^{\gamma\, \vv_i^\top \vt_i}}{\sum_{j} e^{\gamma\, \vv_j^\top \vt_i}}\right],
\label{eq:infonce}
\end{equation}
where $\vv_i$ and $\vt_i$ are the L2-normalized embeddings of the $i$-th visual input and its text, and
$\gamma$ is a learned inverse temperature (logit scale).

\subsubsection{Stage 1: image--text training}
\label{sec:method_image}

\begin{figure}[t]
\centering
\includegraphics[width=\linewidth]{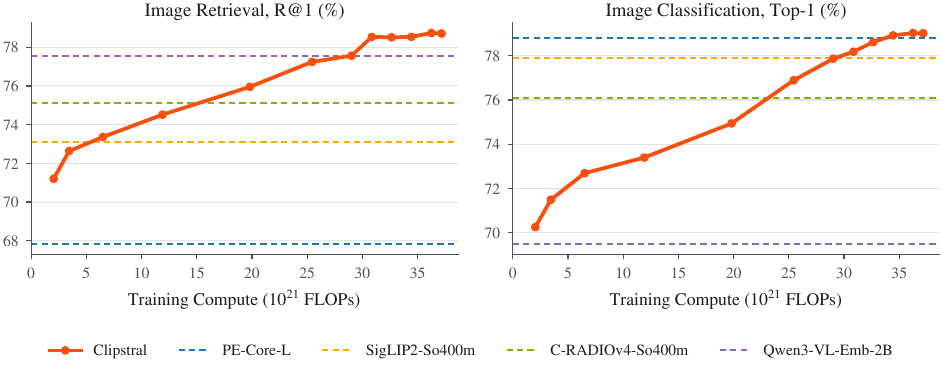}
\caption{\textbf{Stage~1 image performance vs.\ training compute.}
Mean R@1 over the image retrieval sets of \cref{tab:eval_image_ret} (both directions) and mean top-1 accuracy
over the classification sets of \cref{tab:eval_image_cls}, for checkpoints of the Stage~1 run, with the same
protocol (best over resolutions and transforms); dashed lines: baselines from the same tables.
Stage~1 ran for 6.1 days on 512 H100 GPUs (${\approx}$75k H100-hours).}
\label{fig:stage1_progress}
\end{figure}

The image training mix draws from a deduplicated pool of ${\sim}$7.6B image--text pairs from different
sources.
Stage 1 trains the vision and text towers (without the ST encoder) for ${\sim}$5.2 trillion tokens (370k
steps).
We use a global batch size of ${\sim}$14.2M tokens (counting both image and text tokens) over 512 H100 GPUs
with data parallelism (${\sim}$150k image--text pairs per step).
Stage~1 takes 6.1 days (${\approx}$75k H100-hours); \cref{fig:stage1_progress} shows image retrieval and
classification over the course of training.

We use AdamW ($\beta_1=0.9$, $\beta_2=0.95$, $w_{\text{decay}}=0.1$) with a 2k-step warm-up to a peak learning
rate of $6{\times}10^{-4}$, followed by a cosine schedule whose last 5\% of steps are a linear decay.
We clip gradients at norm 5.0 and use a token dropout of 0.2.

\paragraph{Spatial transformations.}
Each training image is first randomly cropped (area fraction in $[0.5, 1]$) as in \citet{bolya2025perception},
keeping its original aspect ratio.
We then sample a target resolution $L$, which is $L_{\text{low}}$ with probability 0.9 and $L_{\text{high}}$
with probability 0.1.
If the longer side exceeds $L$, we downscale the image so that its longer side equals $L$ (keeping the aspect
ratio).
Otherwise, the image keeps its original size:
\begin{equation}
L = \begin{cases} L_{\text{low}} & \text{with probability } 0.9,\\ L_{\text{high}} & \text{with probability }
0.1;\end{cases}
\qquad (h, w) \mapsto \frac{(h, w)}{\max\!\left(1, \max(h, w)/L\right)} .
\end{equation}
In both cases, both sides are then rounded up to multiples of 28\,px.
For this stage, we chose $L_{\text{low}}=224$\,px and $L_{\text{high}}=1036$\,px.
The model thus sees a continuum of sizes and aspect ratios throughout training, unlike other approaches that
increase the resolution during training~\citep{bolya2025perception,tschannen2025siglip2}.
This keeps the model from specializing in a specific resolution, so its performance stays consistent across a
wide range of input resolutions, as shown in \cref{sec:results_resolution}.

\subsubsection{Stage 2: video--text training}
\label{sec:method_video}

After the image-only stage, we further train the model to support video inputs.
In this stage, we use two data mixtures: a mixture of 48M video clips collected from public sources across
different domains (open-source mixture), and a high-quality mixture of 8.8M video clips that we curated
(curated mixture).
The curation includes cutting long videos into clips of at most 1 minute.
From each clip, we sample $T=16$ uniformly spaced frames, resized as in Stage~1 but without random cropping,
with one target size per video ($L_{\text{low}}=448$\,px and $L_{\text{high}}=1036$\,px).
The choice of $T$ determines the temporal resolution at which the model sees the video.
We ablated $T$ while keeping the total token count constant.
More frames reduce the number of (video, text) pairs within the batch, which lowers retrieval performance,
whereas fewer frames lower video classification performance, as expected from the lower temporal resolution
(\cref{app:frames}, \cref{fig:app_frames}).
We also experimented with content-aware frame sampling, but for our short videos it adds significant
complexity without measurable gains over uniform sampling (\cref{fig:app_smart}).

\paragraph{ST encoder alignment and full fine-tuning.}
Starting from the Stage~1 image-only checkpoint, we add a randomly initialized ST encoder and train in three
phases:
\begin{enumerate}[leftmargin=*,itemsep=1pt,topsep=2pt]
\item \emph{ST encoder alignment on the open-source mixture} (${\approx}$1 epoch), peak learning rate
$10^{-4}$.
\item \emph{ST encoder alignment on the curated mixture} (${\approx}$1 epoch), peak learning rate $10^{-4}$.
\item \emph{Full fine-tuning on the curated mixture} (${\approx}$9 epochs), with 10\% of batches replayed from
the Stage~1 image--text mixture and a peak learning rate of $8.4{\times}10^{-7}$, chosen from scaling laws
extracted from preliminary runs.
\end{enumerate}

In \cref{sec:abl_phases}, we ablate the data mixture used in the full fine-tuning phase and find that the
curated mixture brings clear video gains over the open-source mixture.
For all three phases, we use a global batch size of 24.6M tokens (counting both frame and text tokens) over
256 ranks with data parallelism (${\approx}$7k video--text pairs per step).
The three phases last 8{,}868, 1{,}325 and 13{,}248 steps, respectively.
We use AdamW ($\beta_1=0.9$, $\beta_2=0.95$, $w_{\text{decay}}=0.1$) with a learning-rate schedule restarted
at each phase: a linear warm-up (2{,}000 steps in the first phase, 500 in the others), a cosine decay, and a
linear
decay to 0.01 of the peak learning rate over the last 5\% of steps.
We clip gradients at norm 5.0 and use a token dropout of 0.2.
Stage~2 takes ${\approx}$4.4k H200-hours in total.

\section{Evaluations}
\label{sec:eval}

We evaluate \ours{} on \emph{zero-shot} image and video retrieval and classification under one protocol, and
compare it with 13 open-weight models up to 4B parameters, plus Qwen3-VL-Embedding-8B as a reference.
We describe the evaluation protocol and baselines in \cref{sec:protocol} and benchmarks in
\cref{sec:benchmarks}, where we also introduce \vstarcls{}, our new benchmark for small-detail classification
at high resolution.
We report image and video retrieval and classification, each averaged over several benchmarks, in
\cref{sec:results_image,sec:results_video} (individual scores in
\crefrange{tab:eval_image_ret}{tab:eval_video_cls} and \cref{app:perres}).
We then focus on long-caption and multilingual retrieval in \cref{sec:long_captions,sec:multilinguality}, on
robustness to input resolution in \cref{sec:results_resolution}, and on \vstarcls{} in
\cref{sec:results_vstar}.
We report performance versus inference throughput in \cref{sec:throughput}.
Finally, we ablate how to turn the image-only embedder into a video embedder: the video architecture in
\cref{sec:abl_video_pool} and the video training data in \cref{sec:abl_phases}.

\subsection{Evaluation protocol}
\label{sec:protocol}

\paragraph{Baselines.}
We compare \ours{} (1.24B parameters) with 13 open-weight embedding models up to 4B parameters from Meta,
Google, NVIDIA and Alibaba: PE-Core-L and -G~\citep{bolya2025perception}; SigLIP2-So400m in its
native-aspect-ratio NaFlex variant and SigLIP2-g~\citep{tschannen2025siglip2}; C-RADIOv4-So400m and
-H~\citep{ranzinger2026cradiov4}; TIPSv2-L and -g~\citep{cao2026tipsv2}; MetaCLIP~2-H and
-bigG~\citep{chuang2025metaclip2}; VideoPrism-LvT-L~\citep{zhao2024videoprism} for video; and the VLM
embedders EmbeddingGemma~2~\citep{google2026embeddinggemma2} and
Qwen3-VL-Embedding-2B~\citep{li2026qwen3vlembedding}.
We also show Qwen3-VL-Embedding-8B (8.14B parameters) as a reference for much larger models.
Additional baselines are in \cref{app:jina_eva}.
EmbeddingGemma~2 has 0.74B parameters in total, including a 0.3B audio encoder that we do not use, so we
report its text and vision parameters (0.44B) throughout.
Each model uses its own tokenizer and native text context, which is 32--77 tokens for the dual-encoder
baselines. \ours{} and the VLM embedders support longer contexts, as they do not truncate the text in any
evaluated setting (\cref{tab:truncation}).

\paragraph{Spatial transformations.}
Every model is evaluated at 224, 336, 672 and 1036\,px.
Where a model's patch size does not divide the target resolution, we round the image side to the nearest
multiple of the patch size (e.g.\ 1036$\to$1024\,px for C-RADIOv4), and each model resizes its position
embeddings with its official implementation (details in \cref{app:baseline_protocol}).
\crefrange{tab:eval_image_ret}{tab:eval_video_cls} report the best-over-resolution score for each dataset and
retrieval direction. \cref{sec:results_resolution} shows the averages for image/video classification and
retrieval at each resolution, and \cref{app:perres} gives the per-benchmark breakdown.

For each model, dataset and resolution, we take the best score over three image transforms.
For videos, we apply the same transform to every frame.
We extend the evaluation protocol of \citet{bolya2025perception} by considering the following transforms:
\begin{itemize}[leftmargin=*,itemsep=1pt,topsep=2pt]
\item \texttt{squash}: resize to a target-size square, changing the aspect ratio.
\item \texttt{center\_crop}: resize the shorter side to the target while preserving the aspect ratio, then
take a target-size square center crop.
\item \texttt{keep\_ratio}: resize the longer side to the target while preserving the aspect ratio.
\end{itemize}

\paragraph{Video sampling and embeddings.}
We sample 16 uniformly spaced frames from each test video. \ours{}, VideoPrism-LvT-L, Qwen3-VL-Embedding and
EmbeddingGemma~2 use their video-capable encoders.
For PE-Core, which was trained on video with this pooling~\citep{bolya2025perception}, and for the image-only
SigLIP2, C-RADIOv4, TIPSv2 and MetaCLIP~2, we average L2-normalized frame embeddings.

\paragraph{Metrics.}
A retrieval task consists of a search query and a collection of candidates.
We embed all samples and rank the candidates by cosine similarity to the query.
We report Recall@1 (R@1), the fraction of queries whose top-ranked item is correct, for text-to-image/video
and image/video-to-text.
If an item has multiple correct captions, the retrieval counts as correct if any of them is ranked first.
In model comparisons, we do not use the query-set re-ranking of \citet{bolya2025perception} (\cref{app:dsl}):
although it boosts performance, it is unrealistic, as it re-ranks the results of each query based on all other
queries.
For all classification tasks, we report top-1 accuracy.
The only exception is PA-100K~\citep{liu2017hydraplus}, for which we report mean Average Precision (mAP) over
26 attributes: we compute the average precision of each attribute over all test images and then take the mean
over attributes.

\paragraph{Instruction prompting.}
VLM-based embedders can optionally take an instruction prompt alongside the image, video or text input.
We evaluate Qwen3-VL-Embedding with either its default instruction, ``Represent the user's input.'' (Neutral),
or task-specific instructions (Tuned).
For tuned runs, we use published MMEB/MMEB-V2 instructions where
available~\citep{jiang2025vlm2vec,meng2026vlm2vecv2,li2026qwen3vlembedding} and craft instructions in the same
style otherwise (\cref{app:qwen_prompts}).
EmbeddingGemma~2 uses fixed task prefixes rather than free-form instructions.
Its Neutral and Tuned settings are detailed in \cref{app:qwen_prompts}.

On average, tuned instructions improve over neutral ones (\cref{tab:instruction_variants}).
\crefrange{tab:eval_image_ret}{tab:eval_video_cls} and the aggregate plots
(\cref{fig:larger_models,fig:resolution,fig:acc_latency}) report results with the tuned settings of both
Qwen3-VL-Embedding and EmbeddingGemma~2; the neutral settings are compared in \cref{app:qwen_prompts}.
On \vstarcls{} (\cref{tab:vstar_main}), Qwen uses a \vstarcls{}-specific instruction
(\cref{app:qwen_prompts}).

\subsection{Benchmark tasks}
\label{sec:benchmarks}

\paragraph{Image retrieval.}
We evaluate on COCO~\citep{lin2014coco,chen2015cococaptions}, Flickr30k~\citep{young2014flickr30k},
TextCaps~\citep{sidorov2020textcaps}, DOCCI~\citep{onoe2024docci}, XM3600~\citep{thapliyal2022xm3600} (36
languages), and VisualNews~\citep{liu2021visualnews} in its MMEB evaluation version.
In that version, each query is ranked within its own 1{,}000-candidate gallery~\citep{jiang2025vlm2vec}.

\paragraph{Image classification.}
We evaluate on 18 datasets grouped into natural images and fine-grained recognition, using CLIP prompt
ensembles~\citep{radford2021clip} (templates in \cref{app:templates}). \cref{app:benchmarks} lists and cites
every dataset.

\paragraph{Video retrieval.}
We evaluate on MSR-VTT~\citep{xu2016msrvtt}, MSVD~\citep{chen2011msvd}, ActivityNet
Captions~\citep{krishna2017activitynet}, PVD-Bench~\citep{bolya2025perception}, and
DiDeMo~\citep{hendricks2017didemo}.
For MSR-VTT, we use the original 2{,}990-video test split rather than the 1k-A split used in the released PE
evaluation~\citep{bolya2025perception}, as the 1k-A test set overlaps with the 1k-A training partition
(\cref{app:retrieval_protocol}).

\paragraph{Video classification.}
We evaluate on Kinetics-400/600/700~\citep{kay2017kinetics}, UCF101~\citep{soomro2012ucf101},
HMDB51~\citep{kuehne2011hmdb}, Something-Something~v2 (SSv2)~\citep{goyal2017ssv2},
EPIC-Kitchens-100~\citep{damen2022epic}, and Breakfast~\citep{kuehne2014breakfast}.

\begin{figure}[t]
\centering
\includegraphics[width=\linewidth]{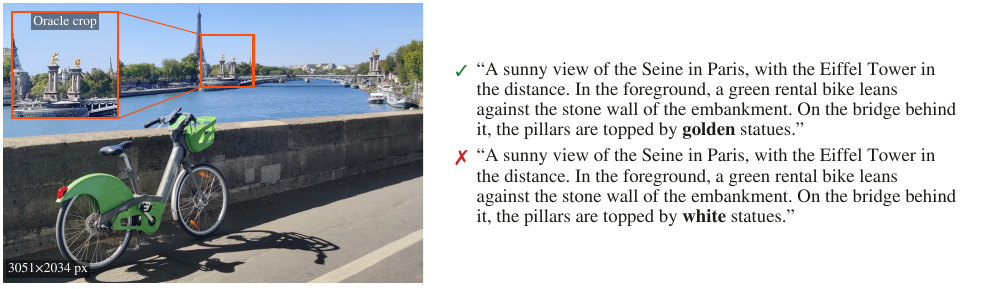}
\caption{\textbf{An illustration of a typical \vstarcls{} item:} the full native image with the oracle crop
(orange box: the target region at native pixel density), the positive caption (\cmark) and a hard-negative
caption (\xmark), which differ only in the answer phrase (bold).}
\label{fig:vstar_example}
\end{figure}
\paragraph{\vstarcls{}.}
\label{sec:vstar}

The benchmarks above can be run at high resolution, but they may not require it: their labels refer to the
predominant foreground content of each image.
They therefore do not test whether an encoder can resolve a small detail that is only visible at native
resolution.
To test this, we build \vstarcls{}, a zero-shot image classification benchmark in which the correct caption
depends on the attributes of a small region in a large image.
It reuses the 191 images of the multiple-choice question-answering benchmark V$^*$Bench~\citep{wu2024vstar},
which are $2246{\times}1583$ pixels on average.

We convert each V$^*$Bench (question, answer) pair into a classification problem.
Each image is paired with one positive caption, which describes the overall scene and states the correct
visual detail in natural language, and 1--3 hard-negative captions, which keep the same scene description and
differ only in the answer-bearing detail (see \cref{fig:vstar_example} for an illustrative example).
Given the image, an embedding model selects the closest caption by cosine similarity.
Because all candidates describe the same scene, the task cannot be solved from broad image semantics alone:
the model must resolve the small annotated detail.

We evaluate each model on two variants of \vstarcls{}: the \emph{full native image}, resized only to fit the
model's patch grid, and an \emph{oracle crop} of the annotated target at native pixel density, which is an
easier version of the task.
Humans identify the correct detail: choosing among the same candidate captions, human annotators reach 98.6\%
accuracy on the full native image and 99.6\% on the oracle crop (\cref{tab:vstar_main}, \cref{app:vstar}).

To measure text-side bias, we also evaluate the models on a \emph{gray-image} version of the benchmark, in
which the image is replaced by a gray square (\cref{app:vstar}, \cref{fig:vstar_cls}).
This measures how well a model can solve the task from the caption text alone, without looking at the image.
All models score close to chance on the gray-image version of \vstarcls{}.

\subsection{Image results}
\label{sec:results_image}

\begin{table}[!b]
{\renewenvironment{table}[1][]{}{\par}\begin{table}[tbp]
\centering
\caption{\textbf{Image--text retrieval.} Best top-1 recall (R@1) over input resolutions and transforms. Bold:
  best, underlined: second best among models up to 4B parameters.}
\label{tab:eval_image_ret}
\footnotesize
\setlength{\tabcolsep}{2.5pt}
\begin{tabular}{@{}l r cc cc cc cc cc cc c@{}}
\toprule
 &  & \multicolumn{2}{c}{COCO} & \multicolumn{2}{c}{Flickr30k} & \multicolumn{2}{c}{TextCaps} &
  \multicolumn{2}{c}{DOCCI} & \multicolumn{2}{c}{XM3600} & \multicolumn{2}{c}{VisualNews} & \\
\cmidrule(lr){3-4}\cmidrule(lr){5-6}\cmidrule(lr){7-8}\cmidrule(lr){9-10}\cmidrule(lr){11-12}\cmidrule(lr){13-14}
Model & Params & t2i & i2t & t2i & i2t & t2i & i2t & t2i & i2t & t2i & i2t & t2i & i2t & Avg. \\
\midrule
\textbf{\textcolor{accentorange}{\ours{}}} & 1.24B & \cellcolor{heat!27}55.5 & \cellcolor{heat!40}75.0 &
  \cellcolor{heat!29}84.1 & \cellcolor{heat!33}95.7 & \cellcolor{heat!32}\underline{79.3} &
  \cellcolor{heat!40}\underline{93.3} & \cellcolor{heat!41}\underline{93.8} & \cellcolor{heat!42}91.6 &
  \cellcolor{heat!45}\textbf{56.6} & \cellcolor{heat!45}\textbf{62.7} & \cellcolor{heat!43}\underline{85.6} &
  \cellcolor{heat!45}\underline{89.5} & \cellcolor{heat!45}\textbf{80.2} \\
PE-Core-L & 0.67B & \cellcolor{heat!32}56.7 & \cellcolor{heat!42}\underline{75.6} &
  \cellcolor{heat!39}\underline{85.9} & \cellcolor{heat!36}96.0 & \cellcolor{heat!27}76.7 &
  \cellcolor{heat!35}91.8 & 71.3 & 68.7 & \cellcolor{heat!5}11.0 & \cellcolor{heat!7}14.7 &
  \cellcolor{heat!38}80.9 & \cellcolor{heat!39}85.0 & \cellcolor{heat!17}67.9 \\
PE-Core-G & 2.42B & \cellcolor{heat!34}57.3 & \cellcolor{heat!35}74.1 & \cellcolor{heat!45}\textbf{87.0} &
  \cellcolor{heat!45}\textbf{96.9} & \cellcolor{heat!27}76.7 & \cellcolor{heat!34}91.7 &
  \cellcolor{heat!25}84.9 & \cellcolor{heat!23}81.3 & \cellcolor{heat!7}12.9 & \cellcolor{heat!9}16.8 &
  \cellcolor{heat!42}84.8 & \cellcolor{heat!42}87.6 & \cellcolor{heat!24}71.0 \\
MetaCLIP~2-H & 1.86B & 49.1 & 66.7 & 78.9 & 92.0 & \cellcolor{heat!17}71.2 & \cellcolor{heat!20}87.1 &
  \cellcolor{heat!17}80.4 & \cellcolor{heat!17}78.1 & \cellcolor{heat!41}52.2 & \cellcolor{heat!41}57.4 &
  \cellcolor{heat!43}85.5 & \cellcolor{heat!44}88.7 & \cellcolor{heat!31}74.0 \\
MetaCLIP~2-bigG & 3.63B & \cellcolor{heat!7}50.8 & \cellcolor{heat!7}68.2 & \cellcolor{heat!4}79.6 &
  \cellcolor{heat!13}93.4 & \cellcolor{heat!22}73.8 & \cellcolor{heat!25}88.7 & \cellcolor{heat!19}81.8 &
  \cellcolor{heat!21}80.0 & \cellcolor{heat!43}53.9 & \cellcolor{heat!42}\underline{59.3} &
  \cellcolor{heat!45}\textbf{87.1} & \cellcolor{heat!45}\textbf{89.9} & \cellcolor{heat!35}75.5 \\
SigLIP2-So400m & 1.14B & \cellcolor{heat!29}56.1 & \cellcolor{heat!24}71.7 & \cellcolor{heat!33}84.8 &
  \cellcolor{heat!31}95.4 & \cellcolor{heat!28}76.9 & \cellcolor{heat!32}90.9 & \cellcolor{heat!5}73.9 & 68.5
  & \cellcolor{heat!37}47.9 & \cellcolor{heat!36}50.8 & \cellcolor{heat!35}78.7 & \cellcolor{heat!35}81.6 &
  \cellcolor{heat!29}73.1 \\
SigLIP2-g & 1.87B & \cellcolor{heat!29}56.1 & \cellcolor{heat!23}71.6 & \cellcolor{heat!27}83.8 &
  \cellcolor{heat!23}94.6 & \cellcolor{heat!25}75.4 & \cellcolor{heat!27}89.4 & \cellcolor{heat!8}75.9 &
  \cellcolor{heat!1}69.0 & \cellcolor{heat!38}49.0 & \cellcolor{heat!35}50.6 & \cellcolor{heat!38}80.9 &
  \cellcolor{heat!38}84.3 & \cellcolor{heat!30}73.4 \\
TIPSv2-L & 0.49B & \cellcolor{heat!36}\underline{57.6} & \cellcolor{heat!30}73.0 & \cellcolor{heat!21}82.6 &
  \cellcolor{heat!9}93.0 & 61.2 & 81.0 & \cellcolor{heat!17}80.8 & \cellcolor{heat!17}77.7 & 4.7 & 5.7 & 49.8
  & 53.4 & 60.1 \\
TIPSv2-g & 1.53B & \cellcolor{heat!45}\textbf{59.9} & \cellcolor{heat!35}74.0 & \cellcolor{heat!33}84.8 &
  \cellcolor{heat!31}95.5 & \cellcolor{heat!4}63.8 & \cellcolor{heat!11}84.3 & \cellcolor{heat!21}82.7 &
  \cellcolor{heat!21}79.9 & 5.0 & 5.8 & \cellcolor{heat!4}53.3 & \cellcolor{heat!6}58.5 &
  \cellcolor{heat!5}62.3 \\
EmbeddingGemma~2 & 0.44B & \cellcolor{heat!34}57.3 & \cellcolor{heat!19}70.6 & \cellcolor{heat!36}85.3 &
  \cellcolor{heat!28}95.1 & \cellcolor{heat!45}\textbf{86.8} & \cellcolor{heat!45}\textbf{95.0} &
  \cellcolor{heat!45}\textbf{96.0} & \cellcolor{heat!45}\textbf{93.4} & \cellcolor{heat!45}\textbf{56.6} &
  \cellcolor{heat!36}50.9 & \cellcolor{heat!16}62.8 & \cellcolor{heat!7}59.1 & \cellcolor{heat!35}75.8 \\
C-RADIOv4-So400m & 1.15B & \cellcolor{heat!34}57.3 & \cellcolor{heat!28}72.6 & \cellcolor{heat!33}84.9 &
  \cellcolor{heat!30}95.3 & \cellcolor{heat!29}77.8 & \cellcolor{heat!28}89.6 & \cellcolor{heat!19}81.9 &
  \cellcolor{heat!18}78.6 & \cellcolor{heat!40}50.9 & \cellcolor{heat!36}51.8 & \cellcolor{heat!35}78.4 &
  \cellcolor{heat!36}82.5 & \cellcolor{heat!34}75.1 \\
C-RADIOv4-H & 1.37B & \cellcolor{heat!35}57.5 & \cellcolor{heat!28}72.6 & \cellcolor{heat!37}85.6 &
  \cellcolor{heat!35}95.9 & \cellcolor{heat!30}78.6 & \cellcolor{heat!29}90.0 & \cellcolor{heat!20}82.0 &
  \cellcolor{heat!19}78.9 & \cellcolor{heat!40}50.6 & \cellcolor{heat!36}51.9 & \cellcolor{heat!36}79.5 &
  \cellcolor{heat!37}83.8 & \cellcolor{heat!35}75.6 \\
Qwen3-VL-Emb-2B & 2.13B & \cellcolor{heat!35}57.5 & \cellcolor{heat!45}\textbf{76.2} & \cellcolor{heat!39}85.8
  & \cellcolor{heat!42}\underline{96.6} & \cellcolor{heat!31}79.1 & \cellcolor{heat!38}92.8 &
  \cellcolor{heat!35}90.3 & \cellcolor{heat!43}\underline{92.6} & \cellcolor{heat!45}\underline{56.2} &
  \cellcolor{heat!38}53.7 & \cellcolor{heat!28}72.6 & \cellcolor{heat!29}77.1 &
  \cellcolor{heat!39}\underline{77.5} \\
\modelseparator
\rowcolor{black!6}Qwen3-VL-Emb-8B & 8.14B & 61.1 & 80.4 & 87.9 & 97.3 & 82.5 & 93.1 & 94.1 & 95.5 & 67.4 &
  65.0 & 81.3 & 86.4 & 82.7 \\
\bottomrule
\end{tabular}
\end{table}
}\vspace{0.3em}
{\renewenvironment{table}[1][]{}{\par}\begin{table}[tbp]
\centering
\caption{\textbf{Image classification.} Best top-1 accuracy (mAP for PA-100K) over input resolutions and
  transforms. Bold: best, underlined: second best among models up to 4B parameters.}
\label{tab:eval_image_cls}
\vspace{9pt}
\scriptsize
\setlength{\tabcolsep}{1.8pt}
\resizebox{\linewidth}{!}{%
\begin{tabular}{@{}l r ccccccc ccccccccccc c@{}}
\toprule
 &  & \multicolumn{7}{c}{Natural images} & \multicolumn{11}{c}{Fine-grained} & \\
\cmidrule(lr){3-9}\cmidrule(lr){10-20}
Model & Params & \rotatebox{90}{ImageNet} & \rotatebox{90}{ImageNet-V2} & \rotatebox{90}{ImageNet-A} &
  \rotatebox{90}{ImageNet-R} & \rotatebox{90}{ImageNet-Sketch} & \rotatebox{90}{ObjectNet} &
  \rotatebox{90}{Dollar St.} & \rotatebox{90}{DTD} & \rotatebox{90}{Flowers} & \rotatebox{90}{Aircraft} &
  \rotatebox{90}{Pets} & \rotatebox{90}{Food101} & \rotatebox{90}{Country211} & \rotatebox{90}{SUN397} &
  \rotatebox{90}{RESISC45} & \rotatebox{90}{Cars} & \rotatebox{90}{Stanford40} & \rotatebox{90}{PA-100K} &
  Avg. \\
\midrule
\textbf{\textcolor{accentorange}{\ours{}}} & 1.24B & \cellcolor{heat!34}82.3 &
  \cellcolor{heat!45}\textbf{79.9} & \cellcolor{heat!35}88.5 & \cellcolor{heat!24}93.9 &
  \cellcolor{heat!31}72.6 & \cellcolor{heat!35}84.0 & \cellcolor{heat!22}58.2 & \cellcolor{heat!38}72.5 &
  \cellcolor{heat!45}\underline{91.3} & \cellcolor{heat!38}67.8 & \cellcolor{heat!42}96.4 &
  \cellcolor{heat!39}95.1 & \cellcolor{heat!36}48.3 & \cellcolor{heat!38}77.6 & \cellcolor{heat!44}77.6 &
  \cellcolor{heat!44}95.2 & \cellcolor{heat!33}96.1 & \cellcolor{heat!31}47.6 &
  \cellcolor{heat!37}\underline{79.2} \\
PE-Core-L & 0.67B & \cellcolor{heat!39}83.5 & \cellcolor{heat!38}77.9 & \cellcolor{heat!37}89.3 &
  \cellcolor{heat!34}95.2 & \cellcolor{heat!34}73.5 & \cellcolor{heat!37}84.5 & \cellcolor{heat!28}59.5 &
  \cellcolor{heat!25}69.8 & \cellcolor{heat!37}85.7 & \cellcolor{heat!39}68.4 & \cellcolor{heat!42}96.5 &
  \cellcolor{heat!43}96.1 & \cellcolor{heat!32}45.1 & \cellcolor{heat!41}\underline{78.3} &
  \cellcolor{heat!44}77.6 & \cellcolor{heat!40}93.6 & \cellcolor{heat!42}\underline{97.5} &
  \cellcolor{heat!27}46.6 & \cellcolor{heat!36}78.8 \\
PE-Core-G & 2.42B & \cellcolor{heat!45}\textbf{85.2} & \cellcolor{heat!45}\textbf{79.9} &
  \cellcolor{heat!45}\textbf{92.9} & \cellcolor{heat!45}\textbf{96.5} & \cellcolor{heat!43}\underline{76.2} &
  \cellcolor{heat!45}\textbf{86.8} & \cellcolor{heat!45}\textbf{62.5} & \cellcolor{heat!39}72.9 &
  \cellcolor{heat!43}89.9 & \cellcolor{heat!45}\textbf{77.7} & \cellcolor{heat!45}\textbf{97.2} &
  \cellcolor{heat!45}\textbf{96.7} & \cellcolor{heat!45}\textbf{57.2} & \cellcolor{heat!45}\textbf{79.2} &
  \cellcolor{heat!41}76.9 & \cellcolor{heat!43}94.6 & \cellcolor{heat!45}\textbf{98.0} &
  \cellcolor{heat!34}48.2 & \cellcolor{heat!45}\textbf{81.6} \\
MetaCLIP~2-H & 1.86B & \cellcolor{heat!31}81.4 & \cellcolor{heat!29}75.4 & \cellcolor{heat!13}79.0 &
  \cellcolor{heat!19}93.3 & \cellcolor{heat!29}71.7 & \cellcolor{heat!18}78.8 & \cellcolor{heat!19}57.8 &
  \cellcolor{heat!33}71.5 & \cellcolor{heat!39}87.5 & \cellcolor{heat!31}56.3 & \cellcolor{heat!38}95.5 &
  \cellcolor{heat!37}94.6 & \cellcolor{heat!36}48.7 & \cellcolor{heat!37}77.4 &
  \cellcolor{heat!44}\underline{77.7} & \cellcolor{heat!34}90.5 & \cellcolor{heat!23}94.4 & 40.7 &
  \cellcolor{heat!28}76.2 \\
MetaCLIP~2-bigG & 3.63B & \cellcolor{heat!37}83.1 & \cellcolor{heat!38}77.9 & \cellcolor{heat!32}87.2 &
  \cellcolor{heat!32}94.9 & \cellcolor{heat!37}74.2 & \cellcolor{heat!31}82.7 & \cellcolor{heat!23}58.5 &
  \cellcolor{heat!38}72.6 & \cellcolor{heat!45}\textbf{91.4} & \cellcolor{heat!39}\underline{69.4} &
  \cellcolor{heat!42}96.5 & \cellcolor{heat!40}95.5 & \cellcolor{heat!44}\underline{56.5} &
  \cellcolor{heat!35}76.9 & \cellcolor{heat!33}75.0 & \cellcolor{heat!38}92.4 & \cellcolor{heat!30}95.6 &
  \cellcolor{heat!10}43.0 & \cellcolor{heat!37}79.1 \\
SigLIP2-So400m & 1.14B & \cellcolor{heat!39}83.7 & \cellcolor{heat!37}77.7 & \cellcolor{heat!35}88.6 &
  \cellcolor{heat!40}95.8 & \cellcolor{heat!41}75.4 & \cellcolor{heat!41}85.5 & \cellcolor{heat!10}56.0 &
  \cellcolor{heat!30}70.9 & \cellcolor{heat!43}90.0 & \cellcolor{heat!32}58.2 & \cellcolor{heat!42}96.5 &
  \cellcolor{heat!43}\underline{96.2} & \cellcolor{heat!23}36.4 & \cellcolor{heat!16}72.9 &
  \cellcolor{heat!43}77.3 & \cellcolor{heat!45}\underline{95.6} & \cellcolor{heat!30}95.6 &
  \cellcolor{heat!40}49.5 & \cellcolor{heat!33}77.9 \\
SigLIP2-g & 1.87B & \cellcolor{heat!40}\underline{84.0} & \cellcolor{heat!40}\underline{78.5} &
  \cellcolor{heat!34}88.3 & \cellcolor{heat!44}\underline{96.4} & \cellcolor{heat!45}\textbf{76.8} &
  \cellcolor{heat!42}\underline{85.9} & \cellcolor{heat!28}59.4 & \cellcolor{heat!45}\textbf{74.1} &
  \cellcolor{heat!42}89.3 & \cellcolor{heat!30}54.5 & \cellcolor{heat!45}\underline{97.1} &
  \cellcolor{heat!42}96.0 & \cellcolor{heat!25}37.7 & \cellcolor{heat!35}76.9 & \cellcolor{heat!40}76.7 &
  \cellcolor{heat!45}\textbf{95.8} & \cellcolor{heat!24}94.6 & \cellcolor{heat!42}50.0 &
  \cellcolor{heat!35}78.4 \\
TIPSv2-L & 0.49B & \cellcolor{heat!17}77.8 & \cellcolor{heat!12}70.5 & \cellcolor{heat!3}74.4 & 91.0 &
  \cellcolor{heat!14}67.3 & 73.6 & 54.2 & 64.3 & \cellcolor{heat!32}82.5 & 9.8 & \cellcolor{heat!27}92.9 &
  \cellcolor{heat!6}86.4 & 14.2 & \cellcolor{heat!4}70.2 & 67.2 & \cellcolor{heat!23}85.1 & 90.8 & 40.7 & 67.4
  \\
TIPSv2-g & 1.53B & \cellcolor{heat!25}79.8 & \cellcolor{heat!23}73.5 & \cellcolor{heat!18}80.9 &
  \cellcolor{heat!15}92.8 & \cellcolor{heat!23}70.1 & \cellcolor{heat!11}77.0 & \cellcolor{heat!6}55.3 &
  \cellcolor{heat!2}64.6 & \cellcolor{heat!33}83.4 & \cellcolor{heat!1}10.7 & \cellcolor{heat!29}93.3 &
  \cellcolor{heat!21}90.3 & \cellcolor{heat!5}18.5 & \cellcolor{heat!2}69.8 & \cellcolor{heat!22}72.4 &
  \cellcolor{heat!30}88.6 & \cellcolor{heat!8}92.1 & \cellcolor{heat!5}41.7 & \cellcolor{heat!7}69.7 \\
EmbeddingGemma~2 & 0.44B & 73.2 & 66.9 & 73.2 & 90.9 & 63.0 & \cellcolor{heat!4}74.9 & \cellcolor{heat!13}56.6
  & \cellcolor{heat!11}66.6 & \cellcolor{heat!11}68.2 & \cellcolor{heat!22}42.7 & \cellcolor{heat!6}87.9 &
  84.8 & \cellcolor{heat!9}22.8 & 69.4 & \cellcolor{heat!8}69.2 & \cellcolor{heat!29}87.9 &
  \cellcolor{heat!28}95.2 & \cellcolor{heat!35}48.3 & \cellcolor{heat!5}69.0 \\
C-RADIOv4-So400m & 1.15B & \cellcolor{heat!33}81.9 & \cellcolor{heat!32}76.3 & \cellcolor{heat!34}87.9 &
  \cellcolor{heat!33}95.0 & \cellcolor{heat!22}69.7 & \cellcolor{heat!27}81.6 & \cellcolor{heat!22}58.4 &
  \cellcolor{heat!43}73.7 & \cellcolor{heat!38}86.3 & \cellcolor{heat!19}39.1 & \cellcolor{heat!41}96.2 &
  \cellcolor{heat!39}95.2 & \cellcolor{heat!21}34.5 & \cellcolor{heat!37}77.5 & \cellcolor{heat!41}76.9 &
  \cellcolor{heat!42}94.2 & \cellcolor{heat!27}95.1 & \cellcolor{heat!44}\underline{50.4} &
  \cellcolor{heat!28}76.1 \\
C-RADIOv4-H & 1.37B & \cellcolor{heat!37}83.0 & \cellcolor{heat!36}77.5 & \cellcolor{heat!38}\underline{89.6}
  & \cellcolor{heat!37}95.4 & \cellcolor{heat!30}72.3 & \cellcolor{heat!32}83.1 & \cellcolor{heat!26}59.0 &
  \cellcolor{heat!44}\underline{73.9} & \cellcolor{heat!42}89.2 & \cellcolor{heat!23}44.1 &
  \cellcolor{heat!42}96.6 & \cellcolor{heat!41}95.6 & \cellcolor{heat!23}36.2 & \cellcolor{heat!40}78.1 &
  \cellcolor{heat!45}\textbf{77.8} & \cellcolor{heat!43}94.6 & \cellcolor{heat!28}95.3 &
  \cellcolor{heat!45}\textbf{50.6} & \cellcolor{heat!32}77.3 \\
Qwen3-VL-Emb-2B & 2.13B & \cellcolor{heat!10}75.9 & \cellcolor{heat!12}70.4 & \cellcolor{heat!27}85.1 &
  \cellcolor{heat!14}92.6 & \cellcolor{heat!12}66.7 & \cellcolor{heat!24}80.8 &
  \cellcolor{heat!37}\underline{61.1} & \cellcolor{heat!3}65.0 & 60.7 & \cellcolor{heat!13}30.1 & 86.5 &
  \cellcolor{heat!16}89.0 & \cellcolor{heat!12}25.4 & \cellcolor{heat!31}76.2 & \cellcolor{heat!10}69.7 & 73.7
  & \cellcolor{heat!42}\underline{97.5} & \cellcolor{heat!19}44.8 & \cellcolor{heat!7}69.5 \\
\modelseparator
\rowcolor{black!6}Qwen3-VL-Emb-8B & 8.14B & 79.5 & 74.5 & 88.3 & 93.3 & 68.0 & 80.8 & 58.2 & 69.6 & 77.9 &
  36.6 & 88.4 & 89.1 & 29.8 & 79.3 & 70.6 & 80.1 & 97.9 & 49.7 & 72.9 \\
\bottomrule
\end{tabular}}
\end{table}
}
\end{table}

\paragraph{Image retrieval.}
\ours{} has the best all-around performance, with the highest mean R@1 (\cref{tab:eval_image_ret}).
However, much of its lead comes from long-caption and multilingual retrieval (DOCCI and XM3600; see
\cref{sec:long_captions,sec:multilinguality}) and from VisualNews, whose news-photo captions mostly name
specific people, places and events.
EmbeddingGemma~2 is a very strong competitor but falls short on VisualNews (62.8 and 59.1 R@1 vs.\ 85.6 and
89.5 for \ours{}).
Still, \ours{} remains the best on average when VisualNews is excluded (78.8 vs.\ 78.7 for EmbeddingGemma~2
and 78.1 for Qwen3-VL-Embedding-2B), and also when XM3600 is excluded (84.3 vs.\ 82.2 for PE-Core-G).
The 8B Qwen reference, \qwenEightParamRatio{}$\times$ larger than \ours{}, performs best among all evaluated
models.

\paragraph{Image classification.}
\ours{} has the second-best all-around performance (\oursImgClsAvg{} mean top-1; \cref{tab:eval_image_cls}),
behind only PE-Core-G (\clsPeG{}), which is ${\sim}$2$\times$ larger (2.42B vs.\ 1.24B).
On fine-grained benchmarks, \ours{} also ranks second (78.7), again behind PE-Core-G (80.8).
On natural images (ImageNet and its variants, ObjectNet and Dollar Street), however, \ours{} averages 79.9,
within 1.4 points of the models ranked second to fifth and 3.0 points behind PE-Core-G (82.9).

\paragraph{Video training improves image performance.}
Compared with the Stage~1 image-only checkpoint from which the video stage starts (\cref{sec:method}), video
training improves mean image retrieval R@1 by \stageOneImgRetGain{} points and keeps image classification on
par (+\stageOneImgClsGain{} points mean top-1), on the benchmarks and protocol of
\cref{tab:eval_image_ret,tab:eval_image_cls}.
This improvement is in line with \citet{bolya2025perception}, who find that fine-tuning on video--text data
also improves image benchmarks.

\FloatBarrier
\subsection{Video results}
\label{sec:results_video}

\begin{table}[!b]
{\renewenvironment{table}[1][]{}{\par}\begin{table}[tbp]
\centering
\caption{\textbf{Video--text retrieval.} Best top-1 recall (R@1) over input resolutions and transforms. Bold:
  best, underlined: second best among models up to 4B parameters.}
\label{tab:eval_video_ret}
\footnotesize
\setlength{\tabcolsep}{3.4pt}
\begin{tabular}{@{}l r cc cc cc cc cc c@{}}
\toprule
 &  & \multicolumn{2}{c}{MSR-VTT} & \multicolumn{2}{c}{MSVD} & \multicolumn{2}{c}{ActivityNet} &
  \multicolumn{2}{c}{PVD} & \multicolumn{2}{c}{DiDeMo} & \\
\cmidrule(lr){3-4}\cmidrule(lr){5-6}\cmidrule(lr){7-8}\cmidrule(lr){9-10}\cmidrule(lr){11-12}
Model & Params & t2v & v2t & t2v & v2t & t2v & v2t & t2v & v2t & t2v & v2t & Avg. \\
\midrule
\textbf{\textcolor{accentorange}{\ours{}}} & 1.24B & \cellcolor{heat!42}39.6 &
  \cellcolor{heat!45}\textbf{73.4} & \cellcolor{heat!27}60.5 & \cellcolor{heat!39}82.7 &
  \cellcolor{heat!45}\textbf{60.4} & \cellcolor{heat!45}\textbf{58.8} & \cellcolor{heat!43}81.4 &
  \cellcolor{heat!45}\textbf{81.4} & \cellcolor{heat!32}50.2 & \cellcolor{heat!45}\textbf{51.8} &
  \cellcolor{heat!45}\textbf{64.0} \\
PE-Core-L & 0.67B & \cellcolor{heat!41}39.4 & \cellcolor{heat!40}69.6 & \cellcolor{heat!38}\underline{63.7} &
  \cellcolor{heat!39}\underline{82.8} & \cellcolor{heat!22}44.9 & \cellcolor{heat!19}41.5 &
  \cellcolor{heat!18}63.6 & \cellcolor{heat!15}65.7 & \cellcolor{heat!16}40.9 & \cellcolor{heat!29}41.5 &
  \cellcolor{heat!26}55.4 \\
PE-Core-G & 2.42B & \cellcolor{heat!42}39.8 & \cellcolor{heat!39}68.7 & \cellcolor{heat!45}\textbf{65.6} &
  \cellcolor{heat!45}\textbf{84.8} & \cellcolor{heat!34}52.9 & \cellcolor{heat!31}49.6 &
  \cellcolor{heat!37}77.2 & \cellcolor{heat!38}77.7 & \cellcolor{heat!23}45.1 & \cellcolor{heat!38}47.2 &
  \cellcolor{heat!38}60.9 \\
MetaCLIP~2-H & 1.86B & \cellcolor{heat!20}32.1 & \cellcolor{heat!16}51.3 & \cellcolor{heat!13}56.3 & 69.6 &
  \cellcolor{heat!10}36.6 & \cellcolor{heat!6}33.2 & 51.7 & \cellcolor{heat!11}63.4 & \cellcolor{heat!7}35.9 &
  \cellcolor{heat!10}30.1 & \cellcolor{heat!5}46.0 \\
MetaCLIP~2-bigG & 3.63B & \cellcolor{heat!23}33.2 & \cellcolor{heat!19}53.7 & \cellcolor{heat!19}58.1 &
  \cellcolor{heat!4}70.9 & \cellcolor{heat!11}37.6 & \cellcolor{heat!7}34.0 & 51.3 & \cellcolor{heat!9}62.5 &
  \cellcolor{heat!3}33.9 & \cellcolor{heat!12}31.3 & \cellcolor{heat!7}46.6 \\
SigLIP2-So400m & 1.14B & \cellcolor{heat!22}32.9 & \cellcolor{heat!16}51.8 & \cellcolor{heat!31}61.6 &
  \cellcolor{heat!19}76.1 & \cellcolor{heat!6}34.0 & 29.3 & \cellcolor{heat!4}54.2 & 57.7 &
  \cellcolor{heat!8}36.6 & \cellcolor{heat!15}33.2 & \cellcolor{heat!7}46.7 \\
SigLIP2-g & 1.87B & \cellcolor{heat!20}32.4 & \cellcolor{heat!16}51.4 & \cellcolor{heat!32}61.8 &
  \cellcolor{heat!14}74.2 & \cellcolor{heat!7}34.6 & \cellcolor{heat!3}31.5 & \cellcolor{heat!7}56.4 &
  \cellcolor{heat!3}59.5 & \cellcolor{heat!11}38.3 & \cellcolor{heat!17}34.3 & \cellcolor{heat!8}47.4 \\
TIPSv2-L & 0.49B & 25.4 & 39.6 & 52.7 & \cellcolor{heat!4}70.7 & 30.1 & 29.2 & \cellcolor{heat!16}62.4 &
  \cellcolor{heat!23}70.0 & 32.0 & 23.7 & 43.6 \\
TIPSv2-g & 1.53B & \cellcolor{heat!3}26.5 & \cellcolor{heat!3}42.1 & \cellcolor{heat!8}55.1 &
  \cellcolor{heat!19}75.8 & \cellcolor{heat!3}32.4 & \cellcolor{heat!5}32.3 & \cellcolor{heat!24}67.9 &
  \cellcolor{heat!32}74.4 & \cellcolor{heat!4}34.6 & \cellcolor{heat!15}33.2 & \cellcolor{heat!8}47.4 \\
VideoPrism-LvT-L & 0.58B & \cellcolor{heat!44}\underline{40.4} & \cellcolor{heat!41}\underline{70.4} &
  \cellcolor{heat!30}61.4 & \cellcolor{heat!34}80.9 & \cellcolor{heat!23}45.8 & \cellcolor{heat!23}44.1 &
  \cellcolor{heat!31}73.0 & \cellcolor{heat!29}73.0 & \cellcolor{heat!28}47.9 & \cellcolor{heat!40}48.7 &
  \cellcolor{heat!33}58.5 \\
EmbeddingGemma~2 & 0.44B & \cellcolor{heat!41}39.5 & \cellcolor{heat!39}68.6 & \cellcolor{heat!17}57.5 &
  \cellcolor{heat!21}76.7 & \cellcolor{heat!36}54.1 & \cellcolor{heat!33}\underline{51.0} &
  \cellcolor{heat!45}\textbf{82.7} & \cellcolor{heat!43}\underline{80.3} & \cellcolor{heat!32}\underline{50.5}
  & \cellcolor{heat!41}49.3 & \cellcolor{heat!38}61.0 \\
C-RADIOv4-So400m & 1.15B & \cellcolor{heat!17}31.3 & \cellcolor{heat!11}47.6 & \cellcolor{heat!23}59.3 & 69.6
  & \cellcolor{heat!5}33.3 & \cellcolor{heat!2}30.8 & \cellcolor{heat!9}57.6 & \cellcolor{heat!15}65.8 &
  \cellcolor{heat!10}37.7 & \cellcolor{heat!12}31.5 & \cellcolor{heat!6}46.4 \\
C-RADIOv4-H & 1.37B & \cellcolor{heat!19}31.8 & \cellcolor{heat!13}49.5 & \cellcolor{heat!27}60.4 &
  \cellcolor{heat!8}72.1 & \cellcolor{heat!7}34.8 & \cellcolor{heat!3}31.5 & \cellcolor{heat!9}57.4 &
  \cellcolor{heat!17}66.5 & \cellcolor{heat!11}38.4 & \cellcolor{heat!14}32.5 & \cellcolor{heat!9}47.5 \\
Qwen3-VL-Emb-2B & 2.13B & \cellcolor{heat!45}\textbf{40.8} & \cellcolor{heat!31}62.5 & \cellcolor{heat!35}62.6
  & \cellcolor{heat!25}77.9 & \cellcolor{heat!41}\underline{57.6} & \cellcolor{heat!31}49.5 &
  \cellcolor{heat!44}\underline{81.6} & \cellcolor{heat!40}78.5 & \cellcolor{heat!45}\textbf{58.0} &
  \cellcolor{heat!43}\underline{50.4} & \cellcolor{heat!40}\underline{61.9} \\
\modelseparator
\rowcolor{black!6}Qwen3-VL-Emb-8B & 8.14B & 44.0 & 68.6 & 65.1 & 78.4 & 63.6 & 56.5 & 86.9 & 80.6 & 63.2 &
  55.5 & 66.2 \\
\bottomrule
\end{tabular}
\end{table}
}\vspace{0.3em}
{\renewenvironment{table}[1][]{}{\par}\begin{table}[tbp]
\centering
\caption{\textbf{Video classification.} Best top-1 accuracy over input resolutions and transforms.
  Kitchens: EPIC-Kitchens-100. Bold: best, underlined: second best among models up to 4B parameters.}
\label{tab:eval_video_cls}
\vspace{9pt}
\footnotesize
\setlength{\tabcolsep}{3.6pt}
\begin{tabular}{@{}l r cccccccc c@{}}
\toprule
Model & Params & K400 & K600 & K700 & UCF101 & HMDB51 & SSv2 & Kitchens & Breakfast & Avg. \\
\midrule
\textbf{\textcolor{accentorange}{\ours{}}} & 1.24B & \cellcolor{heat!39}73.5 &
  \cellcolor{heat!40}\underline{73.3} & \cellcolor{heat!38}64.8 & \cellcolor{heat!45}\underline{90.3} &
  \cellcolor{heat!26}63.9 & \cellcolor{heat!45}\textbf{24.9} & \cellcolor{heat!23}\underline{7.9} &
  \cellcolor{heat!44}\underline{57.1} & \cellcolor{heat!43}\underline{57.0} \\
PE-Core-L & 0.67B & \cellcolor{heat!39}\underline{73.6} & \cellcolor{heat!39}73.0 &
  \cellcolor{heat!39}\underline{65.4} & \cellcolor{heat!34}86.9 & \cellcolor{heat!23}62.0 &
  \cellcolor{heat!9}9.5 & \cellcolor{heat!10}4.9 & \cellcolor{heat!25}44.8 & \cellcolor{heat!32}52.5 \\
PE-Core-G & 2.42B & \cellcolor{heat!45}\textbf{76.5} & \cellcolor{heat!45}\textbf{75.9} &
  \cellcolor{heat!45}\textbf{68.8} & \cellcolor{heat!45}\textbf{90.4} & \cellcolor{heat!26}\underline{64.4} &
  \cellcolor{heat!13}10.9 & \cellcolor{heat!17}6.5 & \cellcolor{heat!29}47.1 & \cellcolor{heat!38}55.1 \\
MetaCLIP~2-H & 1.86B & \cellcolor{heat!28}68.4 & \cellcolor{heat!31}68.6 & \cellcolor{heat!27}59.1 &
  \cellcolor{heat!26}84.5 & \cellcolor{heat!18}58.1 & \cellcolor{heat!5}7.4 & \cellcolor{heat!6}3.8 &
  \cellcolor{heat!10}34.7 & \cellcolor{heat!21}48.1 \\
MetaCLIP~2-bigG & 3.63B & \cellcolor{heat!35}71.8 & \cellcolor{heat!37}71.7 & \cellcolor{heat!34}62.8 &
  \cellcolor{heat!37}88.0 & \cellcolor{heat!18}58.0 & \cellcolor{heat!2}6.2 & \cellcolor{heat!3}3.2 &
  \cellcolor{heat!11}35.7 & \cellcolor{heat!25}49.7 \\
SigLIP2-So400m & 1.14B & \cellcolor{heat!27}67.9 & \cellcolor{heat!29}67.9 & \cellcolor{heat!27}59.0 &
  \cellcolor{heat!40}88.9 & \cellcolor{heat!13}54.4 & \cellcolor{heat!7}8.5 & \cellcolor{heat!8}4.3 &
  \cellcolor{heat!23}43.6 & \cellcolor{heat!24}49.3 \\
SigLIP2-g & 1.87B & \cellcolor{heat!30}69.1 & \cellcolor{heat!31}69.0 & \cellcolor{heat!29}60.3 &
  \cellcolor{heat!44}90.1 & \cellcolor{heat!16}56.4 & \cellcolor{heat!7}8.7 & \cellcolor{heat!6}3.8 &
  \cellcolor{heat!22}42.8 & \cellcolor{heat!26}50.0 \\
TIPSv2-L & 0.49B & 54.8 & 53.0 & 44.9 & 76.6 & 45.0 & 5.4 & 2.4 & \cellcolor{heat!5}31.2 & 39.2 \\
TIPSv2-g & 1.53B & \cellcolor{heat!8}58.6 & \cellcolor{heat!8}56.9 & \cellcolor{heat!7}48.8 &
  \cellcolor{heat!18}82.2 & \cellcolor{heat!2}46.1 & \cellcolor{heat!3}6.6 & 2.3 & 28.1 &
  \cellcolor{heat!5}41.2 \\
VideoPrism-LvT-L & 0.58B & \cellcolor{heat!34}71.4 & \cellcolor{heat!33}70.0 & \cellcolor{heat!34}62.8 &
  \cellcolor{heat!28}85.2 & \cellcolor{heat!22}61.1 & \cellcolor{heat!8}8.8 & \cellcolor{heat!12}5.3 &
  \cellcolor{heat!29}47.1 & \cellcolor{heat!30}51.5 \\
EmbeddingGemma~2 & 0.44B & \cellcolor{heat!26}67.2 & \cellcolor{heat!25}65.7 & \cellcolor{heat!24}57.9 &
  \cellcolor{heat!13}80.7 & \cellcolor{heat!20}59.6 & \cellcolor{heat!11}10.1 & \cellcolor{heat!7}4.1 &
  \cellcolor{heat!38}53.5 & \cellcolor{heat!26}49.9 \\
C-RADIOv4-So400m & 1.15B & \cellcolor{heat!18}63.6 & \cellcolor{heat!21}63.6 & \cellcolor{heat!18}54.3 &
  \cellcolor{heat!29}85.6 & \cellcolor{heat!13}54.8 & \cellcolor{heat!3}6.5 & \cellcolor{heat!3}3.1 &
  \cellcolor{heat!22}42.7 & \cellcolor{heat!18}46.8 \\
C-RADIOv4-H & 1.37B & \cellcolor{heat!22}65.2 & \cellcolor{heat!24}65.3 & \cellcolor{heat!21}56.1 &
  \cellcolor{heat!34}87.0 & \cellcolor{heat!15}55.9 & \cellcolor{heat!3}6.8 & \cellcolor{heat!5}3.5 &
  \cellcolor{heat!19}40.9 & \cellcolor{heat!20}47.6 \\
Qwen3-VL-Emb-2B & 2.13B & \cellcolor{heat!33}70.8 & \cellcolor{heat!34}70.0 & \cellcolor{heat!33}62.3 &
  \cellcolor{heat!42}89.5 & \cellcolor{heat!45}\textbf{78.0} & \cellcolor{heat!35}\underline{20.6} &
  \cellcolor{heat!45}\textbf{13.4} & \cellcolor{heat!45}\textbf{58.1} & \cellcolor{heat!45}\textbf{57.8} \\
\modelseparator
\rowcolor{black!6}Qwen3-VL-Emb-8B & 8.14B & 73.5 & 73.0 & 65.3 & 92.8 & 83.1 & 29.2 & 15.3 & 61.6 & 61.7 \\
\bottomrule
\end{tabular}
\end{table}
}
\end{table}

\paragraph{Video retrieval.}
\ours{} is the best video retriever, with the highest mean R@1 (64.0 vs.\ 61.9 for Qwen3-VL-Embedding-2B;
\cref{tab:eval_video_ret}), and ranks first or second on four of the five benchmarks.
Its lead is largest on ActivityNet Captions~\citep{krishna2017activitynet}, where each query is a paragraph
describing several temporally localized events in a long, untrimmed video (median \vidDurActivityNet{}\,s),
whereas MSR-VTT and MSVD pair short clips (median \vidDurMsrvtt{}\,s and \vidDurMsvd{}\,s) with one-sentence
captions.
On ActivityNet, \ours{} is the best in both directions and 6.1 points ahead of the next model on average (59.6
vs.\ 53.5 for Qwen3-VL-Embedding-2B). \ours{} also performs strongly on PVD, the benchmark introduced with
Perception Encoder~\citep{bolya2025perception}: 15k held-out, motion-centered videos whose captions were
refined by human annotators to add details and remove hallucinations.
On PVD, \ours{} is on par with the best model (81.4 vs.\ 81.5 for EmbeddingGemma~2, averaged over directions)
and is the best in video-to-text.

\paragraph{Video classification.}
\ours{} ranks second among models up to 4B, \dVidClsVsQwen{} points behind Qwen3-VL-Embedding-2B with tuned
instructions, which is ${\sim}$1.7$\times$ larger (2.13B vs.\ 1.24B; \cref{tab:eval_video_cls}).
On benchmarks of fine-grained, temporally structured actions, \ours{} is the best or second-best model.
On SSv2, whose classes describe object interactions such as ``pretending to be tearing something that is not
tearable'' or ``moving something across a surface until it falls down'', \ours{} scores 24.9, compared with at
most 20.6 for the other models up to 4B. On EPIC-Kitchens-100 (first-person kitchen videos labeled with
verb--noun actions such as ``squeeze paste'') and on Breakfast (long videos of people preparing breakfast
dishes), \ours{} is second to Qwen3-VL-Embedding-2B (7.9 vs.\ 13.4 and 57.1 vs.\ 58.1, respectively).
On Kinetics-400/600/700, \ours{} is among the top three models (73.5, 73.3 and 64.8), behind PE-Core-G (76.5,
75.9 and 68.8).
However, Kinetics can largely be solved from the appearance of individual frames: MetaCLIP~2-bigG, an
image-only model, scores 71.8 on Kinetics-400, above the video models VideoPrism-LvT-L (71.4) and
Qwen3-VL-Embedding-2B~(70.8).

\FloatBarrier
\subsection{Long captions}
\label{sec:long_captions}

\begin{table}[!b]
\centering
\caption{\textbf{Long-caption retrieval.} Best top-1 recall (R@1, mean of both directions) over input
  resolutions and transforms. DOCCI: images; ActivityNet, PVD, DiDeMo: videos; Ctx.: maximum text tokens of
  the model. Bold: best, underlined: second best among models up to 4B parameters.}
\label{tab:longcaption}
\footnotesize
\setlength{\tabcolsep}{4pt}
\begin{tabular}{@{}l r r c ccc c@{}}
\toprule
 & & & Image & \multicolumn{3}{c}{Video} & \\
\cmidrule(lr){4-4}\cmidrule(lr){5-7}
Model & Params & Ctx. & DOCCI & ActivityNet & PVD & DiDeMo & Avg. \\
\midrule
\textbf{\textcolor{accentorange}{\ours{}}} & 1.24B & -- & \cellcolor{heat!41}\underline{92.7} &
  \cellcolor{heat!45}\textbf{59.6} & \cellcolor{heat!45}\underline{81.4} & \cellcolor{heat!40}\underline{51.0}
  & \cellcolor{heat!45}\textbf{71.2} \\
PE-Core-L & 0.67B & 32 & 70.0 & \cellcolor{heat!20}43.2 & \cellcolor{heat!15}64.6 & \cellcolor{heat!23}41.2 &
  \cellcolor{heat!13}54.8 \\
PE-Core-G & 2.42B & 72 & \cellcolor{heat!24}83.1 & \cellcolor{heat!32}51.3 & \cellcolor{heat!38}77.5 &
  \cellcolor{heat!31}46.2 & \cellcolor{heat!32}64.5 \\
MetaCLIP~2-H & 1.86B & 77 & \cellcolor{heat!17}79.3 & \cellcolor{heat!8}34.9 & \cellcolor{heat!3}57.5 &
  \cellcolor{heat!9}33.0 & \cellcolor{heat!5}51.2 \\
MetaCLIP~2-bigG & 3.63B & 77 & \cellcolor{heat!20}80.9 & \cellcolor{heat!9}35.8 & \cellcolor{heat!2}56.9 &
  \cellcolor{heat!8}32.6 & \cellcolor{heat!6}51.6 \\
SigLIP2-So400m & 1.14B & 64 & \cellcolor{heat!2}71.2 & \cellcolor{heat!3}31.7 & 56.0 & \cellcolor{heat!12}34.9
  & 48.4 \\
SigLIP2-g & 1.87B & 64 & \cellcolor{heat!4}72.5 & \cellcolor{heat!5}33.0 & \cellcolor{heat!3}57.9 &
  \cellcolor{heat!14}36.3 & \cellcolor{heat!3}49.9 \\
TIPSv2-L & 0.49B & 64 & \cellcolor{heat!17}79.3 & 29.7 & \cellcolor{heat!18}66.2 & 27.8 &
  \cellcolor{heat!5}50.7 \\
TIPSv2-g & 1.53B & 64 & \cellcolor{heat!21}81.3 & \cellcolor{heat!4}32.3 & \cellcolor{heat!27}71.1 &
  \cellcolor{heat!10}33.9 & \cellcolor{heat!12}54.7 \\
VideoPrism-LvT-L & 0.58B & 64 & -- & \cellcolor{heat!23}44.9 & \cellcolor{heat!30}73.0 &
  \cellcolor{heat!35}48.3 & -- \\
EmbeddingGemma~2 & 0.44B & 8{,}192 & \cellcolor{heat!45}\textbf{94.7} & \cellcolor{heat!34}52.6 &
  \cellcolor{heat!45}\textbf{81.5} & \cellcolor{heat!38}49.9 & \cellcolor{heat!42}69.7 \\
C-RADIOv4-So400m & 1.15B & 64 & \cellcolor{heat!19}80.3 & \cellcolor{heat!4}32.1 & \cellcolor{heat!10}61.7 &
  \cellcolor{heat!12}34.6 & \cellcolor{heat!7}52.2 \\
C-RADIOv4-H & 1.37B & 64 & \cellcolor{heat!19}80.5 & \cellcolor{heat!5}33.2 & \cellcolor{heat!11}62.0 &
  \cellcolor{heat!13}35.5 & \cellcolor{heat!9}52.8 \\
Qwen3-VL-Emb-2B & 2.13B & -- & \cellcolor{heat!39}91.4 & \cellcolor{heat!36}\underline{53.5} &
  \cellcolor{heat!43}80.1 & \cellcolor{heat!45}\textbf{54.2} & \cellcolor{heat!42}\underline{69.8} \\
\modelseparator
\rowcolor{black!6}Qwen3-VL-Emb-8B & 8.14B & -- & 94.8 & 60.0 & 83.8 & 59.4 & 74.5 \\
\bottomrule
\end{tabular}
\end{table}

We evaluate how well \ours{} handles retrieval with long captions.
In \cref{tab:longcaption}, we consider the subset of benchmarks from
\cref{sec:results_image,sec:results_video} that have long captions: DOCCI, ActivityNet, PVD and DiDeMo.
Their median caption lengths are 152--598 characters, 2.6--10$\times$ more than in any other public retrieval
or classification benchmark we evaluate (\cref{fig:caption_words}, \cref{app:short_context}).
On this long-caption suite, \ours{} has the highest mean R@1 among models up to 4B (\cref{tab:longcaption}).
Notably, several baselines are penalized by their limited context lengths (32--77 tokens), which severely
truncate the long captions of these benchmarks (\cref{tab:truncation}).

\subsection{Multilingual retrieval}
\label{sec:multilinguality}

\begin{table}[!b]
\centering
\caption{\textbf{Multilingual image--text retrieval on XM3600.} Best top-1 recall (R@1, mean of both
  directions) over input resolutions and transforms. En: English; Others: mean of the 35 other languages; All:
  mean of the 36 languages. Bold: best, underlined: second best among models up to 4B parameters.}
\label{tab:xm3600}
\footnotesize
\setlength{\tabcolsep}{4pt}
\begin{tabular}{@{}l r ccc@{}}
\toprule
 & & \multicolumn{3}{c}{Avg.\ by language} \\
\cmidrule(l){3-5}
Model & Params & En & Others & All \\
\midrule
\textbf{\textcolor{accentorange}{\ours{}}} & 1.24B & \cellcolor{heat!34}57.8 &
  \cellcolor{heat!45}\textbf{59.7} & \cellcolor{heat!45}\textbf{59.7} \\
PE-Core-L & 0.67B & \cellcolor{heat!32}57.4 & \cellcolor{heat!6}11.6 & \cellcolor{heat!6}12.9 \\
PE-Core-G & 2.42B & \cellcolor{heat!27}56.3 & \cellcolor{heat!8}13.7 & \cellcolor{heat!8}14.8 \\
MetaCLIP~2-H & 1.86B & 50.9 & \cellcolor{heat!41}54.9 & \cellcolor{heat!41}54.8 \\
MetaCLIP~2-bigG & 3.63B & \cellcolor{heat!1}51.1 & \cellcolor{heat!43}\underline{56.7} &
  \cellcolor{heat!42}\underline{56.6} \\
SigLIP2-So400m & 1.14B & \cellcolor{heat!8}52.6 & \cellcolor{heat!37}49.3 & \cellcolor{heat!36}49.3 \\
SigLIP2-g & 1.87B & \cellcolor{heat!12}53.4 & \cellcolor{heat!37}49.7 & \cellcolor{heat!37}49.8 \\
TIPSv2-L & 0.49B & \cellcolor{heat!45}\textbf{60.1} & 3.7 & 5.2 \\
TIPSv2-g & 1.53B & \cellcolor{heat!35}58.0 & 3.9 & 5.4 \\
EmbeddingGemma~2 & 0.44B & \cellcolor{heat!37}58.5 & \cellcolor{heat!40}53.6 & \cellcolor{heat!40}53.7 \\
C-RADIOv4-So400m & 1.15B & \cellcolor{heat!20}55.0 & \cellcolor{heat!38}51.2 & \cellcolor{heat!38}51.3 \\
C-RADIOv4-H & 1.37B & \cellcolor{heat!21}55.1 & \cellcolor{heat!38}51.2 & \cellcolor{heat!38}51.3 \\
Qwen3-VL-Emb-2B & 2.13B & \cellcolor{heat!40}\underline{59.1} & \cellcolor{heat!41}54.9 &
  \cellcolor{heat!41}55.0 \\
\modelseparator
\rowcolor{black!6}Qwen3-VL-Emb-8B & 8.14B & 61.7 & 66.3 & 66.2 \\
\bottomrule
\end{tabular}
\end{table}

In \cref{tab:xm3600}, we focus on the multilingual image retrieval benchmark
XM3600~\citep{thapliyal2022xm3600}, which includes 36 languages, each equally represented.
We report R@1 for English, for all other languages, and averaged across all languages. \ours{} is ahead of
every model up to 4B on the all-language metrics and leads especially on the non-English languages.
PE-Core-L and the TIPSv2 models fall to \lxXmOthersEngLo{}--\lxXmOthersEngHi{} R@1 on the other languages, and
PE-Core-G to 13.7.

\subsection{Resolution robustness}
\label{sec:results_resolution}

\begin{figure}[t]
\centering
\includegraphics[width=\linewidth]{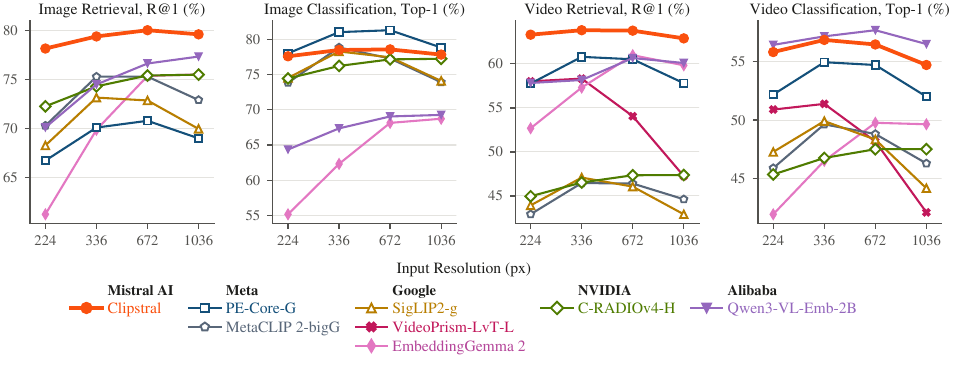}
\caption{\textbf{Performance vs.\ resolution.}
We show \ours{} and, for each of Meta, Google, NVIDIA and Alibaba, the models up to 4B parameters that are the
best of their company in at least one task (VideoPrism-LvT-L is video-only).
At each resolution (224, 336, 672 and 1036\,px), we average benchmark scores using the best of three input
transforms.
The expanded comparison is in \cref{fig:resolution_all} (\cref{app:perres}), and protocol details are in
\cref{sec:protocol} and \cref{app:eval}.}
\label{fig:resolution}
\end{figure}

Image and video collections contain a range of native resolutions, so an encoder should remain accurate as
input size changes. \cref{fig:resolution} compares \ours{} with the selected baselines at 224, 336, 672 and
1036\,px (\cref{sec:protocol}). \ours{} remains stable across this range and leads image and video retrieval
at every resolution.
In classification, it is second at every resolution, except for image classification at 336\,px, where it is
fourth and within 0.4 points of MetaCLIP~2-bigG and \mbox{PE-Core-L}.
PE-Core-G ($\sim$2$\times$ larger than \ours{}) leads image classification, and Qwen3-VL-Embedding-2B (tuned,
$\sim$1.7$\times$ larger) leads video classification, at every resolution.

\subsection{\vstarcls{}}
\label{sec:results_vstar}

To test the recognition of small details at natively high resolution, we introduced \vstarcls{} in
\cref{sec:vstar}.
\cref{tab:vstar_main} reports top-1 accuracy on full images and oracle crops (chance: \vstarChance{}\%).
Note that the \vstarcls{} captions are long (median of 303 characters, see \cref{fig:caption_words}), so the
text encoders of several baselines truncate them, sometimes removing the answer-bearing detail
(\cref{tab:truncation}; \cref{app:vstar}).

\begin{table}[!b]
\centering
\footnotesize
\setlength{\tabcolsep}{5pt}
\caption{\textbf{\vstarcls{} top-1 accuracy (\%).} Full native image and oracle crop, 191 items
(chance: \vstarChance{}\%).
Bold: best, underlined: second best among models up to 4B parameters. $^\dagger$Mean accuracy of four human
annotators per column, choosing among the same candidate captions (\cref{app:vstar}).
Details in \cref{sec:vstar} and \cref{app:vstar}.}
\label{tab:vstar_main}
\begin{tabular}{@{}l cc@{}}
\toprule
Model & Full image & Oracle crop \\
\midrule
Chance & 36.0 & 36.0 \\
Human$^\dagger$ & 98.6 & 99.6 \\
\midrule
\textbf{\textcolor{accentorange}{\ours{}}} & \cellcolor{heat!22}40.3 & \cellcolor{heat!45}\textbf{67.5} \\
PE-Core-L & \cellcolor{heat!3}36.6 & 36.6 \\
PE-Core-G & \cellcolor{heat!10}38.0 & \cellcolor{heat!35}60.5 \\
MetaCLIP~2-H & \cellcolor{heat!45}\textbf{45.0} & \cellcolor{heat!36}61.3 \\
MetaCLIP~2-bigG & \cellcolor{heat!24}40.8 & \cellcolor{heat!34}59.7 \\
SigLIP2-So400m & \cellcolor{heat!33}42.7 & \cellcolor{heat!33}59.4 \\
SigLIP2-g & \cellcolor{heat!18}39.5 & \cellcolor{heat!26}54.7 \\
TIPSv2-L & 36.0 & \cellcolor{heat!26}54.3 \\
TIPSv2-g & \cellcolor{heat!13}38.6 & \cellcolor{heat!26}54.8 \\
EmbeddingGemma~2 & \cellcolor{heat!42}\underline{44.5} & \cellcolor{heat!37}\underline{61.8} \\
C-RADIOv4-So400m & \cellcolor{heat!9}37.7 & \cellcolor{heat!31}57.6 \\
C-RADIOv4-H & \cellcolor{heat!29}41.9 & \cellcolor{heat!35}60.7 \\
Qwen3-VL-Emb-2B & \cellcolor{heat!22}40.3 & \cellcolor{heat!37}\underline{61.8} \\
\modelseparator
\rowcolor{black!6}Qwen3-VL-Emb-8B & 45.5 & 68.1 \\
\bottomrule
\end{tabular}

\end{table}
On full images, every model scores within ten percentage points of chance, whereas humans can discriminate the
relevant visual details.
This suggests that models are still far from solving this benchmark.
Given an oracle crop around the small detail referenced by the captions, \ours{} reaches
\oursVstarCrop{}\%, the highest among models up to 4B parameters and on par with
Qwen3-VL-Embedding-8B. The crop preserves native pixel density and makes the detail easier to locate. \par

\subsection{Inference analysis}
\label{sec:throughput}

\begin{figure}[t]
\centering
\includegraphics[width=\linewidth]{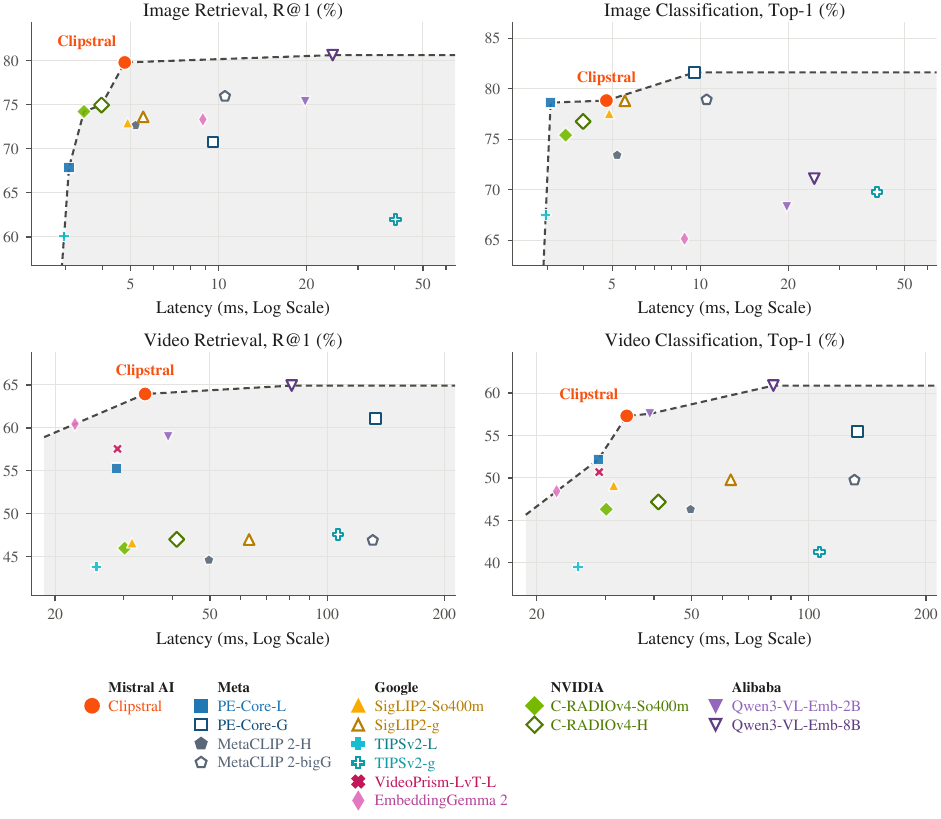}
\caption{\textbf{Performance vs.\ batch-1 latency.}
Mean performance at 448\,px versus encoding latency of one image/video on one H200 with \texttt{torch.compile}
(videos: 16 frames; log scale; lower is better).
Preprocessing and tokenization are excluded from timing.
The dashed black line marks the Pareto frontier.
No model has both lower latency and higher accuracy than \ours{} in image or video retrieval or
classification.
See \cref{sec:protocol} and \cref{app:speed} for evaluation and timing details.}
\label{fig:acc_latency}
\end{figure}

Retrieval and classification scores measure embedding quality, but the choice of which model to deploy also
depends on how fast it produces those embeddings, both when answering a single query and when indexing large
collections of data.
We time every model under one protocol: one H200 GPU, bf16, default \texttt{torch.compile} and a fixed input
resolution of 448\,px, excluding preprocessing and tokenization from the measurement (\cref{app:speed}).
\cref{fig:acc_latency} shows the performance versus the latency of encoding one image/video. \ours{} pushes
the Pareto frontier in image and video retrieval and classification, and in none of them does a model have
both lower latency and higher accuracy.
The latency timings use the released inference code for \ours{} and publicly available model code and weights
for the baselines, with model-specific evaluation adapters (\cref{app:speed}).

\FloatBarrier

\subsection{Video architecture ablations}
\label{sec:abl_video_pool}

\begin{table}[!b]
\centering
\caption{\textbf{Ablation on the ST encoder.} We compare the video retrieval and classification performance of
  configurations that differ in the pooling strategy and in the use of the ST encoder, which together
  determine how the final video embedding is obtained. The image-only checkpoint serves as the reference.
  (a)--(c) apply different pooling methods (attention, mean, max) to the patch tokens of all frames without an
  ST encoder, (d) applies the ST encoder to per-frame embeddings, (e) applies it to the patch tokens of all
  frames, and (f) adds ST encoder alignment before full fine-tuning. A tick in the alignment column means 500
  steps of training only the new module and the contrastive projector: the attention pooler in (a), or the ST
  encoder in (d) and (f).}
\label{tab:arch_ablation}
\small
\setlength{\tabcolsep}{4pt}
\begin{tabular}{@{}l l l c cc@{}}
\toprule
 & Pooling & ST encoder on & Alignment & Video ret. & Video cls. \\
\midrule
\multicolumn{4}{@{}l}{Image-only checkpoint (initialization)} & 53.2 & 57.4 \\
\midrule
(a) & attention & \xmark & \cmark & 53.6 & 55.4 \\
(b) & mean & \xmark & \xmark & 56.8 & 58.7 \\
(c) & max & \xmark & \xmark & 59.4 & 59.0 \\
\midrule
(d) & max & frame embeddings & \cmark & 60.7 & 59.2 \\
(e) & max & patch tokens & \xmark & 61.5 & 59.4 \\
\rowcolor{accentorange!12}(f) & max & patch tokens & \cmark & \textbf{62.9} & \textbf{59.9} \\
\bottomrule
\end{tabular}
\end{table}

The image-only vision tower of \cref{sec:method} encodes each frame independently.
A video embedding therefore requires aggregating the frame features over time, ideally capturing the
relationships among frames.
We ablate two choices for this: a pooling operator over frames and an additional spatio-temporal (ST) encoder
(\cref{tab:arch_ablation}).

\paragraph{Pooling operators.}
We compare mean, max and attention pooling over the patch tokens of all 16 frames (rows (a)--(c) of
\cref{tab:arch_ablation}).
The attention pooler uses one learned query token in a multi-head cross-attention layer followed by a residual
MLP. Like the ST encoder in rows (d) and (f), the attention pooler is first trained alone (alignment phase),
and this phase counts toward the training FLOP budget.

Max pooling gives the highest video retrieval and classification scores of the three.
Mean pooling improves both scores over the image-only checkpoint, while attention pooling stays close to it in
retrieval and falls below it in classification.
This may reflect the use of max pooling during image-only training: retaining the same operator preserves how
patch features are aggregated.
Therefore, we choose max pooling to aggregate the features from each frame.

\paragraph{Adding the ST encoder.}
With max pooling fixed, we test the spatio-temporal (ST) encoder of \cref{sec:method_arch}.
Applying it to the patch tokens of all frames improves video retrieval over max pooling alone (row (e) vs.\
row (c)).
ST encoder alignment before full fine-tuning improves retrieval further and slightly raises video
classification (row (f) vs.\ row (e)).
Applying the ST encoder to patch tokens also outperforms applying it to per-frame embeddings (row (f) vs.\ row
(d)), suggesting that access to spatial detail helps it model temporal relationships.
We also tried making the pretrained ViT's own attention joint over all frames instead, but this degraded image
accuracy (\cref{app:negative}).

\paragraph{Ablation setup.}
All runs in \cref{tab:arch_ablation} are matched in logged training FLOPs.
They use 64 H200 GPUs, 16-frame clips from the curated mixture without image replay, and 96k tokens per GPU
per step.
Both phases use AdamW (weight decay 0.1), with peak learning rates of $10^{-4}$ for alignment and
$4.2{\times}10^{-7}$ for full fine-tuning.
Each phase has a 50-step learning-rate warm-up followed by a cosine and then a linear decay.

\subsection{Video data ablations}
\label{sec:abl_phases}

\begin{figure}[t]
\centering
\includegraphics[width=\linewidth]{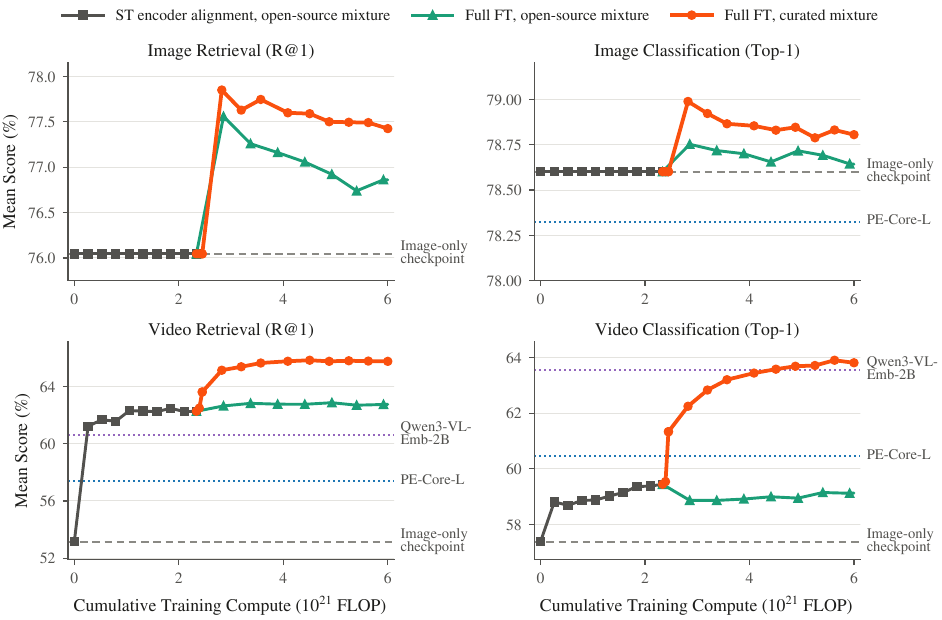}
\caption{\textbf{Curated vs.\ open-source mixture.}
Mean score at 224\,px against cumulative training compute.
Both training branches start from the same ST encoder alignment on the open-source mixture (gray), then either
unfreeze all parameters and continue on the open-source mixture (green), or switch to the curated mixture
(orange, final recipe).
Dashed: image-only checkpoint; dotted: PE-Core-L and Qwen3-VL-Embedding-2B (tuned).
Off scale: image retrieval, PE-Core-L 65.3 and Qwen3-VL-Embedding-2B 70.9; image classification,
Qwen3-VL-Embedding-2B 64.7.}
\label{fig:traj_phases}
\end{figure}

As discussed in \cref{sec:method}, we train \ours{} on two video mixtures, the 48M open-source mixture and the
8.8M curated mixture.
The first phase of training, ST encoder alignment, uses the open-source mixture.
After this phase, we can either continue with full fine-tuning on the open-source mixture or switch to the
curated mixture. \cref{fig:traj_phases} shows that, at the same training compute, switching to the curated
mixture improves the retrieval and classification scores, while continuing on the open-source mixture does
not.
We therefore use the open-source mixture only for the ST encoder alignment.
Omitting it from that phase lowers video retrieval by 0.5 points and changes the other groups by at most 0.24
points (\cref{app:st_alignment}, \cref{fig:abl_warmup_50m}).
Since these differences are at most 0.5 points, the open-source mixture appears helpful but not essential for
the initial ST encoder alignment.

\paragraph{Ablation setup.}
The setup is similar to that of \cref{sec:abl_video_pool} (16-frame clips, 96k tokens per GPU per step, AdamW
with weight decay 0.1), with two differences.
The runs use 256 H200 GPUs (24.6M tokens per step), and full fine-tuning replays 10\% of batches from the
Stage~1 image--text mixture with a peak learning rate of $8.4{\times}10^{-7}$.

\section{Conclusion}
\label{sec:conclusion}

We introduced \ours{}, an open 1.24B image--video--text dual encoder that combines variable-resolution image
and video encoding, long context, multilingual text inputs, and efficient inference.
In our unified evaluation (\cref{fig:larger_models}), it outperforms open-weight models up to 4B parameters on
image and video retrieval, ranks second on image and video classification (slightly behind a few models
1.7--2$\times$ larger), and remains accurate across input resolutions (\cref{fig:resolution}).
We also introduced \vstarcls{}, a novel evaluation task to test classification of small details in native
high-resolution images; on this benchmark, all models are far from saturation.

\paragraph{Limitations and future directions.}
\label{sec:limitations}
Video training focuses on short clips (up to ${\sim}$1 minute) represented by 16 uniformly sampled frames.
This limits coverage of longer videos, for which more than 16 frames or segment-level aggregation will likely
be necessary.
The current model also embeds each modality separately (an image, a video, or text) and does not yet support
flexible multimodal inputs, such as image--text or video--text combinations, or specialized content such as
documents and code.
Future iterations of the model will explore longer videos and multimodal inputs.

To conclude, \ours{} provides state-of-the-art image and video retrieval among open models up to 4B
parameters, in an efficient dual encoder that pushes the Pareto frontier of performance vs.\ size and of
performance vs.\ latency (\cref{fig:larger_models,fig:acc_latency}).
We release the model weights, inference code and \vstarcls{} under an open license to support further
work on these capabilities.

\subsection*{Core contributors}
\label{sec:contributors}
Umberto Tomasini, Federico Baldassarre, Rupert Menneer, Matthieu Futeral, Josselin Somerville Roberts,
Matthieu Andr\'e, Hugo Thimonier, Sean Cha, Kilian Tep, Emmanuel Gottlob, Tomas Hodan, Sagar Vaze

\subsection*{Contributors}
{\hyphenpenalty=10000\exhyphenpenalty=10000
Aaron Buhendwa, Abdelaziz Bounhar, Abhijeet Somani, Aditi Kabra, Adrian Valente, Adrien Petralia, Adrien
Sad\'e, Agnes Valenti, Alan Jeffares, Albert Jiang, Alberto Bietti, Aleksandr Dremov, Aleksandr Timashov,
Alexandre Cahill, Alexandre Gavaudan, Alexandre Laval, Alexandre Sablayrolles, Alexandros Sanchez Vassopoulos,
Am\'elie H\'eliou, Amos You, Andr\'e Jonasson, Andrea Tirinzoni, Andreas F\"urst, Andrew Bai, Andrew Zhao,
Andy Ehrenberg, Andy Lo, Angele Lenglemetz, Anmol Agarwal, Ansgar Pfeifer, Antonia Calvi, Antonio Tirone,
Arata Suzuki, Arjun Majumdar, Arnaud Tournier, Arnav Arora, Artem Korkhov, Arthur Fournier, Arthur Wilcke,
Artjom Joosen, Ash Kumar, Augustin Lombard, Avi Sooriyarachchi, Aylin Guliz Akkus, Aysenur Karaduman Utkur,
Bachir Djermani, Baptiste Bout, Baptiste Rozi\`ere, Barna P\'asztor, Baudouin De Monicault, Benjamin
Holzschuh, Benjamin Tibi, Benno Stein, Bernhard Stadlbauer, Bhargavi Bhaskar, Binbing Hou, B\l{}a\.{z}ej
Osi\'nski, Boris van Breugel, Brennan Sprinkle, Camille Le Scao, Chaoran Yu, Charlotte Cronj\"ager, Chen-Yo
Sun, Chris Bamford, Christian Wallenwein, Christophe Renaudin, Claudio Teixeira Medeiros, Cl\'emence
Lanfranchi, Colin Bonatti, Corentin Barreau, Corentin Sautier, Cyprien Courtot, Daniel Lopes, Daniel Marczak,
Darius Dabert, David Byrne, Diego Couto de las Casas, Dominik Nuss, Dylan Rubini, Dzmitry Soupel, Edouard
Vilain, Elizaveta Demyanenko, Elliot Chane-Sane, Emilien Fugier, Emma Bou Hanna, Erik Aas, Etienne Duchesne,
Etienne Goffinet, Fabian Paischer, Fabian Schlager, Faruk Ahmed, F\'elix Rosseeuw, Filip Szatkowski, Filippo
Vicentini, Florian Sestak, Florian Wiesner, Gabrielle Berrada, Ga\"etan Ecrepont, Gaspard Blanchet, Gaspard
Donada Vidal, Gaurav Mahajan, Gauthier Guinet, Genevieve Hayes, George Christoglou, Georgii Novikov, Giada
Pistilli, Gonzague de Carpentier, Graeme Nail, Guillaume Breton, Guillaume Kunsch, Guillaume Lample, Guillaume
Martin, Guillaume Neau, Guillaume Raille, Guillaume Richard, Guillem Simeon Miquel, Gunjan Dhanuka, Gunshi
Gupta, Hamza Senhaji Rhazi, Han Zhou, H\'edi Hadiji, Himanshu Jaju, Hope McGovern, Hubert de Parseval, Hugo
Cisneros, Hugo Germain, Hugo Venturini, Hyeonju Lee, Hyunwoo Oh, Iliass Lasri, Ilyas Oulkadda, Jacques Sun,
Jade Hardouin, Jakub Krajewski, Jan Ludziejewski, Jason Rute, Jean de Becdeli\`evre, Jean Quentin, Jean-Malo
Delignon, Jessica Chudnovsky, Jianchao Li, Jie Zhang, Joachim Studnia, Joep Barmentlo, Johannes Brandstetter,
John Harvill, Jonas Amar, Jonas Schweizer, Jonathan Crabbe, Jonathan Roberts, Joran Lereec, Jos\'ephine Delas,
Jules Depersin, Julien Denize, Julien Tauran, Julien Verlaguet, Karmesh Yadav, Kartik Khandelwal, Kathleen
Butler, Kelvin Ou, Khemi Chew Jia Jin, Khyathi Chandu, Konstantin Mishchenko, Kris Jankowski, Kriz Chan, Kush
Jain, Larissa Laich, Laura Calem, Laurence Aitchison, Laurent Callot, Laurent Direr, Laurent Fainsin, L\'eo
Cotteleer, L\'eonard Blier, Loic Chadoutaud, Lo\"ick Chambon, Lorenzo Proietti, Louis Faucon, Louis Martin, Lu
Li, Luc Boudier, Lucas Fernandes Martins, Lucas Gestin, Lucas Mebille, Lucien Doustaly, Lucile Saulnier,
Ludovic Ho Fuh, Luis Montero, \L{}ukasz Borchmann, Maarten Buyl, Malik Hassan Raza, Manon Chossegros, Marcin
Mo\.{z}ejko, Margaret Jennings, Markus Hennerbichler, Martin Alexandre, Mathieu Poir\'ee, Mathieu Schmitt,
Mathilde Guillaumin, Mathis Felardos Saint Jean, Matthieu Dinot, Maurits Bleeker, Mauro Comi, Max Mynter,
Maxim Hugo Berman, Maxime Darrin, Maxime Louis, Maximilian Augustin, Maximilian M\"uller, Mehdi Bouaziz,
Melina Jingting binti Laimon, Mert Unsal, Mia Chiquier, Michael Pilcer, Micha\l{} Pietruszka, Micha\l{}
Zaj\k{a}c, Micka\"el Seznec, Mikhail Biriuchinskii, Minh-Quang Pham, Minwoo Kang, Mircea Lica, Morgane
Rivi\`ere, Namit Katariya, Nathan Bout, Nathan Grinsztajn, Nathan Simpson, Neeraj Aggarwal, Neil Ashton, Neil
Mallinar, Nick B\"uhrer, Nicolas Moreau, Nicol\`o Lucchesi, Noh\'e Hinniger-Foray, Ola Mysiak, Oliver Leicht,
Olivier Bousquet, Olivier Koch, Parag Jain, Patricia Wang, Patrick Blies, Patrick von Platen, Patryk Saffer,
Paul Jacob, Paul Teiletche, Paul Wambergue, Paula Kurylowicz, Pavankumar Reddy Muddireddy, Pavel Kuksa,
Philippe Formont, Philippe Pinel, Philom\`ene Chagniot, Pierre Stock, Pierre-Andr\'e Savalle, Piotr
Mi\l{}o\'s, Poojan Shah, Prateek Gupta, Pravesh Agrawal, Quentin Desreumaux, Quentin Selle, Quentin Torroba,
Quentin Wargnier, Quercus Hernandez, Ram Ramrakhya, Randall Isenhour, Ranjit Parva, Rapha\"el Balouka, Raphael
Rialland, Raul Perez Pelaez, Reinhard Sonnleitner, R\'emi Delacourt, Richard Kurle, Rishi Shah, Rob
Romijnders, Rohan Shah, Rohin Arora, Romain R\'eocreux, Romain Sauvestre, Roman Soletskyi, Ronan Riochet,
Rosalie Millner, Samuel Barry, Samuel Belkadi, Samuel Humeau, Samuele Papa, Sanchit Gandhi, Sandeep
Subramanian, Sandeep Suresh Cranganore, Sarthak Mittal, Saskia Marie Adaime, Scott Matthews, Sebastian
Kaltenbach, Seungjae Jung, Shashwat Verma, Sherif Waly, Shrimai Prabhumoye, Shubham Gupta, Siddhant Ramesh
Waghjale, Siddharth Gandhi, Simon Lepage, Siva Reddy Goli, Soham Ghosh, Sophie Marbach, Srijan Mishra,
Stanislas Lange, Steve Hong, Sumukh Aithal, Szymon Antoniak, Tarun Kumar Vangani, Teddy Turquet, Teven Le
Scao, Th\'eo Cachet, Theo Simon Sorg, Thibaut Lavril, Thomas Chabal, Thomas Coste, Thomas Defard, Thomas
Foubert, Thomas Loux, Thomas Robert, Thomas Wang, Tiancheng Chen, Tim Lawson, Tim Wallis, Tobias Kronlachner,
Tom Bewley, Tom Edwards, Triantafyllos Afouras, Tuhin Das, Tyler Wang, Udit Saxena, Ulrick Bl\'e, Umar Jamil,
Valentin Mac\'e, Van Phung, Vibha Masti, Victor Jouault, Victor Letzelter, Victor Paltz, Victor Poucheret,
Vincent Maladi\`ere, Vincent Pfister, Virgile Richard, Vladislav Bataev, Wassim Bouaziz, Wen Ding Li, Will
Flanagan, William Havard, William Marshall, Witold J\k{e}drzejewski, Xiaofei Luo, Xing Han Lu, Xinghui Li,
Xingran Guo, Xinyu Yang, Xuanyu Zhang, Yann Dreze, Yann Viegas, Yannic Neuhaus, Yassine El Ouahidi, Yassine
Turki, Yassir Bendou, Yessin Moakher, Yi He, Yi Xu, Yihan Wang, Yimu Pan, Yuying Zhu, Yves Martin, Zaccharie
Ramzi, Zhenlin Xu, Zs\'ofia Cs\'ak\'any\par}

\bibliographystyle{plainnat}
\bibliography{references}

\newpage
\appendix
\crefalias{section}{appendix}\crefalias{subsection}{subappendix}\crefalias{subsubsection}{subsubappendix}
\section{Evaluation details}
\label{app:eval}

\subsection{Benchmarks}
\label{app:benchmarks}
\emph{Image classification} (18 datasets).
Natural images: ImageNet~\citep{deng2009imagenet}, ImageNet-V2~\citep{recht2019imagenetv2},
ImageNet-A~\citep{hendrycks2021imageneta}, ImageNet-R~\citep{hendrycks2021imagenetr},
ImageNet-Sketch~\citep{wang2019imagenetsketch}, ObjectNet~\citep{barbu2019objectnet} and Dollar
Street~\citep{gaviria2022dollarstreet}.
Fine-grained: DTD~\citep{cimpoi2014dtd}, Flowers102~\citep{nilsback2008flowers},
FGVC-Aircraft~\citep{maji2013aircraft}, Oxford Pets~\citep{parkhi2012pets},
Food101~\citep{bossard2014food101}, Country211~\citep{radford2021clip}, SUN397~\citep{xiao2010sun} and
RESISC45~\citep{cheng2017resisc45}.
People \& vehicles:
Stanford Cars~\citep{krause2013cars}, Stanford40~\citep{yao2011stanford40} and
PA-100K~\citep{liu2017hydraplus} (mAP over 26 attributes, one yes/no prompt pair per attribute).

\emph{Image retrieval.} COCO 5k Karpathy test~\citep{lin2014coco,chen2015cococaptions},
Flickr30k~\citep{young2014flickr30k}, TextCaps~\citep{sidorov2020textcaps}, DOCCI~\citep{onoe2024docci},
XM3600~\citep{thapliyal2022xm3600} (mean over 36 languages) and VisualNews~\citep{liu2021visualnews} in its
MMEB evaluation version~\citep{jiang2025vlm2vec}.

\emph{Video classification.} Kinetics-400/600/700~\citep{kay2017kinetics}, UCF101~\citep{soomro2012ucf101}
(mean of the three splits), HMDB51~\citep{kuehne2011hmdb} (split~1, 1{,}530 test clips),
SSv2~\citep{goyal2017ssv2}, EPIC-Kitchens-100 actions~\citep{damen2022epic} and
Breakfast~\citep{kuehne2014breakfast}.

\emph{Video retrieval.} MSR-VTT~\citep{xu2016msrvtt}, MSVD~\citep{chen2011msvd}, ActivityNet
Captions~\citep{krishna2017activitynet} (paragraph-to-video), PVD~\citep{bolya2025perception} and
DiDeMo~\citep{hendricks2017didemo} (1{,}004 test videos, paragraph queries describing several moments).

\subsection{Retrieval protocol details}
\label{app:retrieval_protocol}
Several MSVD videos share identical captions.
As in the PE evaluation code, we count any of them as a correct text-to-video retrieval, which inflates MSVD
t2v R@1 for all models.
ActivityNet is evaluated with 16 frames for all models, whereas \citet{bolya2025perception} report PE with 32
frames.
For MSR-VTT we use the full test split of \citet{xu2016msrvtt} (2{,}990 videos with 20 captions each) instead
of the 1k-A subset used in the
\href{https://github.com/facebookresearch/perception_models/blob/3e352cca660658d4b5c90f42a7808b11469e4c66/apps/pe/clip_benchmark/datasets/builder.py}{released
PE evaluation code}.
From the source-video URLs and clip timestamps, we find that 52 of the 1k-A test clips overlap in time with
clips in the 1k-A training partition, whereas the original test split shares no source-video URLs with its own
training partition.

\subsection{Baseline protocol}
\label{app:baseline_protocol}
\paragraph{Text preprocessing.}
Each model's text goes through its own default or official preprocessing and tokenizer (e.g.\ C-RADIOv4's
adaptor lower-cases the text and strips punctuation, whereas SigLIP2, PE-Core and TIPSv2 only lower-case it).

\paragraph{Video pooling and TIPSv2 embedding choice.}
For PE-Core, SigLIP2, C-RADIOv4, TIPSv2 and MetaCLIP~2, the video embedding is the mean of L2-normalized frame
embeddings.
PE-Core used this pooling during video training, and the other models have no video-specific encoder.
TIPSv2 trains two global CLS embeddings against different caption sources: web text for the first and
PaliGemma/Gemini synthetic captions for the second (\citealp{cao2026tipsv2}, App.~A.3).
For each model and benchmark, we select the CLS embedding with the higher best-resolution score.
For retrieval, this score is the average of the two directions' best-resolution scores.
We use the selected embedding for every resolution and transform of the benchmark.
For \vstarcls{}, the second CLS embedding has the higher mean across full-image and oracle-crop accuracy for
both TIPSv2 models, so we report it in both conditions.

\paragraph{Resolution and position embeddings.}
When the evaluation resolution differs from the training resolution, each model resizes its position
embeddings with its official implementation.
We do not impose a common interpolation, so the method differs across models.
Learned absolute embeddings are interpolated to the new patch grid: bilinearly for PE-Core, SigLIP2 NaFlex,
TIPSv2 and VideoPrism (the latter two with antialiasing), bicubically for SigLIP2-g and MetaCLIP~2, and
through the CPE interpolation for C-RADIOv4.
PE-Core additionally recomputes its 2D RoPE for the new grid, and Qwen3-VL-Embedding and EmbeddingGemma~2
encode positions natively at any resolution, so nothing is interpolated.
SigLIP2 NaFlex resizes each image to fit a patch budget.
We set this budget to the full $(\text{res}/16)^2$ grid rather than to its 1{,}024-patch training maximum, so
no input is downscaled.
At 672 and 1036\,px, however, it then processes 1{,}764 and ${\sim}$4{,}200 patches, beyond the sequence
lengths seen in training.
Where the patch grid does not divide the nominal resolution, we round the image side to a multiple of the
patch size: C-RADIOv4 to 16\,px (1036$\to$1024), VideoPrism down to 18\,px (224$\to$216) and EmbeddingGemma~2
to 48\,px (224$\to$240).

\subsubsection{Instructions for VLM embedders}
\label{app:qwen_prompts}

VLM embedders accept a text instruction that describes the task as part of the input, which can improve
performance.
We call a setting ``tuned'' when it uses task-specific instructions on the query side, and ``neutral'' when it
keeps the default instruction, if one is provided.

The default instruction of Qwen3-VL-Embedding (2B and 8B) is ``Represent the user's input.'' For the tuned
setting, the published MMEB/MMEB-V2 instructions cover ImageNet, Country211, SUN397, ObjectNet, COCO,
Flickr30k, VisualNews, Kinetics-700, UCF101, HMDB51, SSv2, Breakfast, MSR-VTT, MSVD and
DiDeMo~\citep{jiang2025vlm2vec,meng2026vlm2vecv2,li2026qwen3vlembedding}.
We reuse published instructions for ImageNet-V2/A/R/Sketch and Dollar Street (from ImageNet), TextCaps, DOCCI
and XM3600 (from COCO/Flickr30k), Kinetics-400/600 (from Kinetics-700) and ActivityNet (from DiDeMo).
We write instructions in the same style for FGVC-Aircraft, Stanford Cars, Food101, Oxford Pets, DTD,
Flowers102, RESISC45, Stanford40, PA-100K, EPIC-Kitchens-100 and PVD, and for video-to-text retrieval.
For the 2B model, tuned instructions improve all four group averages, whereas for the 8B model the neutral
instruction is better on video retrieval (\cref{tab:instruction_variants}). \cref{tab:qwen_instr} lists the
tuned query instructions.

EmbeddingGemma~2~\citep{google2026embeddinggemma2} takes no free-form instructions.
Its ``tuned'' variant uses the fixed task prefixes of its model card (``\texttt{task: search result |
query:}'' for queries, ``\texttt{title: none | text:}'' for candidate captions and ``\texttt{task:
classification | query:}'' for class prompts; image and video inputs contain modality markers but no
task-specific text prefix), and its ``neutral'' variant uses ``\texttt{title: none | text:}'' on every text.
\begin{table}[tbp]
\centering
\caption{\textbf{Effect of text instructions on the four task groups.} Mean zero-shot score (R@1 for retrieval, top-1 for classification), taking each dataset at its best evaluated input resolution. Tuned uses the model's task instructions; neutral uses its default instruction. Bold marks the higher setting for each model and task group.}
\label{tab:instruction_variants}
\footnotesize
\setlength{\tabcolsep}{5pt}
\begin{tabular}{@{}l l cccc@{}}
\toprule
Model & Instruction & Image ret. & Image cls. & Video ret. & Video cls. \\
\midrule
EmbeddingGemma~2 & neutral & 73.8 & \textbf{69.8} & 59.4 & \textbf{50.1} \\
EmbeddingGemma~2 & tuned & \textbf{75.8} & 69.0 & \textbf{61.0} & 49.9 \\
Qwen3-VL-Emb-2B & neutral & 74.4 & 65.2 & 61.2 & 54.7 \\
Qwen3-VL-Emb-2B & tuned & \textbf{77.5} & \textbf{69.5} & \textbf{61.9} & \textbf{57.8} \\
\modelseparator
Qwen3-VL-Emb-8B & neutral & 80.6 & 69.8 & \textbf{66.8} & 58.2 \\
Qwen3-VL-Emb-8B & tuned & \textbf{82.7} & \textbf{72.9} & 66.2 & \textbf{61.7} \\
\bottomrule
\end{tabular}
\end{table}

\begin{table}[tbp]
\centering
\caption{\textbf{Tuned query instructions of Qwen3-VL-Embedding.} Candidates always use ``Represent the user's
input.''}
\label{tab:qwen_instr}
\footnotesize
\begin{tabular}{@{}p{0.3\linewidth} p{0.66\linewidth}@{}}
\toprule
Benchmark & Query instruction \\
\midrule
ImageNet family, Dollar Street & Represent the given image for classification \\
Country211 / SUN397 / ObjectNet & Identify the country depicted in the image / the scene shown / the object
shown \\
Aircraft, Cars, Food101, Pets, DTD, Flowers102 & Identify the \{aircraft model / car model / food / pet breed
/ texture / flower species\} shown in the image \\
RESISC45 & Identify the scene shown in the remote sensing image \\
Stanford40 / PA-100K & Identify the action shown in the image / Identify the visual attributes of the person
shown in the image \\
Image retrieval t2i / i2t & Find me an everyday image that matches the given caption: / Find an image caption
describing the given everyday image.
\\
VisualNews t2i / i2t (MMEB) & Retrieve an image of this news caption. / Find a caption for the news in the
given photo. \\
Kinetics-400/600/700 & Recognize the category of the video content. \\
UCF101 & What activities or sports are being performed by the person in the video? \\
HMDB51 & What actions or objects interactions are the person in the video doing? \\
SSv2 & What actions or object interactions are being performed by the person in the video? \\
EPIC-Kitchens-100 & What action and object interaction is the camera wearer performing in the video? \\
Breakfast & Recognize the breakfast type that the person is cooking in the video. \\
MSR-VTT / MSVD / ActivityNet / PVD (t2v) & Find a video that contains the following visual content. / Find the
video snippet that corresponds to the given summary. / Find a video that includes the following described
scenes. / Find the video that matches the following description.
\\
DiDeMo (t2v, MMEB-V2) & Find a video that includes the following described scenes. \\
Video retrieval v2t & Find a video caption describing the given video. \\
\vstarcls{} (image) & Identify the visual attributes and relative positions of the objects shown in the image \\
\bottomrule
\end{tabular}
\end{table}

\subsection{Classification prompt templates}
\label{app:templates}
As is standard~\citep{radford2021clip}, we insert each class label into several prompt templates, average the
resulting text embeddings, L2-normalize the mean, and compare it with each image or video embedding by cosine
similarity.

All models use the same prompt ensembles, listed in full in
\crefrange{tab:templates_imagenet}{tab:templates_pa100k}.
We take them from the CLIP-benchmark suite as used by \citet{bolya2025perception}: the 80 templates of
\citet{radford2021clip} for the ImageNet family, ObjectNet and Dollar Street (\cref{tab:templates_imagenet}),
and per-dataset sets for the other image datasets and for video (\cref{tab:templates_other}).
Breakfast has no standard zero-shot prompts, so we write six templates for it and map its raw labels (e.g.\
\texttt{friedegg}, \texttt{milk}) to the activity names of the dataset (e.g.\ ``a fried egg'', ``chocolate
milk'').
For each of its 26 attributes, PA-100K uses a positive/negative prompt pair (\cref{tab:templates_pa100k})
scored with a two-way softmax, and we report mAP over attributes.

\begin{table}[p]
\centering\scriptsize
\caption{The 80 prompt templates used for ImageNet, ImageNet-V2/A/R/Sketch, ObjectNet and Dollar Street (\{\}
is replaced by the class name).}
\label{tab:templates_imagenet}
\begin{tabular}{@{}rl@{\hspace{2em}}rl@{}}
\toprule
1 & a bad photo of a \{\}. & 41 & the origami \{\}. \\
2 & a photo of many \{\}. & 42 & the \{\} in a video game. \\
3 & a sculpture of a \{\}. & 43 & a sketch of a \{\}. \\
4 & a photo of the hard to see \{\}. & 44 & a doodle of the \{\}. \\
5 & a low resolution photo of the \{\}. & 45 & a origami \{\}. \\
6 & a rendering of a \{\}. & 46 & a low resolution photo of a \{\}. \\
7 & graffiti of a \{\}. & 47 & the toy \{\}. \\
8 & a bad photo of the \{\}. & 48 & a rendition of the \{\}. \\
9 & a cropped photo of the \{\}. & 49 & a photo of the clean \{\}. \\
10 & a tattoo of a \{\}. & 50 & a photo of a large \{\}. \\
11 & the embroidered \{\}. & 51 & a rendition of a \{\}. \\
12 & a photo of a hard to see \{\}. & 52 & a photo of a nice \{\}. \\
13 & a bright photo of a \{\}. & 53 & a photo of a weird \{\}. \\
14 & a photo of a clean \{\}. & 54 & a blurry photo of a \{\}. \\
15 & a photo of a dirty \{\}. & 55 & a cartoon \{\}. \\
16 & a dark photo of the \{\}. & 56 & art of a \{\}. \\
17 & a drawing of a \{\}. & 57 & a sketch of the \{\}. \\
18 & a photo of my \{\}. & 58 & a embroidered \{\}. \\
19 & the plastic \{\}. & 59 & a pixelated photo of a \{\}. \\
20 & a photo of the cool \{\}. & 60 & itap of the \{\}. \\
21 & a close-up photo of a \{\}. & 61 & a jpeg corrupted photo of the \{\}. \\
22 & a black and white photo of the \{\}. & 62 & a good photo of a \{\}. \\
23 & a painting of the \{\}. & 63 & a plushie \{\}. \\
24 & a painting of a \{\}. & 64 & a photo of the nice \{\}. \\
25 & a pixelated photo of the \{\}. & 65 & a photo of the small \{\}. \\
26 & a sculpture of the \{\}. & 66 & a photo of the weird \{\}. \\
27 & a bright photo of the \{\}. & 67 & the cartoon \{\}. \\
28 & a cropped photo of a \{\}. & 68 & art of the \{\}. \\
29 & a plastic \{\}. & 69 & a drawing of the \{\}. \\
30 & a photo of the dirty \{\}. & 70 & a photo of the large \{\}. \\
31 & a jpeg corrupted photo of a \{\}. & 71 & a black and white photo of a \{\}. \\
32 & a blurry photo of the \{\}. & 72 & the plushie \{\}. \\
33 & a photo of the \{\}. & 73 & a dark photo of a \{\}. \\
34 & a good photo of the \{\}. & 74 & itap of a \{\}. \\
35 & a rendering of the \{\}. & 75 & graffiti of the \{\}. \\
36 & a \{\} in a video game. & 76 & a toy \{\}. \\
37 & a photo of one \{\}. & 77 & itap of my \{\}. \\
38 & a doodle of a \{\}. & 78 & a photo of a cool \{\}. \\
39 & a close-up photo of the \{\}. & 79 & a photo of a small \{\}. \\
40 & a photo of a \{\}. & 80 & a tattoo of the \{\}. \\
\bottomrule
\end{tabular}
\end{table}

\begin{table}[p]
\centering\scriptsize
\caption{Prompt templates for the other classification datasets, separated by \textbar{} (\{\} is replaced by
the class name; $N$: number of templates). The ImageNet family is in \cref{tab:templates_imagenet},
PA-100K in \cref{tab:templates_pa100k}.}
\label{tab:templates_other}
\begin{tabular}{@{}lr>{\raggedright\arraybackslash}p{0.75\textwidth}@{}}
\toprule
Dataset & $N$ & Templates \\
\midrule
\multicolumn{3}{@{}l}{\emph{Image}} \\
DTD & 8 & a photo of a \{\} texture. \textbar{} a photo of a \{\} pattern. \textbar{} a photo of a \{\} thing.
\textbar{} a photo of a \{\} object. \textbar{} a photo of the \{\} texture. \textbar{} a photo of the \{\}
pattern. \textbar{} a photo of the \{\} thing. \textbar{} a photo of the \{\} object. \\
Flowers102 & 1 & a photo of a \{\}, a type of flower. \\
FGVC-Aircraft & 2 & a photo of a \{\}, a type of aircraft. \textbar{} a photo of the \{\}, a type of aircraft.
\\
Oxford Pets & 1 & a photo of a \{\}, a type of pet. \\
Food101 & 1 & a photo of \{\}, a type of food. \\
Country211 & 5 & a photo i took in \{\}. \textbar{} a photo i took while visiting \{\}. \textbar{} a photo
from my home country of \{\}. \textbar{} a photo from my visit to \{\}. \textbar{} a photo showing the country
of \{\}. \\
SUN397 & 2 & a photo of a \{\}. \textbar{} a photo of the \{\}. \\
RESISC45 & 4 & a satellite image of \{\} \textbar{} an aerial view of \{\} \textbar{} a satellite photo of
\{\} \textbar{} \{\} from above \\
Stanford Cars & 8 & a photo of a \{\}. \textbar{} a photo of the \{\}. \textbar{} a photo of my \{\}.
\textbar{} i love my \{\}! \textbar{} a photo of my dirty \{\}. \textbar{} a photo of my clean \{\}.
\textbar{} a photo of my new \{\}. \textbar{} a photo of my old \{\}. \\
Stanford40 & 3 & a photo of a person \{\}. \textbar{} a photo of a person who is \{\}. \textbar{} a photo of
someone \{\}. \\
\midrule
\multicolumn{3}{@{}l}{\emph{Video}} \\
Kinetics-400/600/700 & 28 & a photo of \{\}. \textbar{} a photo of a person \{\}. \textbar{} a photo of a
person using \{\}. \textbar{} a photo of a person doing \{\}. \textbar{} a photo of a person during \{\}.
\textbar{} a photo of a person performing \{\}. \textbar{} a photo of a person practicing \{\}. \textbar{} a
video of \{\}. \textbar{} a video of a person \{\}. \textbar{} a video of a person using \{\}. \textbar{} a
video of a person doing \{\}. \textbar{} a video of a person during \{\}. \textbar{} a video of a person
performing \{\}. \textbar{} a video of a person practicing \{\}. \textbar{} a example of \{\}. \textbar{} a
example of a person \{\}. \textbar{} a example of a person using \{\}. \textbar{} a example of a person doing
\{\}. \textbar{} a example of a person during \{\}. \textbar{} a example of a person performing \{\}.
\textbar{} a example of a person practicing \{\}. \textbar{} a demonstration of \{\}. \textbar{} a
demonstration of a person \{\}. \textbar{} a demonstration of a person using \{\}. \textbar{} a demonstration
of a person doing \{\}. \textbar{} a demonstration of a person during \{\}. \textbar{} a demonstration of a
person performing \{\}. \textbar{} a demonstration of a person practicing \{\}. \\
UCF101 & 48 & a photo of a person \{\}. \textbar{} a video of a person \{\}. \textbar{} a example of a person
\{\}. \textbar{} a demonstration of a person \{\}. \textbar{} a photo of the person \{\}. \textbar{} a video
of the person \{\}. \textbar{} a example of the person \{\}. \textbar{} a demonstration of the person \{\}.
\textbar{} a photo of a person using \{\}. \textbar{} a video of a person using \{\}. \textbar{} a example of
a person using \{\}. \textbar{} a demonstration of a person using \{\}. \textbar{} a photo of the person using
\{\}. \textbar{} a video of the person using \{\}. \textbar{} a example of the person using \{\}. \textbar{} a
demonstration of the person using \{\}. \textbar{} a photo of a person doing \{\}. \textbar{} a video of a
person doing \{\}. \textbar{} a example of a person doing \{\}. \textbar{} a demonstration of a person doing
\{\}. \textbar{} a photo of the person doing \{\}. \textbar{} a video of the person doing \{\}. \textbar{} a
example of the person doing \{\}. \textbar{} a demonstration of the person doing \{\}. \textbar{} a photo of a
person during \{\}. \textbar{} a video of a person during \{\}. \textbar{} a example of a person during \{\}.
\textbar{} a demonstration of a person during \{\}. \textbar{} a photo of the person during \{\}. \textbar{} a
video of the person during \{\}. \textbar{} a example of the person during \{\}. \textbar{} a demonstration of
the person during \{\}. \textbar{} a photo of a person performing \{\}. \textbar{} a video of a person
performing \{\}. \textbar{} a example of a person performing \{\}. \textbar{} a demonstration of a person
performing \{\}. \textbar{} a photo of the person performing \{\}. \textbar{} a video of the person performing
\{\}. \textbar{} a example of the person performing \{\}. \textbar{} a demonstration of the person performing
\{\}. \textbar{} a photo of a person practicing \{\}. \textbar{} a video of a person practicing \{\}.
\textbar{} a example of a person practicing \{\}. \textbar{} a demonstration of a person practicing \{\}.
\textbar{} a photo of the person practicing \{\}. \textbar{} a video of the person practicing \{\}. \textbar{}
a example of the person practicing \{\}. \textbar{} a demonstration of the person practicing \{\}. \\
HMDB51 & 24 & a photo of a person \{\}. \textbar{} a video of a person \{\}. \textbar{} a example of a person
\{\}. \textbar{} a demonstration of a person \{\}. \textbar{} a photo of the person \{\}. \textbar{} a video
of the person \{\}. \textbar{} a example of the person \{\}. \textbar{} a demonstration of the person \{\}.
\textbar{} a photo of a person doing \{\}. \textbar{} a video of a person doing \{\}. \textbar{} a example of
a person doing \{\}. \textbar{} a demonstration of a person doing \{\}. \textbar{} a photo of the person doing
\{\}. \textbar{} a video of the person doing \{\}. \textbar{} a example of the person doing \{\}. \textbar{} a
demonstration of the person doing \{\}. \textbar{} a photo of a person during \{\}. \textbar{} a video of a
person during \{\}. \textbar{} a example of a person during \{\}. \textbar{} a demonstration of a person
during \{\}. \textbar{} a photo of the person during \{\}. \textbar{} a video of the person during \{\}.
\textbar{} a example of the person during \{\}. \textbar{} a demonstration of the person during \{\}. \\
SSv2 & 5 & a video of \{\}. \textbar{} a video of a person \{\}. \textbar{} a person \{\}. \textbar{} a
demonstration of \{\}. \textbar{} \{\}. \\
EPIC-Kitchens-100 & 7 & a photo showing the action \{\}. \textbar{} a video showing the action \{\}.
\textbar{} an example of the action \{\}. \textbar{} a demonstration of the action \{\}. \textbar{} a
first-person video showing the action \{\}. \textbar{} an egocentric video showing the action \{\}. \textbar{}
\{\}. \\
Breakfast & 6 & a video of a person making \{\}. \textbar{} a video of a person preparing \{\}. \textbar{} a
photo of a person making \{\}. \textbar{} a video of someone making \{\} in a kitchen. \textbar{} a person
preparing \{\} for breakfast. \textbar{} a demonstration of a person making \{\}. \\
\bottomrule
\end{tabular}
\end{table}

\begin{table}[p]
\centering\scriptsize
\caption{PA-100K prompt pairs: each attribute is scored by a two-way softmax over its positive and negative
prompt, and we report mAP over the 26 attributes.}
\label{tab:templates_pa100k}
\begin{tabular}{@{}l>{\raggedright\arraybackslash}p{0.38\textwidth}%
  >{\raggedright\arraybackslash}p{0.38\textwidth}@{}}
\toprule
Attribute & Positive prompt & Negative prompt \\
\midrule
Female & a photo of a woman & a photo of a man \\
AgeOver60 & a photo of an elderly person over 60 years old & a photo of a person under 60 years old \\
Age18-60 & a photo of an adult between 18 and 60 years old & a photo of a child or an elderly person \\
AgeLess18 & a photo of a child under 18 years old & a photo of an adult \\
Front & a photo of a person seen from the front & a photo of a person seen from the side or back \\
Side & a photo of a person seen from the side & a photo of a person seen from the front or back \\
Back & a photo of a person seen from behind & a photo of a person seen from the front or side \\
Hat & a photo of a person wearing a hat & a photo of a person not wearing a hat \\
Glasses & a photo of a person wearing glasses & a photo of a person not wearing glasses \\
HandBag & a photo of a person carrying a handbag & a photo of a person not carrying a handbag \\
ShoulderBag & a photo of a person carrying a shoulder bag & a photo of a person not carrying a shoulder bag \\
Backpack & a photo of a person carrying a backpack & a photo of a person not carrying a backpack \\
HoldObjectsInFront & a photo of a person holding objects in front of them & a photo of a person not holding
objects in front of them \\
ShortSleeve & a photo of a person wearing short sleeves & a photo of a person wearing long sleeves \\
LongSleeve & a photo of a person wearing long sleeves & a photo of a person wearing short sleeves \\
UpperStride & a photo of a person wearing upper clothing with a stride pattern & a photo of a person wearing
plain upper clothing \\
UpperLogo & a photo of a person wearing upper clothing with a logo & a photo of a person wearing upper
clothing without a logo \\
UpperPlaid & a photo of a person wearing plaid upper clothing & a photo of a person wearing non-plaid upper
clothing \\
UpperSplice & a photo of a person wearing spliced upper clothing of multiple colors & a photo of a person
wearing upper clothing of a single color \\
LowerStripe & a photo of a person wearing striped lower clothing & a photo of a person wearing non-striped
lower clothing \\
LowerPattern & a photo of a person wearing patterned lower clothing & a photo of a person wearing plain lower
clothing \\
LongCoat & a photo of a person wearing a long coat & a photo of a person not wearing a long coat \\
Trousers & a photo of a person wearing trousers & a photo of a person not wearing trousers \\
Shorts & a photo of a person wearing shorts & a photo of a person not wearing shorts \\
Skirt\&Dress & a photo of a person wearing a skirt or a dress & a photo of a person not wearing a skirt or a
dress \\
boots & a photo of a person wearing boots & a photo of a person not wearing boots \\
\bottomrule
\end{tabular}
\end{table}

\FloatBarrier

\subsection{Effect of a short text context}
\label{app:short_context}
\begin{figure}[tbp]
\centering
\includegraphics[width=\linewidth]{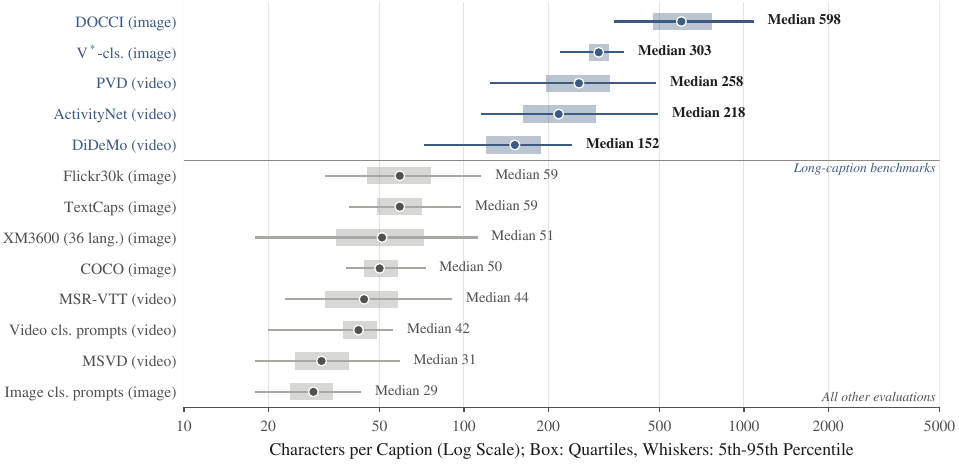}
\caption{\textbf{Characters per caption in each evaluation.} Characters per caption or class prompt; box:
quartiles, whiskers: 5th--95th percentiles, dot and label: median; log x-axis; XM3600 pools its 36 languages.
Top, long captions: DOCCI (median 598 characters), \vstarcls{} (303), PVD (258), ActivityNet (218) and DiDeMo
(152).
Bottom, all other evaluations: medians of 29--59 characters; the closest pair is DiDeMo against Flickr30k and
TextCaps (152 vs.\ 59). \vstarcls{} is evaluated separately as a classification benchmark
(\cref{sec:results_vstar}), and the other four in \cref{tab:longcaption}.}
\label{fig:caption_words}
\end{figure}

As motivated in the introduction (\cref{sec:intro}, property (iii)), we assess how well models handle long
captions on a dedicated suite of long-caption benchmarks.
The caption lengths of \cref{fig:caption_words} define the suite: DOCCI, \vstarcls{}, PVD, ActivityNet and
DiDeMo have a median caption longer than 100 characters, against 29--59 characters for every other evaluation.

On these benchmarks, we also measure the effect of truncation caused by a short text context.
Dual encoders such as PE-Core~\citep{bolya2025perception}, SigLIP2~\citep{tschannen2025siglip2} and
C-RADIOv4~\citep{ranzinger2026cradiov4} truncate texts that exceed the context of their text encoder.
We evaluate the retrieval performance of all models on these benchmarks in \cref{sec:long_captions}
(\vstarcls{} in \cref{sec:results_vstar}). \cref{tab:truncation} reports, for each model, its text context and
the share of truncated texts.

\begin{table}[h]
\centering
\caption{\textbf{Text context length vs.\ share of truncated captions.} Share of evaluation texts (\%) longer than each model's fixed text context (if provided) in our evaluation, counted with the model's own tokenizer and text preprocessing, including special tokens and prompt or instruction prefixes. Columns: benchmarks where some model truncates at least 5\% of the texts (XM3600 pools its 36 languages); on every other benchmark (class prompts included), no model truncates more than \truncOtherMax{}\%. Context: maximum text tokens; --: no context limit (Context column) or not evaluated.}
\label{tab:truncation}
\small
\setlength{\tabcolsep}{4pt}
\begin{tabular}{@{}l r cccccc@{}}
\toprule
Model & Context & DOCCI & XM3600 & ActivityNet & PVD & DiDeMo & V$^*$-cls. \\
\midrule
\ours{} & -- & 0 & 0 & 0 & 0 & 0 & 0 \\
PE-Core-L & 32 & 100 & 43.8 & 87.2 & 91.4 & 55.8 & 100 \\
PE-Core-G & 72 & 97.0 & 15.8 & 18.8 & 25.7 & 0.4 & 11.4 \\
MetaCLIP~2-H & 77 & 95.5 & ${<}0.1$ & 16.5 & 21.7 & 0.2 & 6.7 \\
MetaCLIP~2-bigG & 77 & 95.5 & ${<}0.1$ & 16.5 & 21.7 & 0.2 & 6.7 \\
SigLIP2-So400m & 64 & 98.7 & 0.4 & 24.8 & 35.1 & 0.9 & 33.2 \\
SigLIP2-g & 64 & 98.7 & 0.4 & 24.8 & 35.1 & 0.9 & 33.2 \\
TIPSv2-L & 64 & 99.4 & 2.7 & 30.5 & 42.2 & 1.6 & 70.5 \\
TIPSv2-g & 64 & 99.4 & 2.7 & 30.5 & 42.2 & 1.6 & 70.5 \\
VideoPrism-LvT-L & 64 & -- & -- & 27.2 & 35.2 & 1.6 & -- \\
EmbeddingGemma~2 & 8192 & 0 & 0 & 0 & 0 & 0 & 0 \\
C-RADIOv4-So400m & 64 & 96.8 & 0.3 & 19.5 & 24.5 & 0.7 & 7.4 \\
C-RADIOv4-H & 64 & 96.8 & 0.3 & 19.5 & 24.5 & 0.7 & 7.4 \\
Qwen3-VL-Emb-2B & -- & 0 & 0 & 0 & 0 & 0 & 0 \\
Qwen3-VL-Emb-8B & -- & 0 & 0 & 0 & 0 & 0 & 0 \\
\bottomrule
\end{tabular}
\end{table}

\FloatBarrier
\subsection{\vstarcls{} details}
\label{app:vstar}

\begin{figure}[tbp]
\centering
\includegraphics[width=\linewidth]{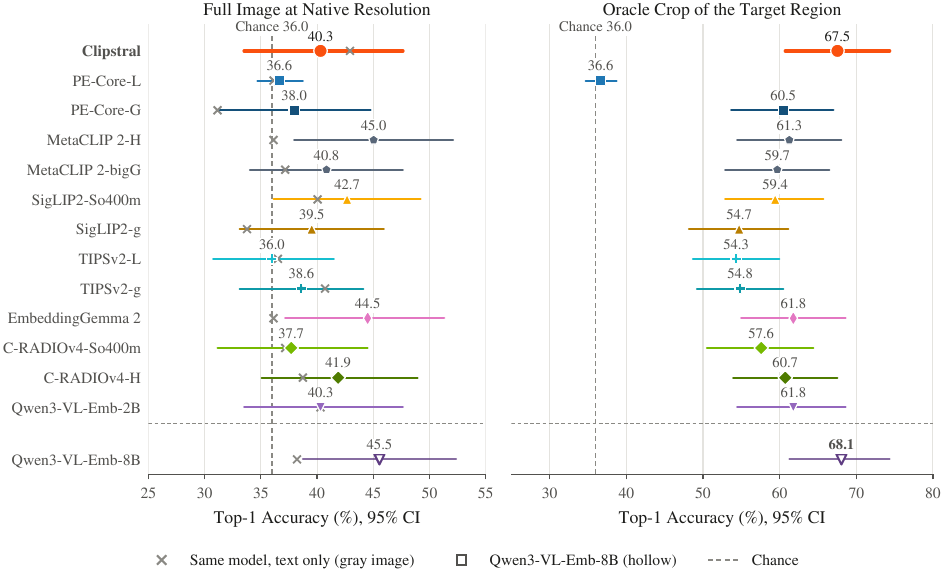}
\caption{\textbf{\vstarcls{} accuracy with the full image vs.\ the oracle crop.}
Top-1 on 191 items (random label among tied candidates) with 95\% bootstrap CIs; dashed line: chance.
Left: full image at native resolution (gray $\times$: uniform gray image, text only).
Right: oracle crop of the target at native pixel density.
Rows as in \cref{tab:vstar_main}; hollow marker: Qwen3-VL-Embedding-8B. Qwen3-VL-Embedding and
EmbeddingGemma~2 use tuned text settings (\cref{app:qwen_prompts}).}
\label{fig:vstar_cls}
\end{figure}

This appendix complements \cref{sec:vstar}, where we build \vstarcls{} by turning each of the 191 questions of
V$^*$Bench~\citep{wu2024vstar} into zero-shot classification over one positive and one to three hard-negative
scene captions that differ only in the answer phrase, and report the main results (\cref{tab:vstar_main}).
Here we give the evaluation protocol, the controls (oracle crop and gray image) and the full per-model results
(\cref{fig:vstar_cls}).

\paragraph{Protocol.} Each image is fed at its native resolution, resized only to the nearest multiple of each
model's patch grid.
We raise each model's maximum input size accordingly (5{,}768\,px for \ours{}, PE-Core, MetaCLIP~2 and TIPSv2;
5{,}760\,px for SigLIP2 (with a NaFlex sequence length of 32{,}768) and C-RADIOv4; 4{,}096 to 8{,}388{,}608
pixels for Qwen3-VL-Embedding; EmbeddingGemma~2 takes the image on its 48-px grid, uncapped).

On the text side, models with a short text context can truncate the answer phrase.
The candidate captions then become identical and receive identical text embeddings, and the predicted label is
sampled at random among them.
The share of truncated \vstarcls{} captions is given for each model in \cref{tab:truncation}.

\paragraph{Controls.} The first two items below are image controls: they change only the image, while
captions, candidate order and answers stay identical.
\begin{itemize}[leftmargin=*,itemsep=1pt,topsep=2pt]
\item \emph{Gray-image control}: to measure how much a model can answer from the candidate captions alone, we
      replace every image with a uniform mid-gray $336{\times}336$ image.
      The resulting accuracy reflects the model's own text bias among the candidates, without scene-specific
      visual evidence.
      On the gray-image control, models score close to chance (\cref{fig:vstar_cls}), as expected.
\item \emph{Oracle crop}: to give each model a simpler version of the task, in which the target no longer
      occupies a few pixels of a large image, we also feed it the target region alone.
      This region is the union of the target bounding boxes, padded by 50\% of its size on each side and grown
      to at least $336{\times}336$ (centered, clamped to the image), at native pixel density.
      For 11 spatial items, V$^*$Bench annotates only the subject of the relation (e.g.\ the dog in ``the dog
      is on
      the left side of the golden tower''), so this region can leave out the reference object; for these items
      we
      extend the crop to include the reference object, or its nearest identifiable part, plus a 25\,px margin.
      Crops range from $336{\times}336$ to $1890{\times}854$\,px.

\item \emph{Confidence intervals}: to assess the uncertainty of the measured accuracies, we draw 10{,}000
bootstrap
      resamples of the 191 items to obtain 95\% percentile intervals~\citep{efron1993bootstrap} for each
      model's accuracy.
      These intervals capture the uncertainty due to the sampling of the items.
\end{itemize}

\paragraph{Human baseline.}
Eight annotators solved \vstarcls{} with the same inputs as the models: each item shows the image and its two
or four candidate captions, and the annotator chooses one caption, without access to the answers.
Four annotators worked on the full native images, using an image viewer with zoom, and four others on the
oracle crops.
Mean accuracy is 98.6\% on full images and 99.6\% on oracle crops (\cref{tab:vstar_main}); the majority vote
is correct on every item in both settings, and all annotators agree on 180 (full image) and 188 (crop) of the
191 items.

\paragraph{Findings.}
On full native images, all models score close to chance (\vstarChance{}\%; \cref{fig:vstar_cls}), from
36.0 to 45.5\%: the 95\% intervals of all models contain chance, except those of MetaCLIP~2-H, SigLIP2-So400m,
EmbeddingGemma~2 and Qwen3-VL-Embedding-8B, whose lower bounds exceed it by at most 2.8 points.
With the oracle crop, every model except PE-Core-L scores far above chance.
\ours{} reaches \oursVstarCrop{}\%, the highest among models up to 4B parameters and on par with
Qwen3-VL-Embedding-8B (68.1\%).

\subsection{Dual-softmax scoring}
\label{app:dsl}
\citet{cheng2021dsl} introduced the dual softmax loss (DSL) to train video--text retrieval models: each
similarity is re-weighted by a softmax over the opposite retrieval direction.
The motivation is that a generic caption (e.g.\ ``a person in a room'') can score highest for many videos; its
text-to-video softmax is spread over them, so the re-weighting down-weights it relative to precise captions.
InternVL~\citep{chen2024internvl} applies the same re-weighting at evaluation time in its released
video-retrieval
code,\footnote{\url{https://github.com/OpenGVLab/InternVL/blob/2410d1dbf208f0e799459aff9376e5747dbf41a2/video_retrieval/test_msrvtt.py\#L97-L109}.}
although the InternVL paper does not describe it.
PE~\citep{bolya2025perception} follows InternVL and applies it to every model it evaluates (their App.~B.1.2):
\begin{equation}
s'(q, c) = s(q, c)\,\frac{\exp s(q, c)}{\sum_{q'} \exp s(q', c)},
\label{eq:dsl}
\end{equation}
where $s$ is the cosine similarity between query $q$ and candidate $c$, and the sum runs over all queries of
the test set.
We use the same form in the comparison below.
However, this softmax requires all queries of the test set, whereas in practice each retrieval query arrives
alone, so we chose to report raw cosine R@1 in all model comparisons.

\begin{figure}[tbp]
\centering
\includegraphics[width=\linewidth]{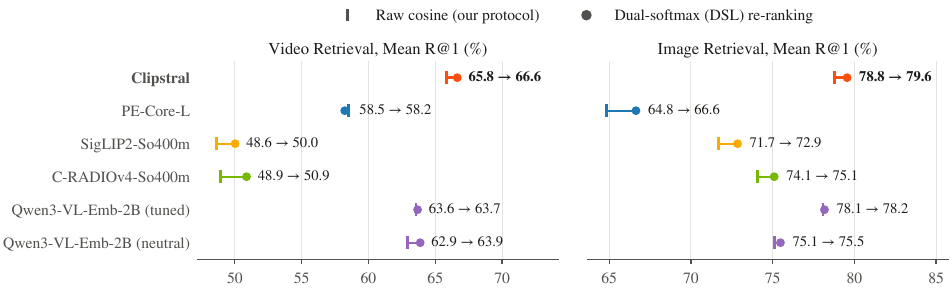}
\caption{\textbf{Mean R@1 with raw cosine vs.\ dual-softmax (DSL) scoring.}
Mean bidirectional R@1 under raw cosine (vertical tick, our protocol) and DSL (dot).
Video: MSR-VTT 1k-A, MSVD, ActivityNet, PVD; image: COCO, Flickr30k, TextCaps, DOCCI, XM3600.
Each metric is the best of 224--1036\,px and three transforms.}
\label{fig:app_dsl}
\end{figure}

\paragraph{Effect on our results.}
We compare raw cosine and DSL on the retrieval benchmarks for which DSL scores are available: MSR-VTT (1k-A
split), MSVD, ActivityNet and PVD for video; COCO, Flickr30k, TextCaps, DOCCI and XM3600 for images.
Each metric takes the best over 224, 336, 672 and 1036\,px and three input transforms, selected separately
under raw cosine and DSL. The comparison covers the size-matched models of the original comparison.
Because of the MSR-VTT split and the benchmark set, the raw-cosine means differ from those of
\cref{tab:eval_image_ret,tab:eval_video_ret}.
DSL changes the mean R@1 of every model by at most 2.0 points (largest change: C-RADIOv4 on video, +2.0;
\ours{} +0.8 on both video and image) and does not change the ranking (\cref{fig:app_dsl}).
Per benchmark, all changes are within $\pm$2.1 points except on XM3600, where DSL adds 7.9 points to PE-Core-L
and about 3 points to SigLIP2, C-RADIOv4 and \ours{}.

\FloatBarrier
\subsection{Encoding latency and throughput}
\label{app:speed}
Whether an embedding model can be deployed depends on its encoding cost as well as its accuracy: we evaluate
the response time for a single query (batch-1 latency), and the cost of indexing a larger collection (batched
throughput).
\begin{figure}[tbp]
\centering
\includegraphics[width=\linewidth]{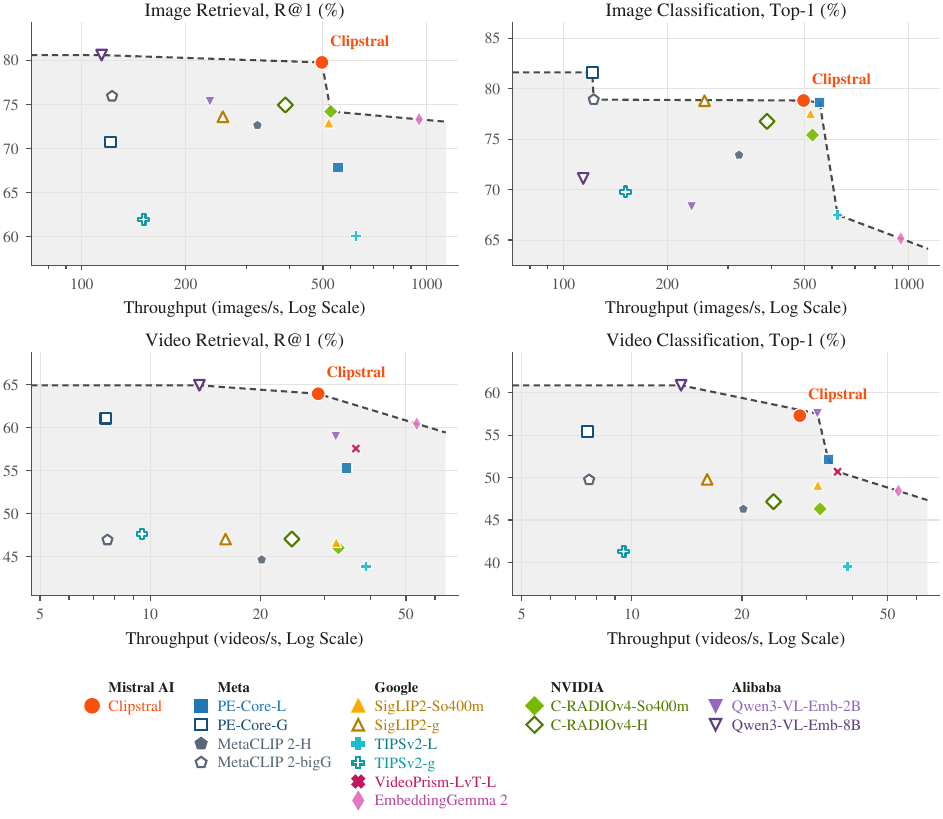}
\caption{\textbf{Performance vs.\ throughput.} As in \cref{fig:acc_latency}, but with batched encoding
throughput (images: batch 64; videos: batch 4, 16 frames; log scale; higher is better) instead of batch-1
latency.
The dashed black line marks the Pareto frontier.
Only in video classification is a model at least as accurate and faster than \ours{}: Qwen3-VL-Embedding-2B,
on par in accuracy.}
\label{fig:acc_speed}
\end{figure}

\paragraph{Setup.}
All models run on one NVIDIA H200 in bf16 with default \texttt{torch.compile} (graph breaks allowed). \ours{}
is timed with its released Hugging Face code and the baselines with their public model code.
Before timing, each model is compiled and warmed up, so the reported times exclude one-time compilation.
Some implementations compile only partially.
The measured graph breaks (image / video) are 27 / 37 for Qwen3-VL-Embedding (2B and 8B), 26 for TIPSv2-g, 13
for EmbeddingGemma~2 and 8 for SigLIP2-So400m; all other models, including \ours{}, compile without graph
breaks.
Qwen3-VL-Embedding uses its default SDPA attention, because its optional FlashAttention-2 path does not run
under \texttt{torch.compile} in transformers 4.57.3 without modifying the code.

\paragraph{What is timed.}
We time only the encode call.
Preprocessing (image transform or processor, chat template, patchification, host-to-device copy) and
tokenization are done before the clock starts.
The input images are synthetic $480{\times}640$ inputs, squash-resized to 448\,px and timed at batch size 1
(latency, \cref{fig:acc_latency}) and 64 (throughput, \cref{fig:acc_speed}).
Videos have 16 frames at 448\,px and are timed at batch sizes 1 and 4.

\paragraph{Results.}
At batch size 1 (\cref{fig:acc_latency}), \ours{} is on the accuracy--latency Pareto frontier in image and
video retrieval and classification: no model has both a lower latency and a higher accuracy.
With batched throughput (\cref{fig:acc_speed}), \ours{} remains on the Pareto frontier for image retrieval,
image classification and video retrieval; in video classification, Qwen3-VL-Embedding-2B is on par in accuracy
and slightly faster.
The two metrics rank models differently: batch-1 latency depends less on model size than on per-call overhead,
i.e., kernel launches and host--device synchronizations at \texttt{torch.compile} graph breaks.

\subsection{Additional comparison}
\label{app:jina_eva}
\cref{tab:jina_eva} compares \ours{} with three further open-weight image-only embedders evaluated under the
same protocol: jina-clip-v2~\citep{koukounas2024jinaclipv2}, a multilingual dual encoder with a long text
context, jina-embeddings-v5-omni-nano~\citep{honicke2026jinav5omni}, a small single-backbone omni embedder,
and EVA02-CLIP-E/14+~\citep{sun2023evaclip}, a 5B-parameter dual encoder.
For jina-embeddings-v5-omni-nano we count the weights used for retrieval, 0.32B (1.04B in total): the vision
tower, the text backbone and the retrieval adapter, excluding the audio tower and the classification,
clustering and text-matching adapters.
We use the retrieval adapter for all its tasks: zero-shot classification matches images against class prompts,
as retrieval does, and the model's classification adapter scores near chance in this setting
(2.7\% top-1 on a class-balanced ImageNet subset).

\ours{} scores highest in both image and video retrieval and classification.
The lead is largest on long-caption image retrieval (DOCCI), multilingual retrieval (XM3600) and also on
retrieval of specific people/places (VisualNews) with respect to the Jina models.
\begin{table}[H]
\centering
\caption{\textbf{\ours{} vs.\ additional baselines} (best resolution per dataset, as in
\crefrange{tab:eval_image_ret}{tab:eval_video_cls}). jina-embeddings-v5-omni-nano uses its model-card prompts.
Params: both towers, from the released weights.
Bold: best per column.}
\label{tab:jina_eva}
\small
\setlength{\tabcolsep}{3.5pt}
\resizebox{\linewidth}{!}{\begin{tabular}{@{}l r cccc c cccc@{}}
\toprule
 & & \multicolumn{4}{c}{Task group} & & \multicolumn{4}{c}{Image retrieval} \\
\cmidrule(lr){3-6}\cmidrule(l){8-11}
Model & Params & Img.\ ret. & Img.\ cls. & Vid.\ ret. & Vid.\ cls. & & Short & DOCCI & XM3600 & VisualNews \\
\midrule
\ours{} & 1.24B & \textbf{80.2} & \textbf{79.2} & \textbf{64.0} & \textbf{57.0} & & \textbf{80.5} & \textbf{92.7} & \textbf{59.7} & \textbf{87.5} \\
jina-clip-v2 & 0.87B & 64.3 & 61.7 & 35.5 & 33.5 & & 65.4 & 78.3 & 51.4 & 60.1 \\
jina-v5-omni-nano & 0.32B & 43.8 & 32.8 & 25.9 & 28.6 & & 52.6 & 45.8 & 22.8 & 36.1 \\
EVA02-CLIP-E/14+ & 5.04B & 68.0 & 76.2 & 45.7 & 46.8 & & 75.5 & 81.5 & 13.5 & 86.8 \\
\bottomrule
\end{tabular}
}
\end{table}

\FloatBarrier
\section{Per-resolution results}
\label{app:perres}
\begin{figure}[tbp]
\centering
\includegraphics[width=\linewidth]{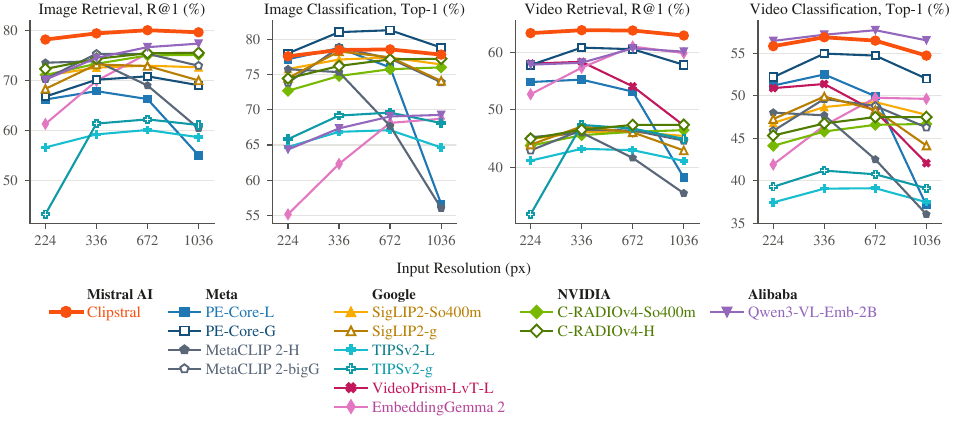}
\caption{\textbf{Expanded performance vs.\ input resolution.} The average score at each input resolution
(224, 336, 672 and 1036\,px) uses the best of three input transforms for every benchmark.
Hollow markers denote the larger model of a family.
Qwen3-VL-Embedding-2B and EmbeddingGemma~2 use tuned instructions. \cref{fig:resolution} shows a subset of
these models.}
\label{fig:resolution_all}
\end{figure}

\crefrange{tab:app_perres_image_cls}{tab:app_perres_video_ret} give, for every benchmark, each model's score
at each input size (best of three transforms) and the best-resolution value used in
\crefrange{tab:eval_image_ret}{tab:eval_video_cls}. \cref{fig:resolution_all} shows each model's average score
at each input size.

\begin{table}[!tbp]
\centering
\caption{\textbf{Image classification per input resolution.} Top-1 accuracy (mAP for PA-100K). Each cell is the best of three input transforms at that input resolution; Best: best resolution per dataset, as in the main tables. Bold: best, underlined: second best among models up to 4B parameters within each block. --: not available.}
\label{tab:app_perres_image_cls}
\scriptsize
\setlength{\tabcolsep}{2.0pt}
\resizebox{\linewidth}{!}{%
\begin{tabular}{@{}l r cccccccccccccccccc c@{}}
\toprule
 & & \multicolumn{7}{c}{Natural images} & \multicolumn{11}{c}{Fine-grained} & \\
\cmidrule(lr){3-9}\cmidrule(lr){10-20}
Model & Params & \rotatebox{90}{ImageNet} & \rotatebox{90}{ImageNet-V2} & \rotatebox{90}{ImageNet-A} & \rotatebox{90}{ImageNet-R} & \rotatebox{90}{ImageNet-Sketch} & \rotatebox{90}{ObjectNet} & \rotatebox{90}{Dollar St.} & \rotatebox{90}{DTD} & \rotatebox{90}{Flowers} & \rotatebox{90}{Aircraft} & \rotatebox{90}{Pets} & \rotatebox{90}{Food101} & \rotatebox{90}{Country211} & \rotatebox{90}{SUN397} & \rotatebox{90}{RESISC45} & \rotatebox{90}{Cars} & \rotatebox{90}{Stanford40} & \rotatebox{90}{PA-100K} & Avg. \\
\midrule
\multicolumn{21}{@{}l}{\textbf{224\,px}} \\
\addlinespace[1pt]
\textbf{\textcolor{accentorange}{\ours{}}} & 1.24B & \cellcolor{heat!41}81.3 & \cellcolor{heat!45}\textbf{79.0} & \cellcolor{heat!43}81.1 & \cellcolor{heat!39}93.0 & \cellcolor{heat!39}71.8 & \cellcolor{heat!45}\underline{81.3} & \cellcolor{heat!29}57.2 & \cellcolor{heat!43}72.3 & \cellcolor{heat!45}\textbf{91.2} & \cellcolor{heat!42}65.2 & \cellcolor{heat!45}\textbf{96.2} & \cellcolor{heat!44}94.7 & \cellcolor{heat!39}41.7 & \cellcolor{heat!42}76.4 & \cellcolor{heat!44}\underline{76.9} & \cellcolor{heat!45}94.8 & \cellcolor{heat!39}95.3 & \cellcolor{heat!37}47.6 & \cellcolor{heat!44}\underline{77.6} \\
PE-Core-L & 0.67B & \cellcolor{heat!43}\underline{82.5} & \cellcolor{heat!41}76.7 & \cellcolor{heat!45}\textbf{83.4} & \cellcolor{heat!43}94.6 & \cellcolor{heat!41}72.7 & \cellcolor{heat!44}80.9 & \cellcolor{heat!33}58.0 & \cellcolor{heat!32}69.4 & \cellcolor{heat!37}85.1 & \cellcolor{heat!42}\underline{65.3} & \cellcolor{heat!45}96.0 & \cellcolor{heat!45}\textbf{95.5} & \cellcolor{heat!39}41.8 & \cellcolor{heat!45}\textbf{77.7} & \cellcolor{heat!39}75.0 & \cellcolor{heat!42}93.2 & \cellcolor{heat!44}\underline{96.9} & \cellcolor{heat!26}44.9 & \cellcolor{heat!43}77.2 \\
\bottomrule
\end{tabular}}
\end{table}
\begin{table}[!tbp]
\centering
\ContinuedFloat
\caption[]{\textbf{Image classification per input resolution.} (continued).}
\scriptsize
\setlength{\tabcolsep}{2.0pt}
\resizebox{\linewidth}{!}{%
\begin{tabular}{@{}l r cccccccccccccccccc c@{}}
\toprule
 & & \multicolumn{7}{c}{Natural images} & \multicolumn{11}{c}{Fine-grained} & \\
\cmidrule(lr){3-9}\cmidrule(lr){10-20}
Model & Params & \rotatebox{90}{ImageNet} & \rotatebox{90}{ImageNet-V2} & \rotatebox{90}{ImageNet-A} & \rotatebox{90}{ImageNet-R} & \rotatebox{90}{ImageNet-Sketch} & \rotatebox{90}{ObjectNet} & \rotatebox{90}{Dollar St.} & \rotatebox{90}{DTD} & \rotatebox{90}{Flowers} & \rotatebox{90}{Aircraft} & \rotatebox{90}{Pets} & \rotatebox{90}{Food101} & \rotatebox{90}{Country211} & \rotatebox{90}{SUN397} & \rotatebox{90}{RESISC45} & \rotatebox{90}{Cars} & \rotatebox{90}{Stanford40} & \rotatebox{90}{PA-100K} & Avg. \\
\midrule
\multicolumn{21}{@{}l}{\textbf{1036\,px}} \\
\addlinespace[1pt]
\textbf{\textcolor{accentorange}{\ours{}}} & 1.24B & \cellcolor{heat!41}82.0 & \cellcolor{heat!45}\textbf{79.2} & \cellcolor{heat!41}86.7 & \cellcolor{heat!41}93.0 & \cellcolor{heat!41}71.4 & \cellcolor{heat!44}83.3 & \cellcolor{heat!32}58.2 & \cellcolor{heat!41}71.0 & \cellcolor{heat!45}\textbf{91.3} & \cellcolor{heat!45}\underline{67.8} & \cellcolor{heat!44}\underline{96.3} & \cellcolor{heat!44}94.2 & \cellcolor{heat!37}47.1 & \cellcolor{heat!44}\underline{77.5} & \cellcolor{heat!37}67.5 & \cellcolor{heat!44}\underline{95.0} & \cellcolor{heat!39}95.8 & \cellcolor{heat!25}44.2 & \cellcolor{heat!43}\underline{77.9} \\
PE-Core-L & 0.67B & \cellcolor{heat!1}65.3 & \cellcolor{heat!10}62.1 & \cellcolor{heat!19}62.3 & \cellcolor{heat!2}66.1 & 40.8 & \cellcolor{heat!18}64.9 & 51.2 & 44.4 & \cellcolor{heat!13}69.1 & \cellcolor{heat!28}45.9 & \cellcolor{heat!31}89.0 & 44.6 & \cellcolor{heat!18}29.8 & 59.9 & 24.4 & \cellcolor{heat!14}73.8 & 84.6 & \cellcolor{heat!14}40.9 & \cellcolor{heat!1}56.6 \\
\bottomrule
\end{tabular}}
\end{table}

\begin{table}[!tbp]
\centering
\caption{\textbf{Image--text retrieval per input resolution.} R@1 (XM3600: mean over 36 languages). Each cell is the best of three input transforms at that input resolution; Best: best resolution per dataset, as in the main tables. Bold: best, underlined: second best among models up to 4B parameters within each block. --: not available.}
\label{tab:app_perres_image_ret}
\scriptsize
\setlength{\tabcolsep}{2.6pt}
\makebox[\linewidth][c]{%
\begin{tabular}{@{}l r cccccccccccc c@{}}
\toprule
 & & \multicolumn{2}{c}{COCO} & \multicolumn{2}{c}{Flickr30k} & \multicolumn{2}{c}{TextCaps} & \multicolumn{2}{c}{DOCCI} & \multicolumn{2}{c}{XM3600} & \multicolumn{2}{c}{VisualNews} & \\
\cmidrule(lr){3-4}\cmidrule(lr){5-6}\cmidrule(lr){7-8}\cmidrule(lr){9-10}\cmidrule(lr){11-12}\cmidrule(lr){13-14}
Model & Params & t2i & i2t & t2i & i2t & t2i & i2t & t2i & i2t & t2i & i2t & t2i & i2t & Avg. \\
\midrule
\multicolumn{15}{@{}l}{\textbf{224\,px}} \\
\addlinespace[1pt]
\textbf{\textcolor{accentorange}{\ours{}}} & 1.24B & \cellcolor{heat!40}53.9 & \cellcolor{heat!42}\underline{72.9} & \cellcolor{heat!43}82.6 & \cellcolor{heat!44}\underline{94.5} & \cellcolor{heat!45}\textbf{74.0} & \cellcolor{heat!45}\textbf{89.5} & \cellcolor{heat!45}\textbf{91.8} & \cellcolor{heat!45}\textbf{89.9} & \cellcolor{heat!45}\textbf{55.7} & \cellcolor{heat!45}\textbf{60.4} & \cellcolor{heat!44}84.4 & \cellcolor{heat!45}\underline{88.5} & \cellcolor{heat!45}\textbf{78.2} \\
PE-Core-L & 0.67B & \cellcolor{heat!44}\underline{55.4} & \cellcolor{heat!45}\textbf{74.0} & \cellcolor{heat!45}\textbf{84.6} & \cellcolor{heat!45}\textbf{95.3} & \cellcolor{heat!45}\underline{73.9} & \cellcolor{heat!44}\underline{89.1} & \cellcolor{heat!25}70.1 & \cellcolor{heat!12}67.6 & \cellcolor{heat!6}10.9 & \cellcolor{heat!8}14.6 & \cellcolor{heat!35}76.3 & \cellcolor{heat!39}82.1 & \cellcolor{heat!30}66.2 \\
PE-Core-G & 2.42B & \cellcolor{heat!35}52.1 & \cellcolor{heat!32}68.1 & \cellcolor{heat!39}79.3 & \cellcolor{heat!37}89.3 & \cellcolor{heat!41}71.2 & \cellcolor{heat!41}87.4 & \cellcolor{heat!36}81.7 & \cellcolor{heat!29}79.3 & \cellcolor{heat!8}12.9 & \cellcolor{heat!10}16.4 & \cellcolor{heat!39}80.4 & \cellcolor{heat!40}82.9 & \cellcolor{heat!30}66.7 \\
\bottomrule
\end{tabular}}
\end{table}
\begin{table}[!tbp]
\centering
\ContinuedFloat
\caption[]{\textbf{Image--text retrieval per input resolution.} (continued).}
\scriptsize
\setlength{\tabcolsep}{2.6pt}
\makebox[\linewidth][c]{%
\begin{tabular}{@{}l r cccccccccccc c@{}}
\toprule
 & & \multicolumn{2}{c}{COCO} & \multicolumn{2}{c}{Flickr30k} & \multicolumn{2}{c}{TextCaps} & \multicolumn{2}{c}{DOCCI} & \multicolumn{2}{c}{XM3600} & \multicolumn{2}{c}{VisualNews} & \\
\cmidrule(lr){3-4}\cmidrule(lr){5-6}\cmidrule(lr){7-8}\cmidrule(lr){9-10}\cmidrule(lr){11-12}\cmidrule(lr){13-14}
Model & Params & t2i & i2t & t2i & i2t & t2i & i2t & t2i & i2t & t2i & i2t & t2i & i2t & Avg. \\
\midrule
\multicolumn{15}{@{}l}{\textbf{1036\,px}} \\
\addlinespace[1pt]
\textbf{\textcolor{accentorange}{\ours{}}} & 1.24B & \cellcolor{heat!36}55.2 & \cellcolor{heat!42}\underline{73.8} & \cellcolor{heat!38}83.2 & \cellcolor{heat!40}95.1 & \cellcolor{heat!37}\underline{79.3} & \cellcolor{heat!43}\underline{93.3} & \cellcolor{heat!43}\underline{93.8} & \cellcolor{heat!43}91.2 & \cellcolor{heat!45}\textbf{56.5} & \cellcolor{heat!45}\textbf{60.1} & \cellcolor{heat!44}\underline{85.6} & \cellcolor{heat!45}\underline{88.4} & \cellcolor{heat!45}\textbf{79.6} \\
PE-Core-L & 0.67B & \cellcolor{heat!5}41.5 & 49.5 & \cellcolor{heat!11}72.3 & 82.6 & \cellcolor{heat!31}72.5 & \cellcolor{heat!29}81.6 & 52.6 & 43.2 & \cellcolor{heat!5}9.9 & \cellcolor{heat!5}11.7 & \cellcolor{heat!27}70.6 & \cellcolor{heat!24}71.4 & 55.0 \\
PE-Core-G & 2.42B & \cellcolor{heat!33}53.9 & \cellcolor{heat!36}70.1 & \cellcolor{heat!44}85.3 & \cellcolor{heat!41}95.3 & \cellcolor{heat!33}74.5 & \cellcolor{heat!40}90.8 & \cellcolor{heat!31}82.1 & \cellcolor{heat!31}78.0 & \cellcolor{heat!6}11.6 & \cellcolor{heat!9}16.4 & \cellcolor{heat!42}83.7 & \cellcolor{heat!43}86.5 & \cellcolor{heat!26}69.0 \\
MetaCLIP~2-H & 1.86B & 39.1 & \cellcolor{heat!7}53.7 & 67.9 & \cellcolor{heat!2}83.3 & 42.1 & 57.1 & \cellcolor{heat!15}67.4 & \cellcolor{heat!19}64.7 & \cellcolor{heat!37}47.8 & \cellcolor{heat!37}50.2 & \cellcolor{heat!31}74.2 & \cellcolor{heat!31}77.1 & \cellcolor{heat!10}60.4 \\
\bottomrule
\end{tabular}}
\end{table}

\begin{table}[!tbp]
\centering
\caption{\textbf{Video classification per input resolution.} Top-1 accuracy, 16 frames (K400/600/700: Kinetics; Kitchens: EPIC-Kitchens-100). Each cell is the best of three input transforms at that input resolution; Best: best resolution per dataset, as in the main tables. Bold: best, underlined: second best among models up to 4B parameters within each block. --: not available.}
\label{tab:app_perres_video_cls}
\scriptsize
\setlength{\tabcolsep}{2.6pt}
\makebox[\linewidth][c]{%
\begin{tabular}{@{}l r cccccccc c@{}}
\toprule
Model & Params & K400 & K600 & K700 & UCF101 & HMDB51 & SSv2 & Kitchens & Breakfast & Avg. \\
\midrule
\multicolumn{11}{@{}l}{\textbf{224\,px}} \\
\addlinespace[1pt]
\textbf{\textcolor{accentorange}{\ours{}}} & 1.24B & \cellcolor{heat!44}\underline{72.5} & \cellcolor{heat!44}\underline{72.2} & \cellcolor{heat!43}63.4 & \cellcolor{heat!45}\textbf{89.7} & \cellcolor{heat!29}\underline{63.5} & \cellcolor{heat!45}\textbf{24.0} & \cellcolor{heat!21}\underline{6.9} & \cellcolor{heat!44}\underline{54.6} & \cellcolor{heat!44}\underline{55.8} \\
PE-Core-L & 0.67B & \cellcolor{heat!43}72.2 & \cellcolor{heat!43}71.6 & \cellcolor{heat!43}\underline{63.5} & \cellcolor{heat!35}86.1 & \cellcolor{heat!26}60.9 & \cellcolor{heat!10}9.3 & \cellcolor{heat!10}4.4 & \cellcolor{heat!24}41.6 & \cellcolor{heat!33}51.2 \\
PE-Core-G & 2.42B & \cellcolor{heat!45}\textbf{73.0} & \cellcolor{heat!45}\textbf{72.6} & \cellcolor{heat!45}\textbf{64.6} & \cellcolor{heat!43}88.9 & \cellcolor{heat!28}62.7 & \cellcolor{heat!12}9.9 & \cellcolor{heat!12}4.8 & \cellcolor{heat!23}40.9 & \cellcolor{heat!35}52.2 \\
MetaCLIP~2-H & 1.86B & \cellcolor{heat!34}68.4 & \cellcolor{heat!37}68.6 & \cellcolor{heat!34}59.1 & \cellcolor{heat!31}84.4 & \cellcolor{heat!23}58.1 & \cellcolor{heat!6}7.4 & \cellcolor{heat!6}3.4 & \cellcolor{heat!14}34.7 & \cellcolor{heat!25}48.0 \\
MetaCLIP~2-bigG & 3.63B & \cellcolor{heat!31}66.8 & \cellcolor{heat!33}66.8 & \cellcolor{heat!29}56.9 & \cellcolor{heat!35}85.9 & \cellcolor{heat!20}55.9 & 5.1 & 1.9 & \cellcolor{heat!4}27.6 & \cellcolor{heat!20}45.9 \\
SigLIP2-So400m & 1.14B & \cellcolor{heat!27}65.2 & \cellcolor{heat!29}65.0 & \cellcolor{heat!27}55.7 & \cellcolor{heat!38}87.2 & \cellcolor{heat!16}52.1 & \cellcolor{heat!6}7.6 & \cellcolor{heat!6}3.3 & \cellcolor{heat!20}38.3 & \cellcolor{heat!22}46.8 \\
\bottomrule
\end{tabular}}
\end{table}
\begin{table}[!tbp]
\centering
\ContinuedFloat
\caption[]{\textbf{Video classification per input resolution.} (continued).}
\scriptsize
\setlength{\tabcolsep}{2.6pt}
\makebox[\linewidth][c]{%
\begin{tabular}{@{}l r cccccccc c@{}}
\toprule
Model & Params & K400 & K600 & K700 & UCF101 & HMDB51 & SSv2 & Kitchens & Breakfast & Avg. \\
\midrule
\multicolumn{11}{@{}l}{\textbf{1036\,px}} \\
\addlinespace[1pt]
\textbf{\textcolor{accentorange}{\ours{}}} & 1.24B & \cellcolor{heat!42}\underline{72.2} & \cellcolor{heat!43}\underline{71.9} & \cellcolor{heat!41}\underline{63.3} & \cellcolor{heat!44}\underline{88.0} & \cellcolor{heat!25}61.0 & \cellcolor{heat!45}\textbf{23.8} & \cellcolor{heat!25}\underline{7.7} & \cellcolor{heat!35}49.9 & \cellcolor{heat!41}\underline{54.7} \\
PE-Core-L & 0.67B & \cellcolor{heat!3}54.0 & \cellcolor{heat!3}52.5 & \cellcolor{heat!2}42.8 & \cellcolor{heat!3}70.5 & \cellcolor{heat!3}45.4 & 5.0 & 1.6 & \cellcolor{heat!4}25.2 & \cellcolor{heat!2}37.1 \\
PE-Core-G & 2.42B & \cellcolor{heat!45}\textbf{73.7} & \cellcolor{heat!45}\textbf{72.8} & \cellcolor{heat!45}\textbf{65.1} & \cellcolor{heat!45}\textbf{88.3} & \cellcolor{heat!26}\underline{61.1} & \cellcolor{heat!10}9.4 & \cellcolor{heat!15}5.3 & \cellcolor{heat!23}40.3 & \cellcolor{heat!35}52.0 \\
MetaCLIP~2-H & 1.86B & 52.7 & \cellcolor{heat!1}51.4 & 41.5 & 69.4 & 43.4 & \cellcolor{heat!2}5.8 & \cellcolor{heat!1}1.8 & 22.4 & 36.1 \\
MetaCLIP~2-bigG & 3.63B & \cellcolor{heat!31}67.3 & \cellcolor{heat!33}67.1 & \cellcolor{heat!31}57.5 & \cellcolor{heat!35}84.2 & \cellcolor{heat!16}54.4 & \cellcolor{heat!2}5.9 & \cellcolor{heat!4}2.6 & \cellcolor{heat!11}31.1 & \cellcolor{heat!22}46.3 \\
SigLIP2-So400m & 1.14B & \cellcolor{heat!29}66.0 & \cellcolor{heat!30}65.7 & \cellcolor{heat!29}56.7 & \cellcolor{heat!43}87.6 & \cellcolor{heat!13}52.4 & \cellcolor{heat!6}7.5 & \cellcolor{heat!11}4.3 & \cellcolor{heat!25}41.8 & \cellcolor{heat!26}47.8 \\
\bottomrule
\end{tabular}}
\end{table}

\begin{table}[!tbp]
\centering
\caption{\textbf{Video--text retrieval per input resolution.} R@1, 16 frames (MSR-VTT: full test split). Each cell is the best of three input transforms at that input resolution; Best: best resolution per dataset, as in the main tables. Bold: best, underlined: second best among models up to 4B parameters within each block. --: not available.}
\label{tab:app_perres_video_ret}
\scriptsize
\setlength{\tabcolsep}{2.6pt}
\makebox[\linewidth][c]{%
\begin{tabular}{@{}l r cccccccccc c@{}}
\toprule
 & & \multicolumn{2}{c}{MSR-VTT} & \multicolumn{2}{c}{MSVD} & \multicolumn{2}{c}{ActivityNet} & \multicolumn{2}{c}{PVD} & \multicolumn{2}{c}{DiDeMo} & \\
\cmidrule(lr){3-4}\cmidrule(lr){5-6}\cmidrule(lr){7-8}\cmidrule(lr){9-10}\cmidrule(lr){11-12}
Model & Params & t2v & v2t & t2v & v2t & t2v & v2t & t2v & v2t & t2v & v2t & Avg. \\
\midrule
\multicolumn{13}{@{}l}{\textbf{224\,px}} \\
\addlinespace[1pt]
\textbf{\textcolor{accentorange}{\ours{}}} & 1.24B & \cellcolor{heat!43}39.1 & \cellcolor{heat!45}\textbf{72.9} & \cellcolor{heat!39}59.8 & \cellcolor{heat!45}\underline{82.5} & \cellcolor{heat!45}\textbf{58.4} & \cellcolor{heat!45}\textbf{57.8} & \cellcolor{heat!45}\textbf{80.8} & \cellcolor{heat!45}\textbf{80.7} & \cellcolor{heat!34}\underline{49.4} & \cellcolor{heat!45}\textbf{51.8} & \cellcolor{heat!45}\textbf{63.3} \\
PE-Core-L & 0.67B & \cellcolor{heat!42}38.7 & \cellcolor{heat!41}69.1 & \cellcolor{heat!45}\textbf{63.0} & \cellcolor{heat!45}\textbf{82.8} & \cellcolor{heat!28}43.5 & \cellcolor{heat!23}40.7 & \cellcolor{heat!27}63.3 & \cellcolor{heat!21}65.4 & \cellcolor{heat!24}40.9 & \cellcolor{heat!28}40.8 & \cellcolor{heat!33}54.8 \\
PE-Core-G & 2.42B & \cellcolor{heat!41}38.2 & \cellcolor{heat!38}66.5 & \cellcolor{heat!43}61.6 & \cellcolor{heat!40}79.9 & \cellcolor{heat!34}48.5 & \cellcolor{heat!29}\underline{45.6} & \cellcolor{heat!40}75.6 & \cellcolor{heat!37}\underline{75.6} & \cellcolor{heat!27}43.4 & \cellcolor{heat!31}42.8 & \cellcolor{heat!37}57.8 \\
MetaCLIP~2-H & 1.86B & \cellcolor{heat!28}31.9 & \cellcolor{heat!19}49.6 & \cellcolor{heat!33}56.3 & \cellcolor{heat!24}69.6 & \cellcolor{heat!19}36.2 & \cellcolor{heat!12}32.2 & \cellcolor{heat!13}49.6 & \cellcolor{heat!18}62.9 & \cellcolor{heat!15}34.4 & \cellcolor{heat!11}29.6 & \cellcolor{heat!19}45.2 \\
\bottomrule
\end{tabular}}
\end{table}
\begin{table}[!tbp]
\centering
\ContinuedFloat
\caption[]{\textbf{Video--text retrieval per input resolution.} (continued).}
\scriptsize
\setlength{\tabcolsep}{2.6pt}
\makebox[\linewidth][c]{%
\begin{tabular}{@{}l r cccccccccc c@{}}
\toprule
 & & \multicolumn{2}{c}{MSR-VTT} & \multicolumn{2}{c}{MSVD} & \multicolumn{2}{c}{ActivityNet} & \multicolumn{2}{c}{PVD} & \multicolumn{2}{c}{DiDeMo} & \\
\cmidrule(lr){3-4}\cmidrule(lr){5-6}\cmidrule(lr){7-8}\cmidrule(lr){9-10}\cmidrule(lr){11-12}
Model & Params & t2v & v2t & t2v & v2t & t2v & v2t & t2v & v2t & t2v & v2t & Avg. \\
\midrule
\multicolumn{13}{@{}l}{\textbf{1036\,px}} \\
\addlinespace[1pt]
\textbf{\textcolor{accentorange}{\ours{}}} & 1.24B & \cellcolor{heat!45}\textbf{38.0} & \cellcolor{heat!45}\textbf{70.7} & \cellcolor{heat!37}59.1 & \cellcolor{heat!45}\textbf{80.7} & \cellcolor{heat!45}\textbf{59.6} & \cellcolor{heat!45}\textbf{58.7} & \cellcolor{heat!43}81.2 & \cellcolor{heat!45}\textbf{81.3} & \cellcolor{heat!41}48.5 & \cellcolor{heat!45}\textbf{51.1} & \cellcolor{heat!45}\textbf{62.9} \\
PE-Core-L & 0.67B & \cellcolor{heat!9}26.4 & \cellcolor{heat!8}39.1 & 43.5 & 51.8 & \cellcolor{heat!6}32.1 & \cellcolor{heat!2}26.1 & \cellcolor{heat!12}55.7 & 51.6 & \cellcolor{heat!4}31.3 & \cellcolor{heat!4}24.6 & \cellcolor{heat!4}38.2 \\
PE-Core-G & 2.42B & \cellcolor{heat!42}37.2 & \cellcolor{heat!38}64.4 & \cellcolor{heat!45}\textbf{62.3} & \cellcolor{heat!44}\underline{80.3} & \cellcolor{heat!31}50.0 & \cellcolor{heat!30}47.1 & \cellcolor{heat!36}75.3 & \cellcolor{heat!35}75.0 & \cellcolor{heat!28}42.1 & \cellcolor{heat!34}44.3 & \cellcolor{heat!37}57.8 \\
MetaCLIP~2-H & 1.86B & 23.6 & 32.3 & \cellcolor{heat!4}45.3 & 51.8 & 27.9 & 24.4 & 45.6 & \cellcolor{heat!1}52.4 & 29.2 & 22.2 & 35.5 \\
MetaCLIP~2-bigG & 3.63B & \cellcolor{heat!24}31.2 & \cellcolor{heat!19}48.7 & \cellcolor{heat!29}55.8 & \cellcolor{heat!24}67.5 & \cellcolor{heat!12}36.7 & \cellcolor{heat!11}33.0 & \cellcolor{heat!5}49.7 & \cellcolor{heat!14}60.7 & \cellcolor{heat!10}33.7 & \cellcolor{heat!11}29.3 & \cellcolor{heat!15}44.6 \\
\bottomrule
\end{tabular}}
\end{table}
\FloatBarrier

\FloatBarrier
\section{Additional ablations}
\label{app:extra_ablations}

\subsection{ST encoder alignment}
\label{app:st_alignment}
In the ST encoder alignment phase, only the newly added ST encoder trains; full fine-tuning then unfreezes all
parameters (\cref{sec:method_video}).
In the final recipe, the alignment runs first on the open-source mixture and then on the curated mixture.
The open-source mixture has about 5$\times$ more clips than the curated one (48M vs.\ 8.8M) but is noisier, so
we test whether it helps the alignment.

\begin{figure}[tbp]
\centering
\includegraphics[width=\linewidth]{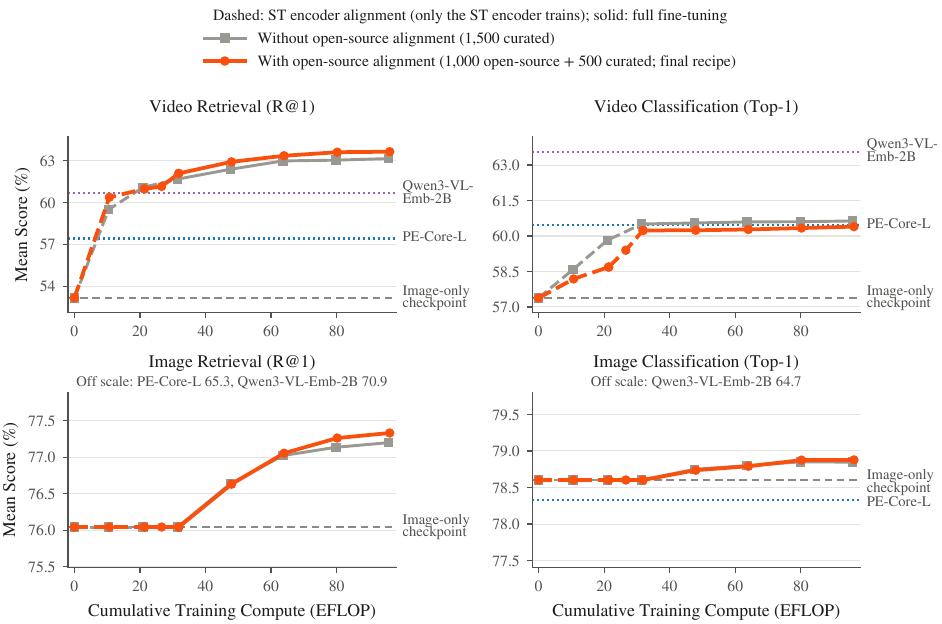}
\caption{\textbf{Open-source vs.\ curated mixture for ST encoder alignment.}
Mean score at 224\,px against cumulative training compute.
Orange (final recipe): 1{,}000 alignment steps on the open-source mixture, then 500 on the curated mixture;
gray: 1{,}500 alignment steps on the curated mixture.
Both then run 1{,}000 full fine-tuning steps on the curated mixture.
Dashed: ST encoder alignment; solid: full fine-tuning; gray dashed line: image-only checkpoint; dotted:
PE-Core-L and Qwen3-VL-Embedding-2B. Single runs.}
\label{fig:abl_warmup_50m}
\end{figure}

The reference run follows the final recipe described in \cref{sec:method_video} at this reduced scale: it
aligns the ST encoder for 1{,}000 steps on the open-source mixture and then 500 steps on the curated mixture.
The ablated run uses the curated mixture for the first 1{,}000 steps as well.
Both runs then perform 1{,}000 full fine-tuning steps on the curated mixture, so their training compute is
matched; \cref{fig:abl_warmup_50m} reports their performance against training compute.

During the alignment, image scores stay at the image-only checkpoint, since the image path does not use the ST
encoder, while video retrieval rises quickly in both runs.
After full fine-tuning, the reference run is 0.5 points higher in video retrieval, and the two runs differ by
at most 0.24 points in image retrieval, image classification and video classification.
The open-source mixture therefore helps the alignment, but only modestly; we keep it in the final recipe
(\cref{sec:method_video}), where the alignment covers about one epoch of each mixture.

\paragraph{Setup.}
Both runs in \cref{fig:abl_warmup_50m} start from the Stage~1 image-only checkpoint and use 64 H200 GPUs, 96k
tokens per GPU (6.144M tokens per step), 16 uniform video frames and no image replay.
The alignment phase trains only the ST encoder and the contrastive projector, at a peak learning rate of
$10^{-4}$; full fine-tuning trains all parameters, at a peak learning rate of $4.2{\times}10^{-7}$. Each phase
uses AdamW with weight decay 0.1, 50 warm-up steps and a cosine-then-linear schedule down to $0.01\times$ the
peak learning rate.

\subsection{Frame sampling per video}
\label{app:frames}
In \cref{sec:method_video}, we sample $T$ uniformly spaced frames per video clip.
Here we ablate the number of frames $T$ and the sampling method, since both could affect the model's ability
to discriminate actions that unfold over time.
Every run has the same FLOP-matched compute budget on 32 GPUs.
Training follows the full fine-tuning setup of \cref{app:st_alignment}, without the ST encoder and with a
lower peak learning rate of $3{\times}10^{-7}$.

\begin{figure}[tbp]
\centering
\includegraphics[width=\linewidth]{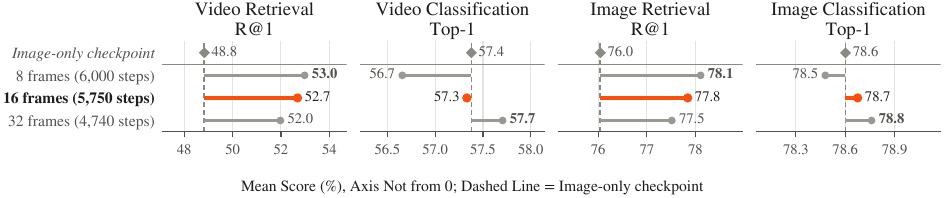}
\caption{\textbf{Number of frames per clip at fixed compute.} PVD excluded.
Classification and retrieval for 8, 16 and 32 uniform frames after 5 hours on 32 GPUs.}
\label{fig:app_frames}
\end{figure}

\paragraph{Number of frames.}
We find that the number of frames $T$ per clip trades retrieval against classification
(\cref{fig:app_frames}).
The smaller $T$, the better the retrieval, as fewer frames per clip allow more clips per batch and therefore
more negatives: eight frames give the strongest retrieval but the weakest video classification.
Instead, the larger $T$, the better the video classification: 32 frames give the best video classification but
the lowest video retrieval.
We keep 16 frames as a compromise in this retrieval--classification trade-off.

\begin{figure}[tbp]
\centering
\includegraphics[width=\linewidth]{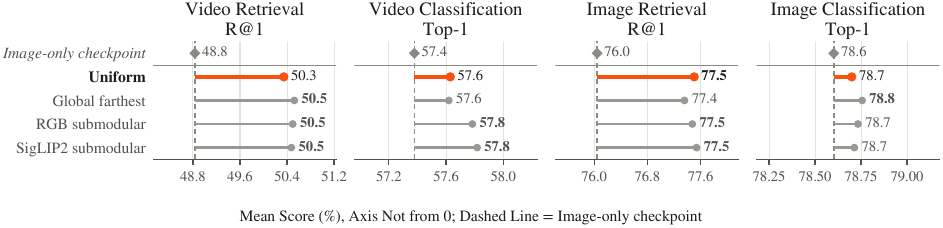}
\caption{\textbf{Content-aware vs.\ uniform frame selection.} PVD excluded.
Selecting 8 of 32 frames by global-farthest, RGB-submodular or SigLIP2-submodular selection, vs.\ uniform
sampling; evaluation with 16 uniform frames.}
\label{fig:app_smart}
\end{figure}

\paragraph{Content-aware sampling.}
Uniform sampling can spend frames on redundant or uninformative moments, and prior work selects frames by
content instead, for action recognition~\citep{wu2019adaframe,korbar2019scsampler,zhi2021mgsampler} and for
long-video understanding with VLMs~\citep{tang2025aks}.
We compare uniform sampling of 8 frames with three content-aware selections of 8 of 32 stored frames
(\cref{fig:app_smart}).
For each ablation, we first discard near-uniform frames such as black frames and fades (pixel standard
deviation below 5 on the 0--255 scale).
Then, \emph{global farthest} keeps the 8 frames whose $32{\times}32$ RGB thumbnails differ most from the
clip's mean thumbnail.
The two \emph{submodular} methods greedily maximize a sum of a coverage term (facility location: every frame
should be similar to a selected one) and a diversity term (dispersion: selected frames should differ from each
other)~\citep{lin2011submodular,gygli2015submodular}, computed on cosine similarities between either the RGB
thumbnails or SigLIP2 image embeddings.
Content-aware selection is within 0.2 points of uniform sampling in image and video retrieval and
classification, i.e., it gives no measurable gain.
We therefore keep uniform frames, which add no preprocessing cost, although content-aware selection may matter
more for longer videos.

\subsection{Approaches that did not work}
\label{app:negative}
\emph{Temporal attention inside the ViT.} Making the ViT's own attention joint over all the frames (using 2D
RoPE plus a time encoding, or 3D RoPE, as positional encoding) degraded image accuracy for any learning rate
large enough to change video accuracy.
This is presumably because the change was too large for a pretrained 2D encoder without full retraining.
A separate ST module avoids this problem, and it is also computationally more efficient.

\end{document}